\documentclass[times,preprint,10pt]{elsarticle}
\providecommand*{\WWbuild}{submission}
\def\WWinternalname{internal}
\newif\ifinternal
\ifx\WWbuild\WWinternalname\internaltrue\fi

\usepackage[T1]{fontenc}
\usepackage[utf8]{inputenc}

\usepackage{amsmath}
\usepackage{graphicx}
\usepackage{xcolor}
\usepackage{xspace}
\usepackage{booktabs}
\usepackage{multirow}
\usepackage{makecell}
\usepackage{array}
\usepackage{url}
\usepackage{hyperref}
\hypersetup{hidelinks}
\makeatletter
\def\ps@pprintTitle{\let\@oddhead\@empty\let\@evenhead\@empty\def\@oddfoot{\hfil\thepage\hfil}\let\@evenfoot\@oddfoot}
\makeatother
\usepackage[a4paper,margin=25mm]{geometry}

\usepackage{siunitx}
\usepackage{placeins}
\usepackage{tabularx}
\usepackage{lastpage}
\AtBeginDocument{}
\newcommand{\method}{When2What\xspace}

\newcommand{\WWfig}[2]{\IfFileExists{#1}{\includegraphics[width=#2]{#1}}{\fbox{\parbox{#2}{\centering\small Figure file \texttt{\detokenize{#1}} not generated yet.}}}}

\newcommand{\NwinDisc}{2{,}286\xspace}
\newcommand{\NpatDisc}{845\xspace}
\newcommand{\NwinLbx}{2{,}267\xspace}
\newcommand{\NpatLbx}{832\xspace}
\newcommand{\NwinVal}{2{,}425\xspace}
\newcommand{\NexclDisc}{11.3\%\xspace}
\newcommand{\NexclLbx}{12.0\%\xspace}
\newcommand{\NexclVal}{11.2\%\xspace}
\newcommand{\NrefCheckPeaks}{2{,}790\xspace}
\newcommand{\NrefCheckErrRateHi}{0.132\%\xspace}
\newcommand{\NrefAgreement}{100\%\xspace}
\newcommand{\NroCFoneVal}{0.968\xspace}
\newcommand{\NroAFoneVal}{0.951\xspace}

\newcommand{\NroCFoneDiscFull}{0.970~[0.967, 0.973]\xspace}

\newcommand{\NroAFoneDiscFull}{0.947~[0.941, 0.953]\xspace}
\newcommand{\NdcondOrigDisc}{\ensuremath{+}0.110\xspace}

\newcommand{\NdcondOrigDiscFull}{\ensuremath{+}0.110~[\ensuremath{+}0.102, \ensuremath{+}0.117]\xspace}

\newcommand{\NdcondOrigLbx}{\ensuremath{+}0.111\xspace}

\newcommand{\NdcondOrigLbxFull}{\ensuremath{+}0.111~[\ensuremath{+}0.103, \ensuremath{+}0.118]\xspace}
\newcommand{\NdcondSplitDisc}{\ensuremath{+}0.067\xspace}

\newcommand{\NdcondSplitDiscFull}{\ensuremath{+}0.067~[\ensuremath{+}0.061, \ensuremath{+}0.074]\xspace}
\newcommand{\NdcondSplitLbx}{\ensuremath{+}0.069\xspace}

\newcommand{\NdcondSplitLbxFull}{\ensuremath{+}0.069~[\ensuremath{+}0.063, \ensuremath{+}0.076]\xspace}

\newcommand{\NratioDiscFull}{0.61~[0.57, 0.65]\xspace}

\newcommand{\NdcondTCNDisc}{\ensuremath{+}0.097\xspace}

\newcommand{\NdcondTCNDiscFull}{\ensuremath{+}0.097~[\ensuremath{+}0.089, \ensuremath{+}0.104]\xspace}
\newcommand{\NdcondMOVER}{\ensuremath{+}0.097\xspace}

\newcommand{\NdcondMOVERFull}{\ensuremath{+}0.097~[\ensuremath{+}0.089, \ensuremath{+}0.106]\xspace}
\newcommand{\NdcondSHDB}{\ensuremath{+}0.110\xspace}

\newcommand{\NdcondSHDBFull}{\ensuremath{+}0.110~[\ensuremath{+}0.100, \ensuremath{+}0.120]\xspace}
\newcommand{\NdcondSplitTunedDisc}{\ensuremath{+}0.074\xspace}

\newcommand{\NdcondSplitTunedDiscFull}{\ensuremath{+}0.074~[\ensuremath{+}0.068, \ensuremath{+}0.081]\xspace}
\newcommand{\NrecoveryDisc}{0.24\xspace}

\newcommand{\NrecoveryTunedDisc}{0.21\xspace}

\newcommand{\NdcondOrigDiscRoAFull}{\ensuremath{+}0.135~[\ensuremath{+}0.123, \ensuremath{+}0.146]\xspace}

\newcommand{\NdcondOrigDiscRoBFull}{\ensuremath{+}0.165~[\ensuremath{+}0.154, \ensuremath{+}0.176]\xspace}

\newcommand{\NratioDiscRoAFull}{0.48~[0.44, 0.52]\xspace}

\newcommand{\NratioDiscRoBFull}{0.57~[0.53, 0.60]\xspace}
\newcommand{\NwaveDirectDisc}{0.537\xspace}

\newcommand{\NwaveDirectDiscCI}{[0.521, 0.554]\xspace}

\newcommand{\NwaveDirectMTDisc}{0.533\xspace}

\newcommand{\NwaveDirectMTDiscCI}{[0.517, 0.550]\xspace}

\newcommand{\NnatDirectMTDisc}{0.670\xspace}

\newcommand{\NnatDirectMTDiscCI}{[0.654, 0.685]\xspace}

\newcommand{\NwaveWWDisc}{0.644\xspace}

\newcommand{\NwaveWWDiscCI}{[0.628, 0.659]\xspace}
\newcommand{\NwaveWWDiscFull}{0.644~[0.628, 0.659]\xspace}
\newcommand{\NnatWWDisc}{0.672\xspace}

\newcommand{\NnatWWDiscCI}{[0.657, 0.688]\xspace}
\newcommand{\NnatWWDiscFull}{0.672~[0.657, 0.688]\xspace}

\newcommand{\NwaveDirectSplitDisc}{0.559\xspace}

\newcommand{\NwaveDirectSplitDiscCI}{[0.544, 0.576]\xspace}

\newcommand{\NwaveDirectMTSplitDisc}{0.560\xspace}

\newcommand{\NwaveDirectMTSplitDiscCI}{[0.545, 0.577]\xspace}

\newcommand{\NnatDirectMTSplitDisc}{0.669\xspace}

\newcommand{\NnatDirectMTSplitDiscCI}{[0.654, 0.685]\xspace}

\newcommand{\NwaveWWSplitDisc}{0.627\xspace}

\newcommand{\NwaveWWSplitDiscCI}{[0.612, 0.642]\xspace}

\newcommand{\NnatWWSplitDisc}{0.655\xspace}

\newcommand{\NnatWWSplitDiscCI}{[0.640, 0.670]\xspace}

\newcommand{\NwaveDirectLbx}{0.544\xspace}

\newcommand{\NwaveDirectMTLbx}{0.541\xspace}

\newcommand{\NnatDirectMTLbx}{0.677\xspace}

\newcommand{\NwaveWWLbx}{0.651\xspace}

\newcommand{\NnatWWLbx}{0.682\xspace}

\newcommand{\NwaveDirectSplitLbx}{0.566\xspace}

\newcommand{\NwaveDirectMTSplitLbx}{0.567\xspace}

\newcommand{\NnatDirectMTSplitLbx}{0.676\xspace}

\newcommand{\NwaveWWSplitLbx}{0.637\xspace}

\newcommand{\NnatWWSplitLbx}{0.666\xspace}

\newcommand{\NguardedDiffDiscFull}{\ensuremath{+}0.032~[\ensuremath{+}0.023, \ensuremath{+}0.041]\xspace}

\newcommand{\NguardedDiffLbxFull}{\ensuremath{+}0.030~[\ensuremath{+}0.021, \ensuremath{+}0.039]\xspace}
\newcommand{\NtauWWDisc}{7.36\xspace}

\newcommand{\NtauCNDisc}{7.26\xspace}

\newcommand{\NmorphRStart}{0.995\xspace}

\newcommand{\NmorphREnd}{0.982\xspace}

\newcommand{\NshapleyTiming}{0.989\xspace}

\newcommand{\NshapleyTimingFull}{0.989~[0.987, 0.991]\xspace}

\newcommand{\NheadRendererDisc}{\ensuremath{-}0.0025\xspace}

\newcommand{\NheadRendererDiscFull}{\ensuremath{-}0.0025~[\ensuremath{-}0.0051, \ensuremath{+}0.0002]\xspace}

\newcommand{\NifSoft}{\ensuremath{-}0.203\xspace}

\newcommand{\NifSoftFull}{\ensuremath{-}0.203~[\ensuremath{-}0.224, \ensuremath{-}0.184]\xspace}
\newcommand{\NmseDirectDisc}{0.959\xspace}

\newcommand{\NmseDirectDiscCI}{[0.950, 0.969]\xspace}

\newcommand{\NmseWWDisc}{1.687\xspace}

\newcommand{\NmseWWDiscCI}{[1.674, 1.701]\xspace}

\newcommand{\NafGuardedDiffFull}{\ensuremath{-}0.0333~[\ensuremath{-}0.0496, \ensuremath{-}0.0207]\xspace}

\newcommand{\NmoverGuardedDiffFull}{\ensuremath{+}0.0005~[\ensuremath{-}0.0094, \ensuremath{+}0.0106]\xspace}
\newcommand{\NectopicDisc}{\ensuremath{-}1.8\%\xspace}

\newcommand{\NdilateSkill}{0.075\xspace}

\newcommand{\NsynTimingSDlow}{4\xspace}
\newcommand{\NsynTimingSDmid}{11\xspace}
\newcommand{\NsynCommitGainMid}{\ensuremath{+}0.139\xspace}

\newcommand{\NsynDirectFoneB}{0.134\xspace}
\newcommand{\NsynCommitFoneB}{0.294\xspace}

\newcommand{\NsoftBeatRmseDiscFull}{\ensuremath{+}0.114~[\ensuremath{+}0.107, \ensuremath{+}0.120]\xspace}
\newcommand{\NppgSkillVsNaive}{\ensuremath{+}0.045\xspace}

\newcommand{\NppgSkillVsNaiveFull}{\ensuremath{+}0.045~[\ensuremath{+}0.035, \ensuremath{+}0.056]\xspace}

\newcommand{\NppgWaveFoneVsDirect}{\ensuremath{+}0.015\xspace}

\newcommand{\NppgWaveFoneVsDirectFull}{\ensuremath{+}0.015~[\ensuremath{+}0.010, \ensuremath{+}0.020]\xspace}

\newcommand{\NppgTauVsNaiveFull}{\ensuremath{-}0.20~[\ensuremath{-}0.53, \ensuremath{+}0.21]\xspace}

\newcommand{\NppgWin}{1{,}860\xspace}
\newcommand{\NppgPat}{792\xspace}

\newcommand{\NppgRefMissedFull}{1.2\%~[0.8\%, 1.6\%]\xspace}

\newcommand{\NppgRefExtraFull}{0.9\%~[0.6\%, 1.3\%]\xspace}
\newcommand{\NppgRefDelay}{714\xspace}

\newcommand{\NabpSkillVsNaive}{\ensuremath{+}0.019\xspace}

\newcommand{\NabpSkillVsNaiveFull}{\ensuremath{+}0.019~[\ensuremath{+}0.008, \ensuremath{+}0.032]\xspace}

\newcommand{\NabpWaveFoneVsDirect}{\ensuremath{+}0.038\xspace}

\newcommand{\NabpWaveFoneVsDirectFull}{\ensuremath{+}0.038~[\ensuremath{+}0.033, \ensuremath{+}0.044]\xspace}

\newcommand{\NabpTauVsNaiveFull}{\ensuremath{-}0.11~[\ensuremath{-}0.58, \ensuremath{+}0.46]\xspace}

\newcommand{\NabpWin}{1{,}118\xspace}
\newcommand{\NabpPat}{438\xspace}

\newcommand{\NabpRefMissedFull}{1.0\%~[0.6\%, 1.4\%]\xspace}

\newcommand{\NabpRefExtraFull}{0.3\%~[0.1\%, 0.6\%]\xspace}
\newcommand{\NabpRefDelay}{212\xspace}

\newcommand{\NecgDiscPatPreGate}{859\xspace}
\newcommand{\NecgLbxPatPreGate}{860\xspace}
\newcommand{\NecgValPatPreGate}{911\xspace}
\newcommand{\NdthreeInfRLev}{\ensuremath{-}0.069\xspace}

\newcommand{\NdthreeInfRLevFull}{\ensuremath{-}0.069~[\ensuremath{-}0.074, \ensuremath{-}0.065]\xspace}
\newcommand{\NdthreeInfRDiff}{\ensuremath{+}0.713\xspace}

\newcommand{\NdthreeInfRDiffFull}{\ensuremath{+}0.713~[\ensuremath{+}0.697, \ensuremath{+}0.729]\xspace}

\newcommand{\NdthreeInfOLevFull}{\ensuremath{-}0.060~[\ensuremath{-}0.064, \ensuremath{-}0.057]\xspace}

\newcommand{\NdthreeInfODiffFull}{\ensuremath{+}0.704~[\ensuremath{+}0.688, \ensuremath{+}0.719]\xspace}
\newcommand{\NdthreeBlxRLev}{0.152\xspace}

\newcommand{\NdthreeBlxRLevFull}{0.152~[0.139, 0.166]\xspace}
\newcommand{\NdthreeBlxRDiff}{\ensuremath{+}0.491\xspace}

\newcommand{\NdthreeBlxRDiffFull}{\ensuremath{+}0.491~[\ensuremath{+}0.474, \ensuremath{+}0.508]\xspace}
\newcommand{\NdthreeHtfOLev}{\ensuremath{-}0.072\xspace}

\newcommand{\NdthreeHtfOLevFull}{\ensuremath{-}0.072~[\ensuremath{-}0.077, \ensuremath{-}0.067]\xspace}
\newcommand{\NdthreeHtfODiff}{\ensuremath{+}0.716\xspace}

\newcommand{\NdthreeHtfODiffFull}{\ensuremath{+}0.716~[\ensuremath{+}0.700, \ensuremath{+}0.731]\xspace}
\newcommand{\NdthreePtALev}{0.712\xspace}

\newcommand{\NdthreePtALevFull}{0.712~[0.697, 0.727]\xspace}
\newcommand{\NdthreePtADiff}{\ensuremath{+}0.121\xspace}

\newcommand{\NdthreePtADiffFull}{\ensuremath{+}0.121~[\ensuremath{+}0.110, \ensuremath{+}0.131]\xspace}

\newcommand{\NdthreePtAtenLevFull}{0.193~[0.181, 0.206]\xspace}

\newcommand{\NdthreePtAtenDiffFull}{\ensuremath{+}0.639~[\ensuremath{+}0.624, \ensuremath{+}0.655]\xspace}

\newcommand{\NdthreePtAthirtyLevFull}{0.711~[0.695, 0.727]\xspace}

\newcommand{\NdthreePtAthirtyDiffFull}{\ensuremath{+}0.122~[\ensuremath{+}0.111, \ensuremath{+}0.132]\xspace}
\newcommand{\NdthreeDlALev}{0.134\xspace}

\newcommand{\NdthreeDlALevFull}{0.134~[0.125, 0.144]\xspace}
\newcommand{\NdthreeDlADiff}{\ensuremath{+}0.698\xspace}

\newcommand{\NdthreeDlADiffFull}{\ensuremath{+}0.698~[\ensuremath{+}0.684, \ensuremath{+}0.713]\xspace}
\newcommand{\NdthreeHtfALev}{0.098\xspace}

\newcommand{\NdthreeHtfALevFull}{0.098~[0.090, 0.107]\xspace}
\newcommand{\NdthreeHtfADiff}{\ensuremath{+}0.734\xspace}

\newcommand{\NdthreeHtfADiffFull}{\ensuremath{+}0.734~[\ensuremath{+}0.720, \ensuremath{+}0.749]\xspace}
\newcommand{\NdthreeDlAtenLev}{\ensuremath{-}0.025\xspace}

\newcommand{\NdthreeDlAtenLevFull}{\ensuremath{-}0.025~[\ensuremath{-}0.030, \ensuremath{-}0.018]\xspace}
\newcommand{\NdthreeDlAtenDiff}{\ensuremath{+}0.668\xspace}

\newcommand{\NdthreeDlAtenDiffFull}{\ensuremath{+}0.668~[\ensuremath{+}0.652, \ensuremath{+}0.683]\xspace}
\newcommand{\NdthreeFadeAtwoLev}{0.782\xspace}

\newcommand{\NdthreeFadeAtwoLevFull}{0.782~[0.769, 0.795]\xspace}
\newcommand{\NdthreeFadeAtwoDiff}{\ensuremath{+}0.050\xspace}

\newcommand{\NdthreeFadeAtwoDiffFull}{\ensuremath{+}0.050~[\ensuremath{+}0.044, \ensuremath{+}0.056]\xspace}
\newcommand{\NdthreeFadeAtenLev}{0.085\xspace}

\newcommand{\NdthreeFadeAtenLevFull}{0.085~[0.079, 0.091]\xspace}
\newcommand{\NdthreeFadeAtenDiff}{\ensuremath{+}0.558\xspace}

\newcommand{\NdthreeFadeAtenDiffFull}{\ensuremath{+}0.558~[\ensuremath{+}0.544, \ensuremath{+}0.573]\xspace}
\newcommand{\NdthreeDlBLev}{\ensuremath{-}0.061\xspace}

\newcommand{\NdthreeDlBLevFull}{\ensuremath{-}0.061~[\ensuremath{-}0.065, \ensuremath{-}0.056]\xspace}
\newcommand{\NdthreeDlBDiff}{\ensuremath{+}0.704\xspace}

\newcommand{\NdthreeDlBDiffFull}{\ensuremath{+}0.704~[\ensuremath{+}0.688, \ensuremath{+}0.720]\xspace}
\newcommand{\NdthreePtBLev}{\ensuremath{-}0.096\xspace}

\newcommand{\NdthreePtBLevFull}{\ensuremath{-}0.096~[\ensuremath{-}0.100, \ensuremath{-}0.092]\xspace}
\newcommand{\NdthreePtBDiff}{\ensuremath{+}0.739\xspace}

\newcommand{\NdthreePtBDiffFull}{\ensuremath{+}0.739~[\ensuremath{+}0.723, \ensuremath{+}0.755]\xspace}
\newcommand{\NdthreeTfmLev}{0.605\xspace}

\newcommand{\NdthreeTfmLevFull}{0.605~[0.587, 0.624]\xspace}
\newcommand{\NdthreeTfmDiff}{\ensuremath{+}0.067\xspace}

\newcommand{\NdthreeTfmDiffFull}{\ensuremath{+}0.067~[\ensuremath{+}0.053, \ensuremath{+}0.081]\xspace}

\newcommand{\NdthreeTfmSixteenLevFull}{0.607~[0.589, 0.625]\xspace}

\newcommand{\NdthreeTfmSixteenDiffFull}{\ensuremath{+}0.065~[\ensuremath{+}0.053, \ensuremath{+}0.078]\xspace}

\newcommand{\NdthreeTfmThirtytwoLevFull}{0.612~[0.595, 0.630]\xspace}

\newcommand{\NdthreeTfmThirtytwoDiffFull}{\ensuremath{+}0.060~[\ensuremath{+}0.047, \ensuremath{+}0.074]\xspace}

\newcommand{\NdthreeTfmSixtyfourLevFull}{0.608~[0.590, 0.626]\xspace}

\newcommand{\NdthreeTfmSixtyfourDiffFull}{\ensuremath{+}0.064~[\ensuremath{+}0.051, \ensuremath{+}0.078]\xspace}
\newcommand{\NdthreePtRrLev}{0.416\xspace}

\newcommand{\NdthreePtRrLevFull}{0.416~[0.398, 0.433]\xspace}
\newcommand{\NdthreePtRrDiff}{\ensuremath{+}0.256\xspace}

\newcommand{\NdthreePtRrDiffFull}{\ensuremath{+}0.256~[\ensuremath{+}0.241, \ensuremath{+}0.272]\xspace}
\newcommand{\NdthreeDlRrLev}{0.264\xspace}

\newcommand{\NdthreeDlRrLevFull}{0.264~[0.250, 0.278]\xspace}
\newcommand{\NdthreeDlRrDiff}{\ensuremath{+}0.408\xspace}

\newcommand{\NdthreeDlRrDiffFull}{\ensuremath{+}0.408~[\ensuremath{+}0.392, \ensuremath{+}0.425]\xspace}

\newcommand{\NdthreeArTwoLevFull}{0.488~[0.470, 0.506]\xspace}

\newcommand{\NdthreeArTwoDiffFull}{\ensuremath{+}0.184~[\ensuremath{+}0.168, \ensuremath{+}0.201]\xspace}

\newcommand{\NdthreeArFourLevFull}{0.511~[0.493, 0.529]\xspace}

\newcommand{\NdthreeArFourDiffFull}{\ensuremath{+}0.161~[\ensuremath{+}0.145, \ensuremath{+}0.178]\xspace}
\newcommand{\NdthreeArEightLev}{0.522\xspace}

\newcommand{\NdthreeArEightLevFull}{0.522~[0.504, 0.540]\xspace}
\newcommand{\NdthreeArEightDiff}{\ensuremath{+}0.150\xspace}

\newcommand{\NdthreeArEightDiffFull}{\ensuremath{+}0.150~[\ensuremath{+}0.135, \ensuremath{+}0.167]\xspace}

\newcommand{\NdthreeArSixteenLevFull}{0.517~[0.496, 0.535]\xspace}

\newcommand{\NdthreeArSixteenDiffFull}{\ensuremath{+}0.156~[\ensuremath{+}0.139, \ensuremath{+}0.174]\xspace}

\newcommand{\NdthreeGruMseLevFull}{0.057~[0.045, 0.069]\xspace}

\newcommand{\NdthreeGruMseDiffFull}{\ensuremath{+}0.616~[\ensuremath{+}0.597, \ensuremath{+}0.635]\xspace}

\newcommand{\NdthreeGruLoneLevFull}{0.402~[0.387, 0.419]\xspace}

\newcommand{\NdthreeGruLoneDiffFull}{\ensuremath{+}0.270~[\ensuremath{+}0.255, \ensuremath{+}0.285]\xspace}

\newcommand{\NdthreeGruNllLevFull}{0.134~[0.120, 0.148]\xspace}

\newcommand{\NdthreeGruNllDiffFull}{\ensuremath{+}0.538~[\ensuremath{+}0.519, \ensuremath{+}0.559]\xspace}
\newcommand{\NdthreeGruHubLev}{0.538\xspace}

\newcommand{\NdthreeGruHubLevFull}{0.538~[0.520, 0.556]\xspace}
\newcommand{\NdthreeGruHubDiff}{\ensuremath{+}0.135\xspace}

\newcommand{\NdthreeGruHubDiffFull}{\ensuremath{+}0.135~[\ensuremath{+}0.121, \ensuremath{+}0.149]\xspace}

\newcommand{\NdthreeGbmLevFull}{0.573~[0.555, 0.591]\xspace}

\newcommand{\NdthreeGbmDiffFull}{\ensuremath{+}0.099~[\ensuremath{+}0.087, \ensuremath{+}0.113]\xspace}
\newcommand{\NdthreeGbmHubLev}{0.615\xspace}

\newcommand{\NdthreeGbmHubLevFull}{0.615~[0.598, 0.633]\xspace}
\newcommand{\NdthreeGbmHubDiff}{\ensuremath{+}0.057\xspace}

\newcommand{\NdthreeGbmHubDiffFull}{\ensuremath{+}0.057~[\ensuremath{+}0.046, \ensuremath{+}0.069]\xspace}
\newcommand{\NdthreeWwFone}{0.833\xspace}

\newcommand{\NdthreeWwFoneFull}{0.833~[0.820, 0.845]\xspace}
\newcommand{\NtrainPat}{3{,}383\xspace}
\newcommand{\NtrainSeg}{29{,}752\xspace}
\newcommand{\NtrainHours}{10{,}722\xspace}
\newcommand{\NppgTrainCases}{930\xspace}
\newcommand{\NabpTrainCases}{538\xspace}
\newcommand{\NwinDiscPreGate}{2{,}577\xspace}
\newcommand{\NwinLbxPreGate}{2{,}575\xspace}
\newcommand{\NwinValPreGate}{2{,}731\xspace}
\newcommand{\NpatVal}{888\xspace}

\newcommand{\NmoverWin}{1{,}641\xspace}
\newcommand{\NmoverPat}{574\xspace}
\newcommand{\NshdbWin}{5{,}879\xspace}
\newcommand{\NshdbRec}{98\xspace}
\newcommand{\NshdbAfWin}{1{,}216\xspace}
\newcommand{\NshdbAfRec}{64\xspace}
\newcommand{\NtauRmssdLow}{8.84\xspace}

\newcommand{\NtauRmssdHigh}{6.46\xspace}

\newcommand{\NtauNsrdb}{1.86\xspace}

\newcommand{\NtauLtafSinus}{3.56\xspace}

\newcommand{\NmseDirectMTDisc}{0.957\xspace}

\newcommand{\NmseDirectMTDiscCI}{[0.948, 0.966]\xspace}

\newcommand{\NmseDirectSplitDisc}{1.034\xspace}

\newcommand{\NmseDirectSplitDiscCI}{[1.024, 1.043]\xspace}

\newcommand{\NmseDirectMTSplitDisc}{1.034\xspace}

\newcommand{\NmseDirectMTSplitDiscCI}{[1.025, 1.044]\xspace}

\newcommand{\NmseWWSplitDisc}{1.762\xspace}

\newcommand{\NmseWWSplitDiscCI}{[1.749, 1.775]\xspace}

\newcommand{\NmseDirectLbx}{0.964\xspace}

\newcommand{\NmseDirectMTLbx}{0.962\xspace}

\newcommand{\NmseWWLbx}{1.696\xspace}

\newcommand{\NmseDirectSplitLbx}{1.038\xspace}

\newcommand{\NmseDirectMTSplitLbx}{1.038\xspace}

\newcommand{\NmseWWSplitLbx}{1.771\xspace}

\newcommand{\NifDelRobust}{\ensuremath{+}0.172\xspace}

\newcommand{\NifDelRobustFull}{\ensuremath{+}0.172~[\ensuremath{+}0.157, \ensuremath{+}0.187]\xspace}
\newcommand{\NifInsRobust}{\ensuremath{+}0.032\xspace}

\newcommand{\NifInsRobustFull}{\ensuremath{+}0.032~[\ensuremath{+}0.027, \ensuremath{+}0.039]\xspace}

\newcommand{\NheadRendererLbxFull}{\ensuremath{-}0.0045~[\ensuremath{-}0.0072, \ensuremath{-}0.0018]\xspace}
\newcommand{\NdirectReadRenderer}{\ensuremath{-}0.148\xspace}

\newcommand{\NdirectReadRendererFull}{\ensuremath{-}0.148~[\ensuremath{-}0.159, \ensuremath{-}0.136]\xspace}
\newcommand{\NguardedTplMinusWW}{\ensuremath{-}0.037\xspace}

\newcommand{\NguardedTplMinusWWFull}{\ensuremath{-}0.037~[\ensuremath{-}0.046, \ensuremath{-}0.028]\xspace}

\newcommand{\NdeviceFifty}{19.6\%\xspace}

\newcommand{\NdeviceFiftyMover}{\ensuremath{-}7.6\%\xspace}

\newcommand{\NdeviceHundred}{0.0\%\xspace}

\newcommand{\NsynTimingSDhigh}{104\xspace}
\newcommand{\NsynTimingSDthree}{34\xspace}
\newcommand{\NsynCommitGainLow}{\ensuremath{+}0.009\xspace}

\newcommand{\NsynCommitGainHigh}{\ensuremath{+}0.160\xspace}

\newcommand{\NsynExcessDirectB}{0.0029\xspace}
\newcommand{\NsynExcessCommitB}{0.0354\xspace}
\newcommand{\NsynAmpDirectB}{0.26\xspace}
\newcommand{\NsynAmpCommitB}{0.93\xspace}

\newcommand{\NhypOne}{6.89\xspace}

\newcommand{\NfirstBeatDisc}{19.8\%\xspace}

\newcommand{\NfirstBeatLbx}{15.8\%\xspace}

\newcommand{\NfirstBeatNonEct}{25.5\%\xspace}

\newcommand{\NdeviceFiftyLbx}{21.8\%\xspace}

\newcommand{\NrefNoArb}{9.5\xspace}
\newcommand{\NrefSuspKept}{2.8\xspace}
\newcommand{\NrefEdge}{0.045\xspace}

\newcommand{\NrefNoArbErrFull}{0.14\%~[0.00\%, 0.80\%]\xspace}

\newcommand{\NrefSuspErrFull}{0.91\%~[0.11\%, 3.26\%]\xspace}
\newcommand{\NrefCheckedAll}{7{,}126\xspace}
\newcommand{\NrefErrAll}{4\xspace}
\newcommand{\NexpMitF}{0.985\xspace}
\newcommand{\NexpLtafF}{0.970\xspace}

\newcommand{\NufHeadTemplate}{\ensuremath{-}0.0076\xspace}

\newcommand{\NufHeadTemplateFull}{\ensuremath{-}0.0076~[\ensuremath{-}0.0105, \ensuremath{-}0.0047]\xspace}

\newcommand{\NufSplitHeadRendererFull}{\ensuremath{-}0.0030~[\ensuremath{-}0.0063, \ensuremath{+}0.0001]\xspace}

\newcommand{\NufSplitDirectRendererFull}{\ensuremath{-}0.0946~[\ensuremath{-}0.1039, \ensuremath{-}0.0855]\xspace}

\newcommand{\NsynGainToneMoneFull}{\ensuremath{+}0.009~[\ensuremath{+}0.006, \ensuremath{+}0.011]\xspace}

\newcommand{\NsynMseToneMoneFull}{\ensuremath{+}0.0024~[\ensuremath{+}0.0020, \ensuremath{+}0.0028]\xspace}

\newcommand{\NsynGainToneMtwoFull}{\ensuremath{+}0.034~[\ensuremath{+}0.028, \ensuremath{+}0.040]\xspace}

\newcommand{\NsynMseToneMtwoFull}{\ensuremath{+}0.0006~[\ensuremath{+}0.0001, \ensuremath{+}0.0011]\xspace}

\newcommand{\NsynGainToneMthreeFull}{\ensuremath{+}0.014~[\ensuremath{+}0.012, \ensuremath{+}0.017]\xspace}

\newcommand{\NsynMseToneMthreeFull}{\ensuremath{+}0.0020~[\ensuremath{+}0.0016, \ensuremath{+}0.0025]\xspace}

\newcommand{\NsynGainToneMfourFull}{\ensuremath{+}0.020~[\ensuremath{+}0.018, \ensuremath{+}0.023]\xspace}

\newcommand{\NsynMseToneMfourFull}{\ensuremath{+}0.0015~[\ensuremath{+}0.0010, \ensuremath{+}0.0019]\xspace}

\newcommand{\NsynGainTtwoMoneFull}{\ensuremath{+}0.139~[\ensuremath{+}0.128, \ensuremath{+}0.150]\xspace}

\newcommand{\NsynMseTtwoMoneFull}{\ensuremath{+}0.0098~[\ensuremath{+}0.0091, \ensuremath{+}0.0105]\xspace}

\newcommand{\NsynGainTtwoMtwoFull}{\ensuremath{+}0.132~[\ensuremath{+}0.122, \ensuremath{+}0.142]\xspace}

\newcommand{\NsynMseTtwoMtwoFull}{\ensuremath{+}0.0089~[\ensuremath{+}0.0081, \ensuremath{+}0.0098]\xspace}

\newcommand{\NsynGainTtwoMthreeFull}{\ensuremath{+}0.092~[\ensuremath{+}0.084, \ensuremath{+}0.100]\xspace}

\newcommand{\NsynMseTtwoMthreeFull}{\ensuremath{+}0.0080~[\ensuremath{+}0.0073, \ensuremath{+}0.0087]\xspace}

\newcommand{\NsynGainTtwoMfourFull}{\ensuremath{+}0.134~[\ensuremath{+}0.124, \ensuremath{+}0.142]\xspace}

\newcommand{\NsynMseTtwoMfourFull}{\ensuremath{+}0.0092~[\ensuremath{+}0.0084, \ensuremath{+}0.0101]\xspace}

\newcommand{\NsynGainTthreeMoneFull}{\ensuremath{+}0.168~[\ensuremath{+}0.153, \ensuremath{+}0.184]\xspace}

\newcommand{\NsynMseTthreeMoneFull}{\ensuremath{+}0.0233~[\ensuremath{+}0.0224, \ensuremath{+}0.0242]\xspace}

\newcommand{\NsynGainTthreeMtwoFull}{\ensuremath{+}0.193~[\ensuremath{+}0.177, \ensuremath{+}0.208]\xspace}

\newcommand{\NsynMseTthreeMtwoFull}{\ensuremath{+}0.0237~[\ensuremath{+}0.0228, \ensuremath{+}0.0246]\xspace}

\newcommand{\NsynGainTthreeMthreeFull}{\ensuremath{+}0.180~[\ensuremath{+}0.164, \ensuremath{+}0.194]\xspace}

\newcommand{\NsynMseTthreeMthreeFull}{\ensuremath{+}0.0221~[\ensuremath{+}0.0213, \ensuremath{+}0.0229]\xspace}

\newcommand{\NsynGainTthreeMfourFull}{\ensuremath{+}0.166~[\ensuremath{+}0.152, \ensuremath{+}0.180]\xspace}

\newcommand{\NsynMseTthreeMfourFull}{\ensuremath{+}0.0194~[\ensuremath{+}0.0187, \ensuremath{+}0.0201]\xspace}

\newcommand{\NsynGainTfourMoneFull}{\ensuremath{+}0.160~[\ensuremath{+}0.148, \ensuremath{+}0.173]\xspace}

\newcommand{\NsynMseTfourMoneFull}{\ensuremath{+}0.0322~[\ensuremath{+}0.0312, \ensuremath{+}0.0332]\xspace}

\newcommand{\NsynGainTfourMtwoFull}{\ensuremath{+}0.172~[\ensuremath{+}0.160, \ensuremath{+}0.185]\xspace}

\newcommand{\NsynMseTfourMtwoFull}{\ensuremath{+}0.0309~[\ensuremath{+}0.0301, \ensuremath{+}0.0318]\xspace}

\newcommand{\NsynGainTfourMthreeFull}{\ensuremath{+}0.144~[\ensuremath{+}0.132, \ensuremath{+}0.155]\xspace}

\newcommand{\NsynMseTfourMthreeFull}{\ensuremath{+}0.0315~[\ensuremath{+}0.0307, \ensuremath{+}0.0323]\xspace}

\newcommand{\NsynGainTfourMfourFull}{\ensuremath{+}0.150~[\ensuremath{+}0.137, \ensuremath{+}0.161]\xspace}

\newcommand{\NsynMseTfourMfourFull}{\ensuremath{+}0.0314~[\ensuremath{+}0.0307, \ensuremath{+}0.0323]\xspace}
\newcommand{\NppgRefResidSd}{20\xspace}
\newcommand{\NppgRefResidGt}{3.5\%\xspace}
\newcommand{\NppgRefCoverage}{94\%\xspace}
\newcommand{\NppgRefDelayQlo}{694\xspace}
\newcommand{\NppgRefDelayQhi}{734\xspace}
\newcommand{\NppgRefPatGtTen}{2.8\%\xspace}
\newcommand{\NabpRefResidSd}{14\xspace}
\newcommand{\NabpRefResidGt}{1.4\%\xspace}
\newcommand{\NabpRefCoverage}{92\%\xspace}
\newcommand{\NabpRefDelayQlo}{196\xspace}
\newcommand{\NabpRefDelayQhi}{234\xspace}
\newcommand{\NabpRefPatGtTen}{1.6\%\xspace}

\newcommand{\NgateHrExclDisc}{14.8\%\xspace}
\newcommand{\NgateHrInclDisc}{3.7\%\xspace}
\newcommand{\NgateRhyExcludedWin}{310\xspace}
\newcommand{\NgateRhyExcludedAf}{142\xspace}
\newcommand{\NgateRhyExcludedAvb}{32\xspace}
\newcommand{\NgateRhyIncludedWin}{6{,}963\xspace}
\newcommand{\NgateRhyIncludedAf}{1{,}611\xspace}
\newcommand{\NgateRhyIncludedAvb}{27\xspace}
\newcommand{\NshapleyTimingZeroTwoMv}{0.940\xspace}

\newcommand{\NshapleyTimingZeroTwoMvFull}{0.940~[0.924, 0.953]\xspace}
\newcommand{\NmoverWinPreGate}{1{,}694\xspace}
\newcommand{\NmoverPatPreGate}{577\xspace}

\newcommand{\NsynSoftGainToneMoneFull}{\ensuremath{+}0.002~[\ensuremath{+}0.000, \ensuremath{+}0.005]\xspace}

\newcommand{\NsynSoftMseToneMoneFull}{\ensuremath{+}0.0008~[\ensuremath{+}0.0005, \ensuremath{+}0.0011]\xspace}

\newcommand{\NsynSoftGainToneMtwoFull}{\ensuremath{+}0.027~[\ensuremath{+}0.022, \ensuremath{+}0.034]\xspace}

\newcommand{\NsynSoftMseToneMtwoFull}{\ensuremath{-}0.0015~[\ensuremath{-}0.0020, \ensuremath{-}0.0010]\xspace}

\newcommand{\NsynSoftGainToneMthreeFull}{\ensuremath{+}0.012~[\ensuremath{+}0.010, \ensuremath{+}0.015]\xspace}

\newcommand{\NsynSoftMseToneMthreeFull}{\ensuremath{-}0.0001~[\ensuremath{-}0.0005, \ensuremath{+}0.0003]\xspace}

\newcommand{\NsynSoftGainToneMfourFull}{\ensuremath{+}0.020~[\ensuremath{+}0.018, \ensuremath{+}0.023]\xspace}

\newcommand{\NsynSoftMseToneMfourFull}{\ensuremath{-}0.0006~[\ensuremath{-}0.0010, \ensuremath{-}0.0002]\xspace}

\newcommand{\NsynSoftGainTtwoMoneFull}{\ensuremath{+}0.045~[\ensuremath{+}0.035, \ensuremath{+}0.055]\xspace}

\newcommand{\NsynSoftMseTtwoMoneFull}{\ensuremath{+}0.0023~[\ensuremath{+}0.0020, \ensuremath{+}0.0026]\xspace}

\newcommand{\NsynSoftGainTtwoMtwoFull}{\ensuremath{+}0.072~[\ensuremath{+}0.062, \ensuremath{+}0.082]\xspace}

\newcommand{\NsynSoftMseTtwoMtwoFull}{\ensuremath{+}0.0018~[\ensuremath{+}0.0013, \ensuremath{+}0.0022]\xspace}

\newcommand{\NsynSoftGainTtwoMthreeFull}{\ensuremath{+}0.046~[\ensuremath{+}0.039, \ensuremath{+}0.052]\xspace}

\newcommand{\NsynSoftMseTtwoMthreeFull}{\ensuremath{+}0.0007~[\ensuremath{+}0.0003, \ensuremath{+}0.0010]\xspace}

\newcommand{\NsynSoftGainTtwoMfourFull}{\ensuremath{+}0.103~[\ensuremath{+}0.094, \ensuremath{+}0.111]\xspace}

\newcommand{\NsynSoftMseTtwoMfourFull}{\ensuremath{+}0.0038~[\ensuremath{+}0.0032, \ensuremath{+}0.0043]\xspace}

\newcommand{\NsynSoftGainTthreeMoneFull}{\ensuremath{+}0.039~[\ensuremath{+}0.027, \ensuremath{+}0.050]\xspace}

\newcommand{\NsynSoftMseTthreeMoneFull}{\ensuremath{+}0.0059~[\ensuremath{+}0.0055, \ensuremath{+}0.0063]\xspace}

\newcommand{\NsynSoftGainTthreeMtwoFull}{\ensuremath{+}0.099~[\ensuremath{+}0.086, \ensuremath{+}0.113]\xspace}

\newcommand{\NsynSoftMseTthreeMtwoFull}{\ensuremath{+}0.0066~[\ensuremath{+}0.0063, \ensuremath{+}0.0070]\xspace}

\newcommand{\NsynSoftGainTthreeMthreeFull}{\ensuremath{+}0.022~[\ensuremath{+}0.010, \ensuremath{+}0.033]\xspace}

\newcommand{\NsynSoftMseTthreeMthreeFull}{\ensuremath{+}0.0042~[\ensuremath{+}0.0038, \ensuremath{+}0.0046]\xspace}

\newcommand{\NsynSoftGainTthreeMfourFull}{\ensuremath{+}0.077~[\ensuremath{+}0.065, \ensuremath{+}0.088]\xspace}

\newcommand{\NsynSoftMseTthreeMfourFull}{\ensuremath{+}0.0059~[\ensuremath{+}0.0055, \ensuremath{+}0.0064]\xspace}

\newcommand{\NsynSoftGainTfourMoneFull}{\ensuremath{+}0.082~[\ensuremath{+}0.073, \ensuremath{+}0.091]\xspace}

\newcommand{\NsynSoftMseTfourMoneFull}{\ensuremath{+}0.0082~[\ensuremath{+}0.0079, \ensuremath{+}0.0084]\xspace}

\newcommand{\NsynSoftGainTfourMtwoFull}{\ensuremath{+}0.066~[\ensuremath{+}0.058, \ensuremath{+}0.074]\xspace}

\newcommand{\NsynSoftMseTfourMtwoFull}{\ensuremath{+}0.0136~[\ensuremath{+}0.0130, \ensuremath{+}0.0142]\xspace}

\newcommand{\NsynSoftGainTfourMthreeFull}{\ensuremath{+}0.088~[\ensuremath{+}0.078, \ensuremath{+}0.098]\xspace}

\newcommand{\NsynSoftMseTfourMthreeFull}{\ensuremath{+}0.0185~[\ensuremath{+}0.0176, \ensuremath{+}0.0194]\xspace}

\newcommand{\NsynSoftGainTfourMfourFull}{\ensuremath{+}0.090~[\ensuremath{+}0.082, \ensuremath{+}0.099]\xspace}

\newcommand{\NsynSoftMseTfourMfourFull}{\ensuremath{+}0.0179~[\ensuremath{+}0.0169, \ensuremath{+}0.0189]\xspace}
\newcommand{\NfollowDiscCMTExpr}{72.04\%\xspace}

\newcommand{\NfollowDiscCWWExpr}{95.62\%\xspace}

\ifinternal\else\fi

\begin{document}
\begin{frontmatter}

\title{When2What: A Timing-Conditioned Framework for Long-Horizon Forecasting of Continuous Cardiovascular Waveforms\tnoteref{orcid}}

\author[1]{Chaoqi Zhang}

\author[2]{Shujie Yan\fnref{fn1}}

\author[2]{Zhao Zhang\fnref{fn2}}

\author[2]{Yanhua Sun\fnref{fn3}}

\author[3]{Yu Wang}

\author[4]{Li Li}

\author[2]{Lingzhong Meng\fnref{fn4}}

\author[1]{Xiaoyi Chen\corref{cor1}\fnref{fn5}}

\author[1]{Haixu Tang}

\affiliation[1]{organization={Department of Computer Science, Luddy School of Informatics, Computing, and Engineering, Indiana University Bloomington},
                city={Bloomington},
                state={IN},
                country={USA}}

\affiliation[2]{organization={Department of Anesthesia, Indiana University School of Medicine},
                city={Indianapolis},
                state={IN},
                country={USA}}

\affiliation[3]{organization={Department of Intelligent Systems Engineering, Luddy School of Informatics, Computing, and Engineering, Indiana University Bloomington},
                city={Bloomington},
                state={IN},
                country={USA}}

\affiliation[4]{organization={Department of Critical Care Medicine, Xiangya Hospital, Central South University},
                city={Changsha},
                state={Hunan},
                country={China}}

\cortext[cor1]{Corresponding author: Xiaoyi Chen. E-mail: \href{mailto:chxiaoyi@iu.edu}{chxiaoyi@iu.edu}. Additional contact emails: \href{mailto:cz42@iu.edu}{cz42@iu.edu} (Chaoqi Zhang); \href{mailto:hatang@iu.edu}{hatang@iu.edu} (Haixu Tang).}

\fntext[fn1]{Home institution during the visiting appointment: Department of Cardiopulmonary Bypass, Fuwai Hospital, National Centre for Cardiovascular Diseases, Chinese Academy of Medical Sciences \& Peking Union Medical College, National Clinical Research Centre for Cardiovascular Diseases, Beijing, China.}
\fntext[fn2]{Home institution during the visiting appointment: Department of Anesthesiology, Tianjin Chest Hospital, Tianjin University, Tianjin, China.}
\fntext[fn3]{Home institution during the visiting appointment: Department of Anesthesiology, Nanjing Drum Tower Hospital, Affiliated Hospital of Medical School, Nanjing University, Nanjing, Jiangsu, China.}
\fntext[fn4]{Present address: Department of Anesthesiology and Pain Medicine, University of California Davis School of Medicine, Sacramento, California, USA.}
\fntext[fn5]{Present address: Microsoft, Seattle, WA, USA.}

\tnotetext[orcid]{ORCID: Chaoqi Zhang \mbox{0000-0001-5113-7160}; Shujie Yan \mbox{0000-0002-6211-3145}; Zhao Zhang \mbox{0000-0001-6314-1116}; Yanhua Sun \mbox{0009-0006-6086-6684}; Yu Wang \mbox{0000-0003-0361-1563}; Li Li \mbox{0009-0001-0301-7575}; Lingzhong Meng \mbox{0000-0002-5168-5084}; Xiaoyi Chen \mbox{0000-0002-6224-791X}; Haixu Tang \mbox{0000-0001-8963-8155}.}

\begin{abstract}
Long-horizon forecasting of quasi-periodic cardiovascular waveforms, such as the electrocardiogram (ECG), arterial blood pressure (ABP), and photoplethysmogram (PPG), requires predicting both \emph{when} future cardiac events will occur (beat timing) and \emph{what} the local structural pattern looks like around each event (beat morphology). However, standard sequence forecasters trained with pointwise mean squared error (MSE) average across uncertain future heartbeat phases, flattening sharp R peaks and pulse upstrokes into featureless baselines. 
Our analysis of intraoperative ECG reveals a clear temporal asymmetry: beat-time predictability from history decays within seconds due to stochastic heart-rate variability, whereas local beat morphology aligned at true beat times remains stable over several minutes. 
Guided by this insight, we introduce When2What, a timing-conditioned framework comprising three stages: Predict, Commit, and Generate. Given a 120-s history, When2What predicts a dense 60-s timing map, selects and sharpens discrete beat events into unit-height Gaussian impulses, and conditions a neural waveform generator on both these committed events and the observed history. 
Across intraoperative ECG, When2What substantially outperforms existing time-series and physiological forecasters, achieving a near-term waveform event skill of 0.644 and reproducing 95.62\% of its predicted heartbeats in the generated waveform where direct long-horizon forecasters collapse into flat baselines. 
Furthermore, When2What transfers zero-shot to external hospital and atrial fibrillation cohorts, extends effectively to ABP and PPG pulse waveforms, and is supported by ablations and synthetic analyses that establish explicit timing conditioning as an effective paradigm for quasi-periodic long-horizon waveform forecasting.
\end{abstract}

\begin{keyword}
Time series forecasting \sep Physiological signals \sep Electrocardiogram \sep Event timing \sep Conditional generation \sep Photoplethysmography \sep Arterial blood pressure
\end{keyword}

\end{frontmatter}
\hypersetup{pdfauthor={Chaoqi Zhang, Shujie Yan, Zhao Zhang, Yanhua Sun, Yu Wang, Li Li, Lingzhong Meng, Xiaoyi Chen, Haixu Tang}}

\section{Introduction}\label{sec:intro}

Continuous cardiovascular waveforms, including the electrocardiogram (ECG), arterial blood pressure (ABP), and photoplethysmogram (PPG), are representative quasi-periodic, event-driven time series that are fundamental to physiological pattern analysis and critical care monitoring~\cite{ASA2025Monitoring,lee2022}.
While predictive monitoring often compresses these signals into a single risk score for adverse events such as hypotension~\cite{hatib2018,Jo2022WaveformHypotension}, full-waveform forecasting aims to predict the continuous future evolution of the physiological signal itself.
Because cardiovascular signals are driven by recurring heartbeats, a meaningful forecast must capture two complementary properties of each cardiac cycle: \emph{beat timing}, which marks when individual events occur (ventricular R peaks in ECG or maximal-upslope pulse anchors in ABP and PPG), and \emph{beat morphology}, which describes the local waveform shape surrounding each event.
Preserving the structural integrity of a forecast therefore requires more than matching average signal levels; the predicted waveform must produce identifiable beats at the right times with realistic shapes.

A standard approach to waveform forecasting maps the observed history directly to future samples using general-purpose time-series models or ECG-specific networks trained with pointwise objectives such as mean squared error (MSE)~\cite{nie2023,zacarias2024,li2026}.
However, existing ECG forecasters typically operate on short windows of one to a few seconds, and direct samplewise regression breaks down over longer multi-beat horizons (such as 10 to 60~s) when future beat timing is uncertain. 
The optimal point forecast is the conditional mean of plausible future trajectories~\cite{gneiting2011}, which means that when heartbeats shift slightly in time across these trajectories, averaging them together inevitably attenuates or broadens sharp waveform features such as the R peak and flattens the forecast waveform. 
To better preserve waveform structure, prior studies have modified the training objective. For example, FADE~\cite{ruizbarroso2025} weights different amplitude bands separately in its loss function.
Other methods organize forecasts around recurring patterns or events: BeatLex~\cite{hooi2017} forecasts sequences of learned waveform patterns and concatenates them into a continuous signal, RDiffGAN-ECG~\cite{lin2026} synthesizes ECG conditioned on supplied R-peak locations and diagnostic categories, and DiffECG~\cite{neifar2024diffecg} forecasts only a single next heartbeat from the preceding beat using segments aligned around annotated R peaks.
Yet, forecasting an uninterrupted, unsegmented multi-beat recording poses a distinct challenge: future beat locations are neither supplied in advance nor pre-aligned, which requires a forecaster to estimate multiple future beat times from history and express those beats in a continuous output waveform. 

Our analysis of intraoperative ECG motivates decoupling these two steps by revealing a marked temporal asymmetry between \emph{beat timing} and \emph{beat morphology}.
When estimated from historical waveforms alone, beat-time forecast skill decays within a few seconds as beat-to-beat interval variations accumulate.
By contrast, when future beats are aligned at their reference times, their local waveform shape remains highly consistent with a historical beat template over intervals of several minutes.
This contrast between rapidly growing uncertainty in \emph{when} future beats will occur, and multi-minute stability in \emph{what} each beat looks like, motivates representing predicted event times explicitly while drawing on the observed history to construct the local waveform shape. 
Crucially, this raises the central question of our study: \emph{in long-horizon waveform forecasting, how should a model connect uncertain future beat timing (when) to continuous waveform generation (what) so that identifiable heartbeats are preserved in the output?}

To answer this question, we introduce \textbf{When2What}, a timing-conditioned framework for long-horizon cardiovascular waveform forecasting.
Given a 120-s waveform history at 500~Hz, When2What decomposes forecasting into three sequential stages: \textbf{Predict} ($P$), \textbf{Commit} ($C$), and \textbf{Generate} ($G$). %
First, the timing predictor $P$ estimates a dense timing probability map over the 60-s forecast horizon. %
Because beat-time uncertainty causes raw timing predictions to broaden and decay in amplitude as lead time increases, feeding the raw map directly into a generator would still result in flattened waveforms.
The commit step $C$ overcomes this problem by selecting discrete beat events and re-encoding them into a sharpened timing map of unit-height Gaussian impulses.
Finally, the waveform generator $G$ conditions on both this sharpened timing map and the encoded history at coarse token and fine sample scales to synthesize a continuous 60-s waveform. %
We evaluate When2What against existing time-series and physiological forecasting baselines across multiple cohorts and pulse signals, and conduct systematic ablations against unconditioned regression variants (Direct and Direct-MT), unsharpened timing maps, and static historical beat templates.

Extensive experiments on intraoperative ECG, ABP, and PPG waveforms demonstrate the effectiveness of When2What.
Across all evaluated settings, When2What consistently outperforms existing time-series and ECG forecasters, including Informer~\cite{zhou2021}, PatchTST~\cite{nie2023}, DLinear~\cite{zeng2023}, HTF~\cite{zhang2026htf}, BeatLex~\cite{hooi2017}, FADE~\cite{ruizbarroso2025}, and the foundation model TimesFM~\cite{das2024}, achieving a 0--10-s waveform event skill of \NwaveWWDisc{} where direct long-horizon forecasters collapse into flat baselines. %
Our ablation study reveals why this improvement occurs: although a multi-task network with auxiliary timing supervision (Direct-MT) predicts future beat times accurately in its timing head, its unconditioned waveform generator reproduces only \NfollowDiscCMTExpr{} of those correctly predicted beats in the output waveform, whereas When2What reproduces \NfollowDiscCWWExpr{} and improves waveform event skill by \NdcondOrigDisc{}.
Moreover, our learned generator outperforms static historical beat templates across event accuracy, 60-s waveform MSE, and resilience to event-input perturbations, while controlled synthetic experiments confirm that preserving sharp heartbeats under timing uncertainty mathematically requires trading off pointwise squared error. %

Our contributions are fourfold:
\begin{itemize}
    \item \textbf{Timing and morphology characterization:} We show that beat-time forecast accuracy from history decays within seconds in intraoperative ECG, whereas beat morphology aligned at reference times remains stable over several minutes, motivating the decoupling of beat-time estimation and waveform construction.
    \item \textbf{Timing-conditioned long-horizon waveform forecasting:} We develop When2What, a predict, commit, and generate pipeline that sharpens predicted future beat locations into explicit conditioning inputs to synthesize 60-s, multi-beat continuous physiological waveforms without the peak flattening of direct regression. %
    \item \textbf{Effectiveness and cross-signal generalization:} We demonstrate that When2What outperforms existing time-series and ECG forecasters and generalizes zero-shot to external hospital and arrhythmia cohorts (MOVER and SHDB-AF) as well as ABP and PPG pulse waveforms.
    \item \textbf{Pipeline ablations and theoretical trade-off analysis:} Through systematic ablations and controlled synthetic experiments, we show that both discrete event sharpening and learned dynamic waveform generation are essential, and establish why preserving sharp cardiac events under timing uncertainty inevitably increases pointwise MSE.
\end{itemize}

\section{Related work}\label{sec:related}

\subsection{Continuous waveform forecasting and cardiac-event prediction}\label{sec:rel-forecasting}

Existing physiological forecasters typically operate over short windows of a few hundred samples or target discrete event times rather than minute-scale continuous waveforms.
DiffECG~\cite{neifar2024diffecg} forecasts only the next single heartbeat from the preceding beat on pre-segmented, R-peak-aligned windows, whereas Yaldiz et al.~\cite{yaldiz2026} predict future cardiac-event timings from detected features without generating a waveform.
Among continuous waveform predictors, ECG-specific models such as LSTM forecasters~\cite{zacarias2024}, Clairvoyant-Net~\cite{li2026}, and FADE~\cite{ruizbarroso2025} focus on short horizons of one to a few seconds, while general long-sequence architectures such as Informer~\cite{zhou2021}, PatchTST~\cite{nie2023}, and DLinear~\cite{zeng2023} regress future samples directly with pointwise squared error.
When extended to a 60-s unsegmented horizon at 500~Hz (30,000 samples), direct pointwise regression averages across slightly shifted future cardiac cycles and rapidly collapses toward a flat baseline; probabilistic models such as DeepAR~\cite{salinas2020} and TimeGrad~\cite{rasul2021} instead sample trajectories from learned distributions.
To generate an uninterrupted 60-s forecast that preserves clear heartbeats across the near-term multi-beat horizon (0--10~s) while characterizing longer-term decay, our approach explicitly decouples multi-beat timing estimation from continuous waveform construction.

\subsection{Structured waveform construction and event conditioning}\label{sec:rel-conditioning}

Conditioning signal generation on intermediate temporal structure has precedents in speech and video modeling (FastSpeech~2~\cite{ren2021}, FLODCAST~\cite{ciamarra2024}) as well as physiological synthesis.
McSharry et al.~\cite{mcsharry2003} generate synthetic ECG from prescribed RR dynamics and morphology parameters, VABAM~\cite{kim2026} decouples waveform shape and amplitude via cascaded filtering, and RDiffGAN-ECG~\cite{lin2026} synthesizes ECG conditioned on supplied R-peak locations and diagnostic classes.
For continuous forecasting, BeatLex~\cite{hooi2017} predicts sequences of learned waveform patterns with a variable-order Markov model and concatenates them into a continuation.
Yet when future beat times must be estimated from history, two questions remain open: whether supplying predicted beat times as explicit generator inputs improves beat expression beyond auxiliary timing supervision, and whether waveform construction requires a learned neural generator rather than a simple historical beat template.
We address both questions by developing When2What and comparing it against a matched multi-task control (Direct-MT) and non-learned template baselines.

\subsection{Structure-aware objectives and event-level evaluation}\label{sec:rel-objectives}

Rather than changing how temporal structure enters the generator, structure-aware training objectives modify how prediction errors are penalized: FADE's Split-MSE~\cite{ruizbarroso2025} weights high-amplitude peak regions separately from baseline segments, while DILATE~\cite{leguen2019} and Soft-DTW~\cite{cuturi2017} penalize shape and temporal distortion using differentiable alignment.
Because loss design and explicit event conditioning are complementary choices, we evaluate When2What and its matched controls under both standard derivative-augmented MSE and Split-MSE.
Moreover, because low samplewise MSE does not guarantee detectable heartbeats in the output waveform, we pair pointwise error metrics with forecast-verification skill scores~\cite{murphy1988,jolliffe2012} using established beat detectors~\cite{pantompkins1985,makowski2021} to separately assess beat-time prediction, waveform beat expression, and samplewise fidelity.

\section{\method}\label{sec:setup}\label{sec:framework}

\subsection{Task formulation and empirical motivation}\label{sec:task}\label{sec:asymmetry}

\paragraph{Forecasting task and cardiac beat events}
We formulate cardiovascular waveform forecasting as a continuous, single-signal prediction problem.
Given an observed history $\mathbf{x} \in \mathbb{R}^{H}$ of $120$~s sampled at $500$~Hz ($H = 60{,}000$ samples), a forecaster generates a single continuous waveform trajectory $\hat{\mathbf{y}} \in \mathbb{R}^{L}$ over the subsequent $60$-s forecast horizon ($L = 30{,}000$ samples), rather than a scalar clinical risk score~\cite{hatib2018} or a probability distribution over trajectories. %
During inference, the model receives only historical samples $\mathbf{x}$ preceding the forecast origin, while future waveform samples $\mathbf{y}$ and reference beat times $E = \{e_1, \dots, e_K\}$ serve exclusively as training targets or evaluation references.
Because cardiovascular signals are organized around recurring cardiac cycles, a useful forecast must preserve both samplewise waveform fidelity and individual beat events.
Each beat event $e_k \in E$ is defined in a signal-specific manner: it marks the ventricular R peak in ECG and the maximal-upslope pulse anchor in ABP and PPG, which follows the ECG R peak after an observed physiological delay (Fig.~\ref{fig:anchors}).

\paragraph{Measuring beat-time predictability}
To evaluate how beat-time accuracy changes with lead time $t$ (the time elapsed from the forecast origin, from $0$ to $60$~s), predicted and reference events are paired one-to-one within a tolerance of $\pm$50~ms and grouped into lead-time bins (Supplementary Section~S1). %
Because higher heart rates increase the probability that uninformative predictions fall within $\pm$50~ms of true beats by chance, we normalize the raw $F_1$ score in each bin against a chance baseline $F_1^{\mathrm{c}}$. %
Specifically, $F_1^{\mathrm{c}}$ is the expected $F_1$ of a constant-rate baseline (a periodic pulse train at the window's true mean RR interval with a uniformly random starting phase):
\begin{equation}\label{eq:skill}
S = \frac{F_1 - F_1^{\mathrm{c}}}{1 - F_1^{\mathrm{c}}},
\end{equation}
where $F_1$ and $F_1^{\mathrm{c}}$ denote cohort averages across all windows in a given lead-time bin (computed after averaging across model seeds within each window; Supplementary Section~S1).
An event skill of $S = 1$ indicates perfect beat placement, $S = 0$ matches random guessing, and $S < 0$ indicates performance worse than chance.
To quantify how rapidly beat-time predictability decays with lead time, we fit an exponential curve, $a\,e^{-t/\tau}+c$, to the event skill $S$ across 2-s lead-time bins over the 60-s horizon and report the decay time constant $\tau$ (Supplementary Section~S1). %

\begin{figure}[!htbp]
\centering
\WWfig{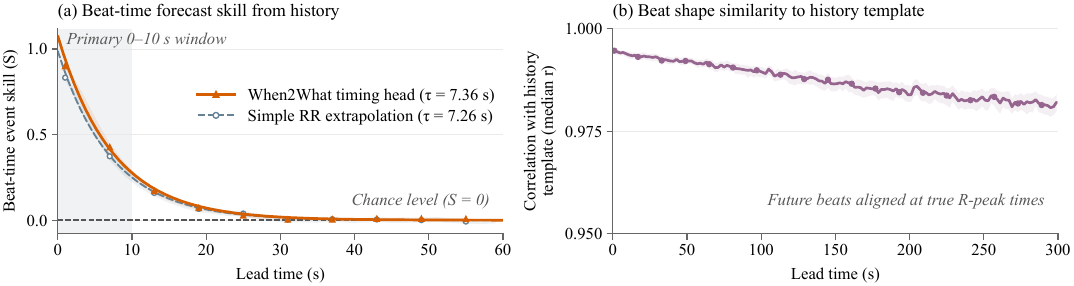}{\linewidth}
\caption{Temporal asymmetry between beat timing and beat morphology on intraoperative VitalDB ECG.
(a)~Beat-time event skill $S$ across 2-s lead-time bins for a simple RR extrapolation and the learned \method timing head, shown with exponential decay fits ($a\,e^{-t/\tau}+c$) and 95\% patient-cluster bootstrap bands. Both predictors decay to the chance level ($S = 0$) within seconds; shading marks the primary 0--10~s evaluation window.
(b)~Median Pearson correlation ($r$) between each patient's historical median beat template and future beats aligned at their true R-peak times over a 5-min horizon.}\label{fig:asymmetry} %
\end{figure}

\paragraph{Motivating observation: predicting beat timing is hard, but preserving beat morphology is easy}
Evaluating intraoperative VitalDB ECG~\cite{lee2022} reveals a temporal asymmetry between beat timing and beat morphology (Fig.~\ref{fig:asymmetry}).
When predicting future beat times from history alone, event skill $S$ drops rapidly within the first 10~s and reaches the random-chance level ($S = 0$) after roughly 25~s. %
For the evaluated predictors, this seconds-scale decay holds across both a simple RR extrapolation that projects the median of the last five corrected beat intervals ($\tau = \NtauCNDisc{}$~s) and our learned \method timing head ($\tau = \NtauWWDisc{}$~s; Fig.~\ref{fig:asymmetry}a), while alternative autoregressive, recurrent, Transformer, and foundation-model RR forecasters achieve even lower near-term skill (Supplementary Section~S4, Table~S8). %
Because the time of the $k$-th future heartbeat is the cumulative sum of $k$ successive beat intervals, even small beat-to-beat variations accumulate across cardiac cycles and quickly push timing uncertainty beyond the $\pm$50-ms alignment window. %
By contrast, when future beats are aligned at their true R-peak times, their local waveform shape remains highly consistent with a historical beat template (the pointwise median of the last 40 complete history beats) over several minutes: the median correlation is $\NmorphRStart{}$ in the first 2~s and remains $\NmorphREnd{}$ four to five minutes ahead (Fig.~\ref{fig:asymmetry}b). %

Although this reference-aligned correlation measures morphological stability rather than forecast accuracy, comparing Fig.~\ref{fig:asymmetry}a and Fig.~\ref{fig:asymmetry}b provides a clear design principle.
Because uncertainty in \emph{when} future beats will occur grows rapidly within seconds while \emph{what} each beat looks like remains stable over minutes, a direct regressor that predicts waveform samples directly is forced to average over phase-shifted future heartbeats, smoothing out sharp R peaks and pulse upstrokes.
To avoid this phase averaging, a forecaster should decouple beat-time estimation from waveform construction: it should first commit to explicit future beat locations and then construct the local waveform shape around those locations using the observed history.
This decay profile also motivates using the first 10~s as our primary window for near-term beat accuracy while retaining the full 60-s horizon to evaluate skill decay and samplewise error. %

\subsection{Overview}\label{sec:method}

\begin{figure}[!htbp]
\centering
\WWfig{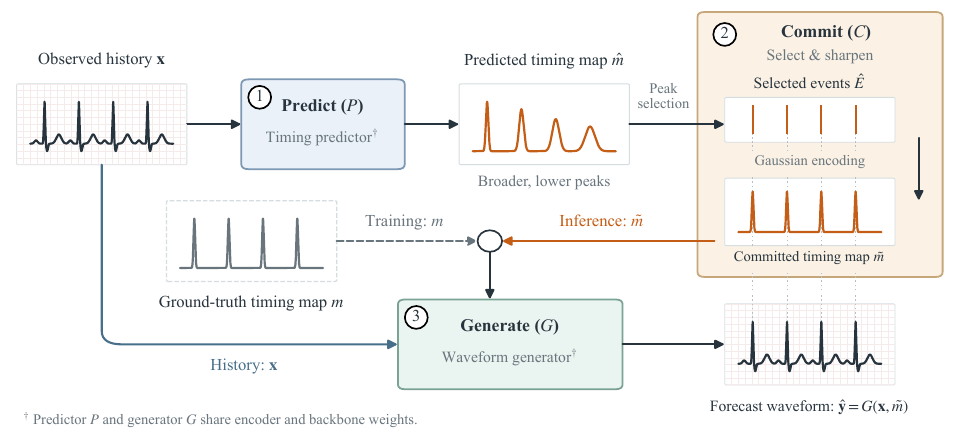}{\linewidth}
\caption{Overview of the \method framework.
(1)~In \textbf{Predict} ($P$), the timing predictor estimates a continuous timing map $\hat{m}$ from the observed history $\mathbf{x}$, whose peaks become broader and lower in amplitude as lead time increases.
(2)~In \textbf{Commit} ($C$), discrete beat events $\hat{E}$ are selected from $\hat{m}$ and re-encoded into a committed timing map $\tilde{m}$ of unit-height Gaussian bumps.
(3)~In \textbf{Generate} ($G$), the waveform generator combines $\mathbf{x}$ with $\tilde{m}$ at inference (or with the ground-truth timing map $m$ during teacher-forced training) to construct the forecast waveform $\hat{\mathbf{y}} = G(\mathbf{x}, \tilde{m})$.
}\label{fig:method} %
\end{figure}

\method organizes continuous waveform forecasting into three sequential stages (Fig.~\ref{fig:method}): \textbf{Predict} ($P$), which estimates \emph{when} future beats will occur; \textbf{Commit} ($C$), which selects and sharpens discrete beat events; and \textbf{Generate} ($G$), which constructs \emph{what} the waveform looks like around those committed beat times:
\begin{equation}\label{eq:pipeline}
\underbrace{\hat{m} = P(\mathbf{x})}_{\text{1. Predict raw timing map}}, \qquad
\underbrace{\hat{E} = C(\hat{m}) \;\Rightarrow\; \tilde{m}}_{\text{2. Commit \& sharpen events}}, \qquad
\underbrace{\hat{\mathbf{y}} = G(\mathbf{x}, \tilde{m})}_{\text{3. Generate waveform}}.
\end{equation}
Specifically, the three stages operate as follows:
\begin{enumerate}
    \item \textbf{Predict ($P$):} The timing predictor $P$ maps the historical waveform $\mathbf{x} \in \mathbb{R}^{H}$ to a dense, continuous timing map $\hat{m} \in [0,1]^{L}$ on the 500-Hz grid of the 60-s horizon, where high values indicate likely beat locations. %
    \item \textbf{Commit ($C$):} Because timing uncertainty causes the raw predictions $\hat{m}$ to broaden and decay in amplitude as lead time increases, the non-neural commit operator $C$ selects a discrete set of beat times $\hat{E} = \{\hat{e}_1, \dots, \hat{e}_{\hat{K}}\}$ from $\hat{m}$ and re-encodes them as a committed, sharpened Gaussian timing map $\tilde{m} \in [0,1]^{L}$ with unit-height peaks. %
    \item \textbf{Generate ($G$):} The waveform generator $G$ conditions on both the committed timing map $\tilde{m}$ and the historical waveform $\mathbf{x}$ to synthesize the continuous 60-s forecast waveform $\hat{\mathbf{y}} \in \mathbb{R}^{L}$. %
\end{enumerate}

\subsection{Shared representation and sequence backbone}\label{sec:architecture}

Modeling a 120-s history ($60{,}000$ samples) and a 60-s horizon ($30{,}000$ samples) directly at 500~Hz would require processing $90{,}000$ time steps per window. %
To make long-horizon sequence modeling computationally tractable while preserving high-frequency waveform morphology, the timing predictor $P$ and waveform generator $G$ share a common token-level representation and sequence backbone (Fig.~\ref{fig:method}):

\begin{itemize}
    \item \textbf{History normalization and autoencoder tokenization ($\mathrm{Enc}, \mathrm{Dec}$):}
    Each input history $\mathbf{x}$ is first z-scored by its own mean and standard deviation. %
    A convolutional autoencoder encoder $\mathrm{Enc}(\cdot)$ compresses the 120-s history $\mathbf{x}$ into $60$ history tokens, where each token summarizes a 2-s segment (roughly two to three cardiac cycles). %
    A symmetric convolutional decoder $\mathrm{Dec}(\cdot)$ is trained jointly to reconstruct $\hat{\mathbf{x}} = \mathrm{Dec}(\mathrm{Enc}(\mathbf{x}))$, ensuring that the token representation retains detailed beat morphology.

    \item \textbf{Horizon query tokens:}
    To represent the 60-s forecast horizon at the same 2-s resolution, we initialize $30$ learnable horizon query tokens. %
    Before entering the backbone, these query tokens are summed with a low-frequency spectral embedding of the history $\mathbf{x}$, providing the queries with initial context regarding the patient's baseline heart rate and respiratory modulation. %

    \item \textbf{Shared bidirectional Mamba--Attention backbone:}
    In both the predictor $P$ and generator $G$, the $60$ history tokens and $30$ horizon query tokens are concatenated into a 90-token sequence and processed by a shared 4-layer backbone (width $512$) composed of bidirectional Mamba state-space blocks~\cite{gu2023} interleaved with self-attention in every second layer (Supplementary Section~S2, Table~S2). %
    Sharing the encoder and backbone across $P$ and $G$ allows both modules to reuse the same learned representation of historical cardiac dynamics without increasing model parameters or inference overhead (Supplementary Section~S2, Tables~S3--S4).
\end{itemize}

\subsection{Detailed design of Predict, Commit, and Generate}

Building on this shared backbone, we now detail how \textbf{Predict} ($P$) estimates future beat probabilities, how \textbf{Commit} ($C$) selects and sharpens discrete beat events, and how \textbf{Generate} ($G$) constructs the continuous waveform around them. %

\paragraph{Predict ($P$): Dense timing map prediction and training loss}
Because the number of heartbeats $K$ within a 60-s window varies across patients and heart rates, directly regressing a variable-length list of timestamps is ill-posed. %
Instead, we represent any set of beat times $E$ on the 500-Hz horizon grid as a continuous Gaussian timing map $m(t)$ by placing a unit-height Gaussian bump of standard deviation $\sigma = 10$~ms at each beat location: %
\begin{equation}\label{eq:bump}
m(t) = \sum_{e \in E} \exp\!\left(-\frac{(t - e)^2}{2\sigma^2}\right), \qquad t \in \{1, \dots, L\}.
\end{equation}
In the prediction stage, the horizon query tokens receive no timing conditioning.
After the shared backbone processes the history and query tokens, a timing head projects the updated horizon tokens back to the 500-Hz grid to produce the raw predicted timing map $\hat{m}(t) = P(\mathbf{x})$. %

Training $P(\mathbf{x})$ with standard MSE against $m(t)$ fails because narrow 10-ms R-peak bumps occupy less than 2\% of the 60-s timeline, causing an unweighted MSE loss to collapse toward an all-zero prediction. %
We therefore train $\hat{m}(t)$ using a composite timing loss $\mathcal{L}_{\mathrm{time}}$ that combines a peak-weighted MSE with a differentiable soft-overlap $F_1$ loss:
\begin{equation}\label{eq:ltime}
\mathcal{L}_{\mathrm{time}}(\hat{m}, m) = \underbrace{\frac{1}{L}\sum_{t=1}^{L} \bigl(1 + w\, m(t)\bigr)\bigl(\hat{m}(t) - m(t)\bigr)^2}_{\text{Peak-weighted MSE}} + \underbrace{\lambda_{\mathrm{F}}\bigl(1 - \mathrm{F}_{\mathrm{soft}}(\hat{m}, m)\bigr)}_{\text{Soft overlap } F_1 \text{ loss}},
\end{equation}
where $w$ heavily penalizes missing true beat peaks, and $\mathrm{F}_{\mathrm{soft}}(\hat{m}, m) = \frac{2\sum_{t}\hat{m}(t)m(t)}{\sum_{t}\hat{m}(t)+\sum_{t}m(t)}$ measures the continuous pointwise overlap $F_1$ between the predicted and ground-truth maps to encourage global peak alignment (Supplementary Section~S2, Table~S2). %

\paragraph{Commit ($C$): Discrete event selection and map sharpening}
During inference, the raw timing map $\hat{m}(t)$ reflects the growing timing uncertainty shown in Fig.~\ref{fig:asymmetry}a: while its peaks are sharp and close to $1.0$ in the first few seconds, they become progressively broader, flatter, and noisier at longer lead times. %
If this blurred map $\hat{m}(t)$ were fed directly into the waveform generator $G$, the generator would receive attenuated, spread-out triggers and reproduce smoothed, flattened heartbeats (as verified by our raw-map substitution in Section~\ref{sec:interface}).

The commit step $C$ prevents this degradation by acting as a discretizing and sharpening bottleneck.
Given $\hat{m}(t)$, $C$ selects discrete beat times $\hat{E} = C(\hat{m}) = \{\hat{e}_1, \dots, \hat{e}_{\hat{K}}\}$ by identifying local maxima above a validation-selected threshold while enforcing a physiological minimum refractory spacing equal to half the median of the last five eligible historical RR intervals (preventing physiologically implausible double-peaks; Supplementary Section~S2). %
We then place a clean, unit-height 10-ms Gaussian bump at each committed time $\hat{e}_k \in \hat{E}$ via Eq.~\eqref{eq:bump} to form the committed, sharpened timing map $\tilde{m}(t)$: %
\begin{equation}\label{eq:m_tilde}
\tilde{m}(t) = \sum_{\hat{e} \in \hat{E}} \exp\!\left(-\frac{(t - \hat{e})^2}{2\sigma^2}\right).
\end{equation}
By converting a decaying, uncertain probability map $\hat{m}(t)$ into a sequence of crisp, unit-amplitude impulses $\tilde{m}(t)$, the commit step ensures that the generator $G$ always receives clean beat triggers matching the format of the ground-truth map $m(t)$.

\paragraph{Generate ($G$): Coarse-to-fine waveform generation}
In the generation stage, the waveform generator $G(\mathbf{x}, \tilde{m})$ synthesizes the 500-Hz forecast waveform $\hat{\mathbf{y}}$ by injecting the committed timing map $\tilde{m}$ at two complementary spatial scales: %
\begin{enumerate}
    \item \textbf{Coarse token-level alignment (2-s scale):}
    A convolutional token embedding of $\tilde{m}$ is added directly to the $30$ horizon query tokens before they enter the shared Mamba--Attention backbone. %
    This informs the sequence backbone of the planned beat schedule within each 2-s window, allowing it to retrieve and adapt the patient's historical beat morphology from the $60$ history tokens. %

    \item \textbf{Fine sample-level anchoring (500-Hz scale):}
    After the backbone processes the tokens, the convolutional decoder $\mathrm{Dec}(\cdot)$ upsamples the $30$ horizon tokens back to a 500-Hz waveform. %
    A final convolutional fusion layer then combines this decoded waveform with the full-resolution 500-Hz timing map $\tilde{m}(t)$. %
    This skip connection provides millisecond-precise temporal anchoring so that sharp R peaks (in ECG) or pulse upstrokes (in ABP and PPG) are rendered accurately at the committed timestamps (without imposing a hard post-hoc constraint on peak amplitudes).
\end{enumerate}

\paragraph{Teacher-forced training versus committed inference}
A critical question is how to train the waveform generator $G$ so that it learns to generate sharp heartbeats rather than smoothed averages.
If the generator $G$ were trained using the predicted map $\tilde{m}$ from $P$ and $C$, any slight timing error between a predicted beat $\hat{e}_k$ and the true beat $e_k$ in the target waveform $\mathbf{y}$ would cause pointwise waveform losses to heavily penalize drawing a sharp beat peak at $\hat{e}_k$, pressuring the generator to smooth out its peaks. %
To decouple morphology learning from timing prediction error during training, we train the generator $G$ with \emph{teacher forcing}: during training, $G(\mathbf{x}, m)$ receives the ground-truth timing map $m(t)$ so that every conditioned beat location aligns exactly with the true beat in $\mathbf{y}$.

The entire network is trained end-to-end by minimizing the total loss
\begin{equation}\label{eq:loss}
\mathcal{L} = \mathcal{L}_{\mathrm{time}}(\hat{m}, m) + \mathcal{L}_{\mathrm{wave}}(\hat{\mathbf{y}}, \mathbf{y}) + \lambda_{\mathrm{h}}\,\mathcal{L}_{\mathrm{wave}}(\hat{\mathbf{x}}, \mathbf{x}),
\end{equation}
where $\hat{\mathbf{x}}$ is the autoencoder reconstruction of the history $\mathbf{x}$ (weighted by $\lambda_{\mathrm{h}} = 0.3$), and the waveform loss $\mathcal{L}_{\mathrm{wave}}$ combines samplewise MSE with first- and second-order finite-difference errors to preserve steep beat upstrokes and sharp peak curvature: %
\begin{equation}\label{eq:lwave}
\mathcal{L}_{\mathrm{wave}}(\hat{\mathbf{y}}, \mathbf{y}) = \|\hat{\mathbf{y}} - \mathbf{y}\|_2^2 + \|\Delta \hat{\mathbf{y}} - \Delta \mathbf{y}\|_2^2 + \|\Delta^2 \hat{\mathbf{y}} - \Delta^2 \mathbf{y}\|_2^2,
\end{equation}
where $\|\cdot\|_2^2$ denotes the mean squared error over samples and $\Delta, \Delta^2$ denote first- and second-order temporal differences.
During inference, teacher forcing is removed and the generator $G$ receives the committed timing map $\tilde{m}$ from $C(\hat{m})$, using no future information.

\section{Experimental evaluation}\label{sec:evaluation}

We organize our experimental evaluation around four research questions:
\begin{itemize}
    \item \textbf{RQ1 (Comparison with existing forecasters):} How does \method compare against existing time-series and physiological forecasting methods in preserving heartbeats over continuous forecasts (Section~\ref{sec:published})?
    \item \textbf{RQ2 (Ablation study on the \method pipeline):} How does each component of \method, namely explicit timing conditioning (versus unconditioned Direct and Direct-MT), the commit sharpening step $C$, the timing predictor $P$, and the learned waveform generator $G$, contribute to overall performance (Section~\ref{sec:rq2})?
    \item \textbf{RQ3 (Generalization and cross-signal robustness):} Does \method generalize effectively across different waveform loss functions, backbone architectures, external ECG cohorts, and ABP and PPG pulse waveforms (Section~\ref{sec:rq3})?
    \item \textbf{RQ4 (Mechanism of the sharpness--MSE trade-off):} Why does preserving sharp heartbeats under timing uncertainty inevitably increase samplewise squared error (Section~\ref{sec:synthetic})?
\end{itemize}

\subsection{Experimental setup}\label{sec:protocol}

\paragraph{Datasets and cohorts}\label{sec:data}
We evaluate our framework primarily on intraoperative physiological recordings from VitalDB~\cite{lee2022} sampled at 500~Hz. %
Each evaluation window uses a 120-s history to forecast the subsequent 60~s within a continuous segment, with at most three non-overlapping windows per patient. %
For ECG, patients are partitioned into disjoint training, validation, test, and lockbox cohorts, with sample sizes summarized in Table~\ref{tab:data}.
Model hyperparameters and decision rules were tuned on the validation cohort, and primary evaluations were conducted on the test cohort.
To guard against over-tuning across analysis rounds, the lockbox cohort consists of patients held out from an earlier training-only pool and excluded from final model training, serving as a secondary held-out test set for independent replication (Supplementary Sections~S3 and~S5).

To evaluate zero-shot generalization across hospitals and cardiac rhythms without retraining, we test the trained ECG models on two external datasets: MOVER~\cite{samad2023}, an intraoperative ECG dataset from a separate medical center, and SHDB-AF~\cite{tsutsui2025}, an ambulatory Holter ECG dataset from patients with atrial fibrillation (AF).
To evaluate cross-signal applicability beyond ECG, we also train and test signal-specific models on VitalDB arterial blood pressure (ABP) and finger photoplethysmogram (PPG) waveforms from subsets of the ECG test cohort.
Cohort screening characteristics, reference verification, and preprocessing details are provided in Supplementary Section~S3.

\begin{table}[!htbp]
\caption{Summary of training and evaluation cohorts across ECG, ABP, and PPG signals: windows / patients (SHDB-AF: records).}\label{tab:data}
\centering
{\small\setlength{\tabcolsep}{3.5pt}%
\begin{tabular}{@{}llllll@{}}
\toprule
Signal & Dataset & Reference & Set & \makecell[l]{Before\\ screen} & Evaluable \\
\midrule
ECG & \makecell[l]{VitalDB\\ (intraoperative)} & \makecell[l]{Single-lead\\ R peaks} & Training & \multicolumn{2}{l}{\makecell[l]{\NtrainPat patients\\\NtrainHours archived h}} \\
 & & Fused R peaks & Validation & \NwinValPreGate{} / \NecgValPatPreGate & \NwinVal{} / \NpatVal \\
 & & & Test & \NwinDiscPreGate{} / \NecgDiscPatPreGate & \NwinDisc{} / \NpatDisc \\
 & & & Lockbox & \NwinLbxPreGate{} / \NecgLbxPatPreGate & \NwinLbx{} / \NpatLbx \\
\midrule
ABP & VitalDB & Pulse anchors & Test & -- & \NabpWin{} / \NabpPat \\
PPG & VitalDB & Pulse anchors & Test & -- & \NppgWin{} / \NppgPat \\
\midrule
ECG & \makecell[l]{MOVER\\ (external)} & \makecell[l]{Single-lead\\ R peaks} & Test & \NmoverWinPreGate{} / \NmoverPatPreGate & \NmoverWin{} / \NmoverPat \\
ECG & \makecell[l]{SHDB-AF\\ (Holter, external)} & \makecell[l]{Automatic\\ R peaks} & \makecell[l]{Test\\ (AF)} & -- & \makecell[l]{\NshdbWin{} / \NshdbRec\\ (\NshdbAfWin{} / \NshdbAfRec)} \\
\bottomrule
\end{tabular}}
\end{table}

\begin{figure}[!htbp]
\centering
\WWfig{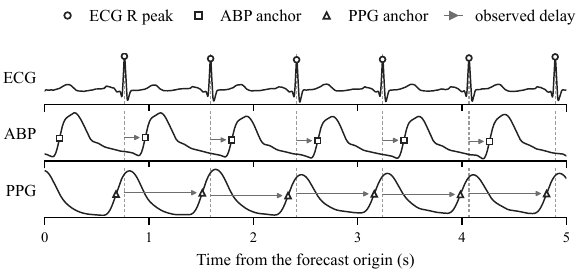}{\linewidth}
\caption{ECG R peaks and maximal-upslope pulse anchors on z-scored signals.
Arrows indicate observed R-to-anchor physiological delays (verification details in Supplementary Section~S3, Table~S7).}\label{fig:anchors}
\end{figure}

\paragraph{Reference events, evaluation channels, and statistical protocol}\label{sec:ref}\label{sec:detector-protocol}\label{sec:stats}\label{sec:metrics}
Ground-truth ECG training labels are extracted using NeuroKit2~\cite{makowski2021}.
For ECG evaluation, reference R peaks on VitalDB and MOVER are obtained via a quality-screened reference pipeline that excludes windows containing unresolvable signal artifacts in the forecast horizon, while SHDB-AF uses its distributed R-peak and expert rhythm annotations (Table~\ref{tab:data}; reference construction and verification details in Supplementary Section~S3).
For ABP and PPG waveforms, reference events correspond to maximal-upslope pulse anchors extracted using the slope-sum onset detector~\cite{zong2003} and the Elgendi detector~\cite{elgendi2013}, which align closely with synchronous ECG R peaks (Fig.~\ref{fig:anchors}; Supplementary Section~S3, Table~S7).

We evaluate forecast performance along two complementary dimensions: beat-timing accuracy and waveform fidelity.
Because our framework produces both discrete beat-time predictions $\hat{E} = C(\hat{m})$ from the timing predictor and a continuous forecast waveform $\hat{\mathbf{y}}$ from the waveform generator, we evaluate beat-timing accuracy along two evaluation paths:
\begin{itemize}
    \item The \textbf{event channel} scores the discrete events $\hat{E}$ output by the timing predictor directly against the reference events $E$.
    \item The \textbf{waveform channel} uses the standard NeuroKit2 detector~\cite{makowski2021} to extract beats from the generated waveform $\hat{\mathbf{y}}$ and scores those detected beats against the same reference events $E$ (additional matching-tolerance and detector sensitivity analyses are provided in Supplementary Sections~S5 and~S6, Tables~S14 and~S21).
\end{itemize}
In both channels, our primary metric is the near-term event skill $S$ over 0--10~s, defined as the unweighted mean of $S$ (Eq.~\eqref{eq:skill}) across the $[0,2)$, $[2,5)$, and $[5,10)$~s lead-time bins, alongside event $F_1$ for 2-s short-horizon baselines. %
For waveform fidelity, we report samplewise MSE over the full 60-s horizon on z-scored waveforms (alongside per-beat morphology RMSE and Shapley timing-versus-morphology error decomposition in Supplementary Section~S5, Table~S11). %
Each model is trained across multiple random seeds (Supplementary Section~S2, Table~S2).
To quantify estimation uncertainty, we average scores across seeds for each window and compute 95\% confidence intervals using patient-cluster bootstrap resampling (full statistical protocols in Supplementary Sections~S3 and~S5). %

\FloatBarrier
\subsection{Comparison with existing forecasting baselines (RQ1)}\label{sec:published}

\begin{table}[!htbp]
\caption{Comparison of \method against existing time-series and physiological forecasting baselines on the VitalDB ECG test cohort (\NwinDisc windows from \NpatDisc patients). %
The last column reports the paired improvement ($\Delta$) of \method over each baseline (95\% confidence intervals and additional hyperparameter configurations are reported in Supplementary Section~S4, Table~S8).}\label{tab:published} %
\centering
{\small\setlength{\tabcolsep}{3pt}%
\begin{tabularx}{\linewidth}{@{}>{\raggedright\arraybackslash}X>{\raggedright\arraybackslash}p{.32\linewidth}cc@{}}
\toprule
Method & Evaluation metric & \makecell{Baseline\\ score} & \makecell{\textbf{\method{}}\\ \textbf{gain ($\Delta$)}} \\
\midrule
\multicolumn{4}{@{}p{\linewidth}@{}}{\emph{Long-horizon waveform forecasting: 120\,s history $\to$ 60\,s forecast (waveform channel)}} \\
Informer~\cite{zhou2021} & Event skill $S$ (0--10\,s) & \NdthreeInfRLev & \textbf{\NdthreeInfRDiff} \\
HTF~\cite{zhang2026htf} & Event skill $S$ (0--10\,s) & \NdthreeHtfOLev & \textbf{\NdthreeHtfODiff} \\
BeatLex~\cite{hooi2017} & Event skill $S$ (0--10\,s) & \NdthreeBlxRLev & \textbf{\NdthreeBlxRDiff} \\
DLinear~\cite{zeng2023} (100\,Hz) & Event skill $S$ (0--10\,s) & \NdthreeDlBLev & \textbf{\NdthreeDlBDiff} \\
PatchTST~\cite{nie2023} (100\,Hz) & Event skill $S$ (0--10\,s) & \NdthreePtBLev & \textbf{\NdthreePtBDiff} \\
\textbf{\method{} (Ours)} & Event skill $S$ (0--10\,s) & \textbf{\NwaveWWDisc} & -- \\
\midrule
\multicolumn{4}{@{}p{\linewidth}@{}}{\emph{Short-horizon waveform forecasting: 10\,s history $\to$ 2\,s or 10\,s forecast (waveform channel)}} \\
HTF~\cite{zhang2026htf} (2-s forecast) & Event $F_1$ (0--2\,s) & \NdthreeHtfALev & \textbf{\NdthreeHtfADiff} \\
DLinear~\cite{zeng2023} (2-s forecast) & Event $F_1$ (0--2\,s) & \NdthreeDlALev & \textbf{\NdthreeDlADiff} \\
PatchTST~\cite{nie2023} (2-s forecast) & Event $F_1$ (0--2\,s) & \NdthreePtALev & \textbf{\NdthreePtADiff} \\
FADE~\cite{ruizbarroso2025} (2-s forecast) & Event $F_1$ (0--2\,s) & \NdthreeFadeAtwoLev & \textbf{\NdthreeFadeAtwoDiff} \\
\textbf{\method{} (Ours, 0--2\,s)} & Event $F_1$ (0--2\,s) & \textbf{\NdthreeWwFone} & -- \\
\cmidrule(lr){1-4}
DLinear~\cite{zeng2023} (10-s forecast) & Event skill $S$ (0--10\,s) & \NdthreeDlAtenLev & \textbf{\NdthreeDlAtenDiff} \\
FADE~\cite{ruizbarroso2025} (10-s forecast) & Event skill $S$ (0--10\,s) & \NdthreeFadeAtenLev & \textbf{\NdthreeFadeAtenDiff} \\
\textbf{\method{} (Ours, 0--10\,s)} & Event skill $S$ (0--10\,s) & \textbf{\NwaveWWDisc} & -- \\
\midrule
\multicolumn{4}{@{}p{\linewidth}@{}}{\emph{Beat-time (RR-series) forecasting over 60\,s (event channel)}} \\
DLinear-RR~\cite{zeng2023} & Event skill $S$ (0--10\,s) & \NdthreeDlRrLev & \textbf{\NdthreeDlRrDiff} \\
PatchTST-RR~\cite{nie2023} & Event skill $S$ (0--10\,s) & \NdthreePtRrLev & \textbf{\NdthreePtRrDiff} \\
AR(8) & Event skill $S$ (0--10\,s) & \NdthreeArEightLev & \textbf{\NdthreeArEightDiff} \\
GRU (Huber loss) & Event skill $S$ (0--10\,s) & \NdthreeGruHubLev & \textbf{\NdthreeGruHubDiff} \\
GBM (Huber loss) & Event skill $S$ (0--10\,s) & \NdthreeGbmHubLev & \textbf{\NdthreeGbmHubDiff} \\
TimesFM (500M, zero-shot)~\cite{das2024} & Event skill $S$ (0--10\,s) & \NdthreeTfmLev & \textbf{\NdthreeTfmDiff} \\
\textbf{\method{} timing head (Ours)} & Event skill $S$ (0--10\,s) & \textbf{\NnatWWDisc} & -- \\
\bottomrule
\end{tabularx}}

\end{table}

We first compare \method on the VitalDB ECG test cohort against a comprehensive set of existing time-series and physiological forecasting baselines, including Informer~\cite{zhou2021}, DLinear~\cite{zeng2023}, PatchTST~\cite{nie2023}, BeatLex~\cite{hooi2017}, FADE~\cite{ruizbarroso2025}, HTF~\cite{zhang2026htf}, zero-shot TimesFM~\cite{das2024}, and standard autoregressive (AR), recurrent (GRU), and gradient-boosted (GBM) models (Table~\ref{tab:published}; implementation details in Supplementary Section~S4).
Because published methods operate under different look-back lengths, sampling rates, and output modalities, Table~\ref{tab:published} evaluates each baseline under its respective task configuration and reports the paired improvement achieved by \method.

Across every comparison group and every baseline in Table~\ref{tab:published}, \method achieves the highest beat-detection accuracy.
In the 60-s continuous waveform forecasting task (120-s look-back), direct long-horizon sequence models average across uncertain future heartbeat phases and flatten into smooth baselines, resulting in negative 0--10-s event skill for Informer (\NdthreeInfRLev), HTF (\NdthreeHtfOLev), DLinear (\NdthreeDlBLev), and PatchTST (\NdthreePtBLev). %
The pattern-concatenation forecaster BeatLex avoids complete flattening but reaches an event skill of only \NdthreeBlxRLev as pattern transitions drift out of phase.
By contrast, \method achieves a waveform event skill of \NwaveWWDisc, outperforming BeatLex by \NdthreeBlxRDiff and direct sequence forecasters by more than 0.70 skill points. %

In short-horizon waveform settings (10-s look-back), PatchTST and the ECG-specific forecaster FADE achieve reasonable beat $F_1$ over the immediate 0--2-s window (\NdthreePtALev and \NdthreeFadeAtwoLev), yet \method still outperforms both with a 0--2-s $F_1$ of \NdthreeWwFone (gains of \NdthreePtADiff and \NdthreeFadeAtwoDiff). %
More importantly, when the evaluation window extends to 0--10~s, FADE's event skill drops sharply to \NdthreeFadeAtenLev and DLinear's to \NdthreeDlAtenLev, whereas \method maintains \NwaveWWDisc (outperforming FADE by \NdthreeFadeAtenDiff). %
Finally, when directly forecasting future beat times from historical RR intervals over 60~s, \method's timing predictor (\NnatWWDisc event-channel skill) consistently surpasses AR (\NdthreeArEightLev), GRU (\NdthreeGruHubLev), PatchTST-RR (\NdthreePtRrLev), GBM (\NdthreeGbmHubLev), and the 500M-parameter foundation model TimesFM (\NdthreeTfmLev, a gain of \NdthreeTfmDiff). %

\FloatBarrier
\subsection{Ablation study on the When2What pipeline (RQ2)}\label{sec:rq2}\label{sec:controls}\label{sec:interface}

\begin{table}[!htbp]
\caption{Configurations of the matched conditioning ablation models. Direct, Direct-MT, and \method share the exact encoder, Mamba--Attention backbone, and waveform loss $\mathcal{L}_{\mathrm{wave}}$, differing only in auxiliary timing supervision and the timing map supplied to the waveform generator $G$.}\label{tab:models}
\centering
{\setlength{\tabcolsep}{3pt}%
\begin{tabularx}{\linewidth}{@{}l>{\centering\arraybackslash}X>{\centering\arraybackslash}X>{\centering\arraybackslash}X@{}}
\toprule
Model variant & \makecell{Timing loss\\ on $\hat{m}$} & \makecell{Generator $G$\\ timing input\\ (training)} & \makecell{Generator $G$\\ timing input\\ (inference)} \\
\midrule
Direct & -- & zero map $\mathbf{0}$ & zero map $\mathbf{0}$ \\
Direct-MT & $\mathcal{L}_{\mathrm{time}}$ & zero map $\mathbf{0}$ & zero map $\mathbf{0}$ \\
\method{} (Ours) & $\mathcal{L}_{\mathrm{time}}$ & \makecell{ground-truth\\ map $m$} & \makecell{committed\\ map $\tilde{m}$} \\
\bottomrule
\end{tabularx}}
\end{table}

\begin{table}[!htbp]
\caption{Ablation study of the \method pipeline on the VitalDB ECG test cohort (\NwinDisc windows from \NpatDisc patients). %
We report first-10-s event skill $S$ in the event and waveform channels, the waveform skill change ($\Delta$) relative to full \method, and 0--60-s waveform MSE (95\% confidence intervals and lockbox replications are provided in Supplementary Section~S5, Tables~S9–S11).}\label{tab:native-waveform}\label{fig:construction} %
\centering
{\small\setlength{\tabcolsep}{4.5pt}%
\begin{tabular}{@{}lcccc@{}}
\toprule
Ablation variant & \makecell{Event\\ skill $S$} & \makecell{Waveform\\ skill $S$} & \makecell{Skill $\Delta$\\ vs.\ Ours} & \makecell{Waveform\\ MSE} \\
\midrule
\multicolumn{5}{@{}l}{\emph{Conditioning path ablations (original waveform loss)}} \\
Direct (w/o $P$, $C$, and $\tilde{m}$) & -- & \NwaveDirectDisc & \ensuremath{-}0.106 & \NmseDirectDisc \\
Direct-MT (with $P$, w/o $\tilde{m}$) & \NnatDirectMTDisc & \NwaveDirectMTDisc & \ensuremath{-}0.110 & \NmseDirectMTDisc \\
\textbf{\method{} (full: $P + C + G$)} & \textbf{\NnatWWDisc} & \textbf{\NwaveWWDisc} & \textbf{Ref.} & \NmseWWDisc \\
\midrule
\multicolumn{5}{@{}l}{\emph{Component ablations on Commit ($C$), Predictor ($P$), and Generator ($G$)}} \\
w/o Commit $C$ (raw map $\hat{m} \to G$) & \NnatWWDisc & 0.441 & \NifSoft & -- \\
Direct waveform peaks $\to G$ & -- & 0.496 & \NdirectReadRenderer & -- \\
Direct-MT timing $\to G$ & \NnatDirectMTDisc & 0.641 & \NheadRendererDisc & 1.668 \\
\method{} timing $\to$ Template & \NnatWWDisc & 0.639 & \ensuremath{-}0.0045 & 1.789 \\
Direct-MT timing $\to$ Template & \NnatDirectMTDisc & 0.636 & \NufHeadTemplate & 1.776 \\
Guarded cadence $\to$ Template & 0.640 & 0.607 & \NguardedTplMinusWW & 1.792 \\
\midrule
\multicolumn{5}{@{}l}{\emph{Conditioning path ablations (Split-MSE waveform loss)}} \\
Direct (Split) & -- & \NwaveDirectSplitDisc & \ensuremath{-}0.068 & \NmseDirectSplitDisc \\
Direct-MT (Split) & \NnatDirectMTSplitDisc & \NwaveDirectMTSplitDisc & \ensuremath{-}0.067 & \NmseDirectMTSplitDisc \\
\textbf{\method{} (Split)} & \NnatWWSplitDisc & \textbf{\NwaveWWSplitDisc} & \textbf{Ref.} & \NmseWWSplitDisc \\
\bottomrule
\end{tabular}}

\end{table}

\begin{figure}[!htbp]
\centering
\WWfig{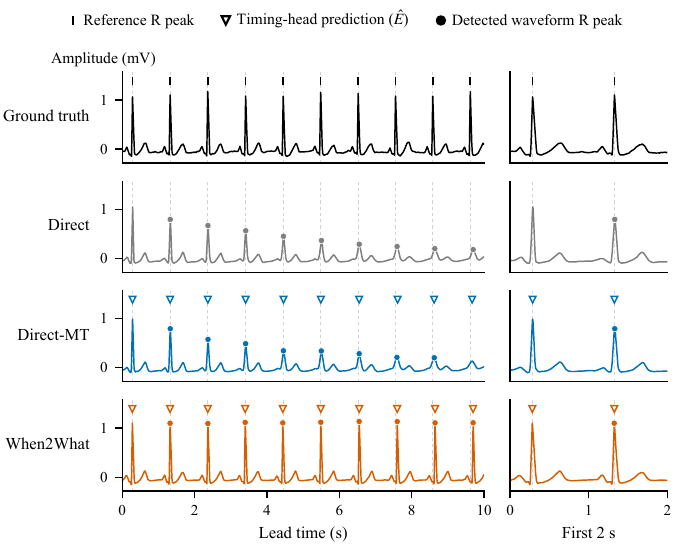}{\linewidth}
\caption{Representative 10-s ECG forecast on the test cohort, with a zoomed 0--2~s view on the right. %
Reference R peaks (vertical ticks), timing-head predictions (triangles), and beats detected from the generated waveform (circles) are shown on a common mV scale.
Direct-MT predicts accurate beat times in its timing head, but its unconditioned waveform output smooths out the corresponding R peaks; \method preserves sharp, detectable R peaks at the predicted times.}\label{fig:example}
\end{figure}

To isolate why \method succeeds and quantify the contribution of each module in our pipeline, we conduct systematic ablations across the conditioning path, the commit sharpening step $C$, the timing predictor $P$, and the waveform generator $G$ in Table~\ref{tab:native-waveform} (full confidence intervals and lockbox replications are provided in Supplementary Section~S5, Tables~S9–S11).

\paragraph{Ablating explicit timing conditioning (\method vs.\ Direct and Direct-MT)}
We first examine whether \method's performance comes from multi-task representation learning or from explicitly feeding the committed timing map $\tilde{m}$ into the waveform generator $G$.
To test this, we compare full \method against two strictly matched ablation variants that share the exact network architecture, parameter count, training schedule, and waveform loss $\mathcal{L}_{\mathrm{wave}}$ (Table~\ref{tab:models}).
The first variant, \textbf{Direct Regression (Direct)}, removes both the timing predictor $P$ and the conditioning input $\tilde{m}$, passing an all-zero timing map $\mathbf{0}$ to the generator $G$ during training and inference.
The second variant, \textbf{Direct Regression with Auxiliary Timing Supervision (Direct-MT)}, retains the exact timing predictor $P(\mathbf{x})$ and timing loss $\mathcal{L}_{\mathrm{time}}$ of \method, predicting $\hat{m}$ and $\hat{\mathbf{y}}$ in parallel, but passes an all-zero timing map $\mathbf{0}$ to the generator $G$ during training and inference.

As shown in Table~\ref{tab:native-waveform}, Direct-MT and \method achieve virtually identical beat-time skill in the event channel (\NnatDirectMTDisc and \NnatWWDisc), confirming that both models learn to predict \emph{when} future beats will occur equally well.
In the waveform channel, however, training the timing head as an auxiliary task without feeding its predictions into the generator brings no benefit: Direct-MT (\NwaveDirectMTDisc) performs no better than Direct (\NwaveDirectDisc), whereas \method raises waveform event skill to \NwaveWWDisc, a gain of \NdcondOrigDisc under the original loss and \NdcondSplitDisc under the Split-MSE loss.
Figure~\ref{fig:example} shows visually why Direct-MT fails: although its timing head accurately localizes future beats, its unconditioned generator averages across uncertain phases and outputs flattened deflections where R peaks cannot be detected.
Tracing each reference beat correctly localized by a model's timing head in the first 10~s confirms that Direct-MT reproduces only \NfollowDiscCMTExpr of its correctly predicted beats in the output waveform, whereas \method reproduces \NfollowDiscCWWExpr (Supplementary Section~S5, Tables~S15–S17). %
At the same time, Table~\ref{tab:native-waveform} shows that generating sharp R peaks at predicted times increases 0--60-s samplewise MSE relative to Direct (\NmseWWDisc versus \NmseDirectDisc), because any slight timing offset of a sharp peak incurs a larger pointwise penalty than a flat line. %

\paragraph{Ablating the commit sharpening step ($C$)}
In Section~\ref{sec:method}, we designed the commit step $C$ to convert the broadened, decaying peaks of the raw timing map $\hat{m}$ into clean, unit-height Gaussian impulses $\tilde{m}$.
When we bypass $C$ at inference time and feed the unthresholded raw timing map $\hat{m}$ directly into the trained generator $G$, first-10-s waveform event skill drops sharply by \NifSoft (Table~\ref{tab:native-waveform}). %
Similarly, feeding noisy peaks extracted from Direct's output waveform into $G$ reduces waveform event skill by \NdirectReadRenderer.
This confirms that discrete event selection and sharpening via $C$ are essential for providing clear triggers to the generator.

\paragraph{Ablating the timing predictor ($P$) and waveform generator ($G$)}
Next, keeping \method's trained generator $G$ fixed at inference and feeding it the beat events predicted by Direct-MT's timing head yields nearly identical performance to full \method, differing by only \NheadRendererDisc in waveform event skill (Table~\ref{tab:native-waveform}).
This confirms that Direct-MT's limitation lies in its unconditioned generator rather than its timing head.
Conversely, replacing the learned neural generator $G$ with a static historical beat template that copies and pastes the median of the patient's last 40 beats at the predicted beat times reduces waveform event skill to 0.639 with \method timing and 0.636 with Direct-MT timing, and further to 0.607 when combined with a non-learned guarded cadence rule (Table~\ref{tab:native-waveform}). %
More importantly, because a static template merely repeats a fixed average beat shape without capturing dynamic beat-to-beat cardiac variations, it degrades 0--60-s waveform MSE from \NmseWWDisc to 1.789 and exhibits much poorer robustness when input events contain misses or false alarms: under 30\% event deletion or insertion perturbations, the learned generator $G$ outperforms the static template in event skill retention by \NifDelRobust and \NifInsRobust, respectively (Supplementary Section~S5, Table~S11). %

\FloatBarrier
\subsection{Generalization across losses, backbones, cohorts, and pulse signals (RQ3)}\label{sec:rq3}\label{sec:pulse}

\begin{table}[!htbp]
\caption{Robustness of the conditioning gain (\method{} minus Direct-MT in first-10-s waveform event skill $S$) across waveform loss functions, network backbones, and evaluation cohorts (95\% confidence intervals are provided in Supplementary Section~S6, Tables~S18–S19).}\label{tab:dcond} %
\centering
\begingroup%
\def\TblOrigDiscBase{0.533}%
\def\TblOrigDiscOurs{0.644}%
\def\TblSplitDiscBase{0.560}%
\def\TblSplitDiscOurs{0.627}%
\def\TblSplitTunedDiscBase{0.557}%
\def\TblSplitTunedDiscOurs{0.631}%
\def\TblTCNDiscBase{0.534}%
\def\TblTCNDiscOurs{0.631}%
\def\TblOrigLbxBase{0.541}%
\def\TblOrigLbxOurs{0.651}%
\def\TblSplitLbxBase{0.567}%
\def\TblSplitLbxOurs{0.637}%
\def\TblMOVERBase{0.534}%
\def\TblMOVEROurs{0.632}%
\def\TblSHDBBase{0.238}%
\def\TblSHDBOurs{0.348}%
\setlength{\tabcolsep}{4pt}
\begin{tabular}{@{}>{\raggedright\arraybackslash}p{3.2cm}lccc@{}}
\toprule
\makecell[l]{Experimental\\ setting} & Cohort & Direct-MT & \makecell{\textbf{\method{}}\\ \textbf{(Ours)}} & \makecell{\textbf{Gain}\\ \textbf{($\Delta$)}} \\
\midrule
\multicolumn{5}{@{}l}{\emph{Waveform loss}} \\
Original loss & Test & \TblOrigDiscBase & \textbf{\TblOrigDiscOurs} & \textbf{\NdcondOrigDisc} \\
Split-MSE loss & Test & \TblSplitDiscBase & \textbf{\TblSplitDiscOurs} & \textbf{\NdcondSplitDisc} \\
Split-MSE loss (tuned) & Test & \TblSplitTunedDiscBase & \textbf{\TblSplitTunedDiscOurs} & \textbf{\NdcondSplitTunedDisc} \\
\midrule
\multicolumn{5}{@{}l}{\emph{Backbone architecture}} \\
TCN backbone, original loss & Test & \TblTCNDiscBase & \textbf{\TblTCNDiscOurs} & \textbf{\NdcondTCNDisc} \\
\midrule
\multicolumn{5}{@{}l}{\emph{Held-out and external cohorts}} \\
Original loss & Lockbox & \TblOrigLbxBase & \textbf{\TblOrigLbxOurs} & \textbf{\NdcondOrigLbx} \\
Split-MSE loss & Lockbox & \TblSplitLbxBase & \textbf{\TblSplitLbxOurs} & \textbf{\NdcondSplitLbx} \\
Original loss (zero-shot) & MOVER & \TblMOVERBase & \textbf{\TblMOVEROurs} & \textbf{\NdcondMOVER} \\
Original loss (zero-shot) & SHDB-AF & \TblSHDBBase & \textbf{\TblSHDBOurs} & \textbf{\NdcondSHDB} \\
\bottomrule
\end{tabular}
\endgroup

\end{table}

\paragraph{Loss functions, sequence backbones, and external ECG cohorts}
We next verify that the conditioning gain of \method over Direct-MT holds across different training objectives, backbone architectures, and unseen patient cohorts (Table~\ref{tab:dcond}).
First, training the models with FADE-style Split-MSE~\cite{ruizbarroso2025} up-weights high-amplitude deflections, yet \method~(Split) continues to substantially outperform unconditioned Direct-MT~(Split) by \NdcondSplitDisc in waveform event skill, and by \NdcondSplitTunedDisc after hyperparameter tuning on validation data (Supplementary Section~S6, Table~S18).
Second, replacing the bidirectional Mamba blocks with a causal dilated TCN backbone in both the predictor $P$ and generator $G$ preserves a strong conditioning gain of \NdcondTCNDisc.
Third, as shown in Table~\ref{tab:dcond}, the conditioning gain replicates consistently on the held-out VitalDB lockbox cohort (\NdcondOrigLbx) and transfers zero-shot without retraining to external intraoperative ECG on MOVER (\NdcondMOVER) and ambulatory Holter ECG on SHDB-AF (\NdcondSHDB; detector sensitivity and participant-cluster analyses are provided in Supplementary Section~S6, Tables~S18, S19, and~S21).

\begin{table}[!htbp]
\caption{Comparison of \method{} against baseline models on the VitalDB ABP and PPG test cohorts over the first 10~s (95\% confidence intervals are provided in Supplementary Section~S6, Table~S20).}\label{tab:pulse} %
\centering
{\small\setlength{\tabcolsep}{3pt}%
\begin{tabularx}{\linewidth}{@{}l>{\raggedright\arraybackslash}Xccc@{}}
\toprule
Signal & Metric (0--10\,s) & Baseline & \makecell{\textbf{\method{}}\\ \textbf{(Ours)}} & \textbf{Gain ($\Delta$)} \\
\midrule
PPG & Waveform event $F_1$ (vs.\ Direct) & 0.671 & \textbf{0.686} & \textbf{\NppgWaveFoneVsDirect} \\
PPG & Event skill $S$ (vs.\ simple RR rule) & 0.621 & \textbf{0.666} & \textbf{\NppgSkillVsNaive} \\
\midrule
ABP & Waveform event $F_1$ (vs.\ Direct) & 0.681 & \textbf{0.719} & \textbf{\NabpWaveFoneVsDirect} \\
ABP & Event skill $S$ (vs.\ simple RR rule) & 0.660 & \textbf{0.679} & \textbf{\NabpSkillVsNaive} \\
\bottomrule
\end{tabularx}}

\end{table}

\paragraph{ABP and PPG pulse waveforms}
Evaluating signal-specific \method and Direct models on arterial blood pressure (ABP) and photoplethysmogram (PPG) waveforms confirms that our framework generalizes to hemodynamic signals (Table~\ref{tab:pulse}).
In the waveform channel, \method improves first-10-s pulse-event $F_1$ over Direct regression by \NabpWaveFoneVsDirect on ABP and \NppgWaveFoneVsDirect on PPG. %
In the event channel, \method's timing head also achieves higher first-10-s event skill than the simple RR-extrapolation rule on both signals (\NabpSkillVsNaive on ABP and \NppgSkillVsNaive on PPG), while exhibiting a similar predictability time scale $\tau$ (Supplementary Section~S6, Table~S20). %

\FloatBarrier
\subsection{Synthetic analysis of the sharpness--MSE trade-off (RQ4)}\label{sec:synthetic}

\begin{figure}[!htbp]
\centering
\WWfig{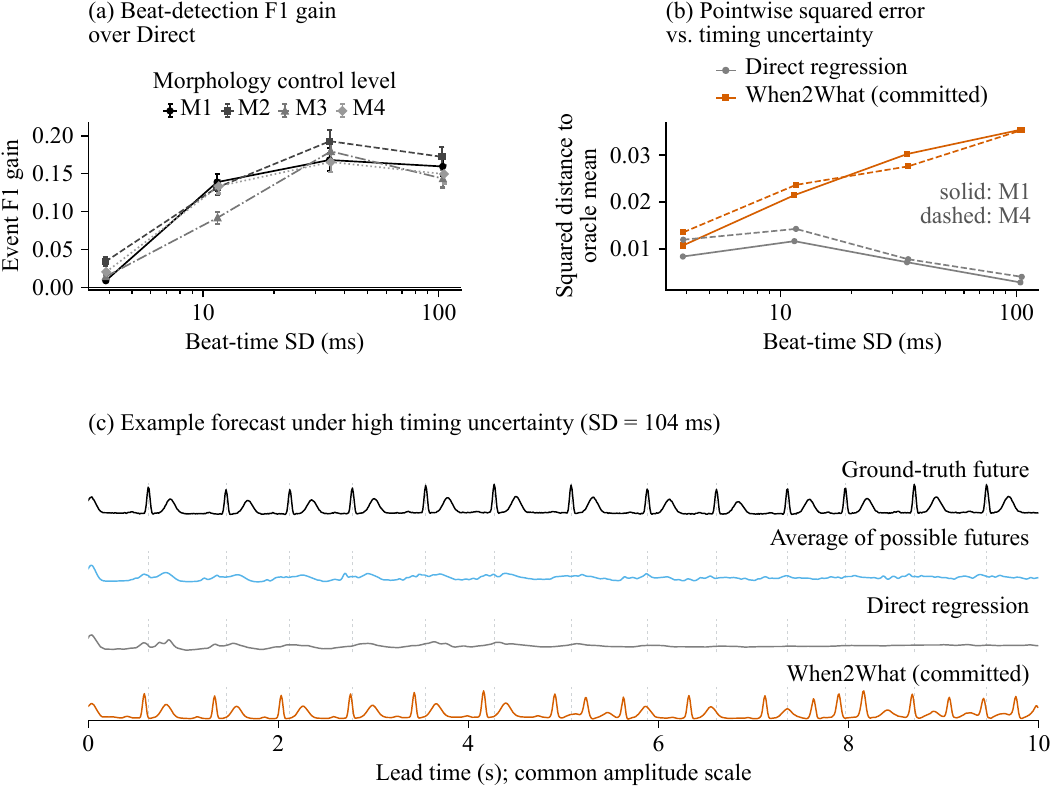}{\linewidth}
\caption{Synthetic verification of the trade-off between beat sharpness and pointwise squared error.
(a)~Improvement in beat-detection $F_1$ (\method{} minus Direct regression) as future beat-time uncertainty increases across four morphology variability levels (M1--M4).
(b)~Pointwise squared error relative to the average of possible futures, showing that preserving sharp peaks increases squared error as timing uncertainty grows.
(c)~Example forecast under high timing uncertainty: the average of possible futures and Direct regression both flatten out due to phase averaging, whereas \method{} preserves the sharp R peaks of the ground-truth future waveform.}\label{fig:synthetic}
\end{figure}

In Table~\ref{tab:native-waveform}, we observed that \method achieves much higher beat-detection accuracy than Direct regression while having a higher samplewise MSE.
To verify why this trade-off occurs, we conduct a controlled synthetic ECG experiment using the dynamical ECG generator of McSharry et al.~\cite{mcsharry2003} (full setup in Supplementary Section~S7, Table~S22).
In this simulation, we explicitly control how uncertain future heartbeat times are by varying the beat-time standard deviation across four levels from nearly deterministic (\NsynTimingSDlow~ms) to highly uncertain (\NsynTimingSDmid, \NsynTimingSDthree, and \NsynTimingSDhigh~ms), alongside four levels of beat-shape variability (M1--M4). %
For each test record, in addition to the actual \emph{ground-truth future} waveform, we simulate 64 possible future trajectories from the same history and average them to obtain the \emph{average of possible futures}, which represents the theoretical MSE-optimal waveform~\cite{gneiting2011}. %
We train Direct regression and our committed event-conditioned model on the same backbone and evaluate retained R-peak height, beat-detection $F_1$, and pointwise squared distance to this average waveform.

Figure~\ref{fig:synthetic} shows how beat accuracy and pointwise error diverge as timing uncertainty increases.
When future beat timing is nearly certain (\NsynTimingSDlow~ms jitter), future beats occur at almost fixed locations, so Direct regression does not blur the peaks and both models achieve similar beat-detection $F_1$, differing by only \NsynCommitGainLow (Fig.~\ref{fig:synthetic}a).
Once beat-time uncertainty increases to \NsynTimingSDmid--\NsynTimingSDhigh~ms, however, averaging across phase-shifted future heartbeats causes the average of possible futures itself to flatten out.
As shown in Fig.~\ref{fig:synthetic}c under high timing uncertainty, Direct regression closely tracks this flattened average waveform, retaining only \NsynAmpDirectB of the true R-peak height and reaching a beat $F_1$ of just \NsynDirectFoneB.
By contrast, conditioning on committed beat times preserves sharp R peaks (\NsynAmpCommitB of the true R-peak height) and boosts beat $F_1$ to \NsynCommitFoneB, yielding $F_1$ gains of \NsynCommitGainMid to \NsynCommitGainHigh across uncertainty levels (Fig.~\ref{fig:synthetic}a,c).
Because a sharp predicted peak that is shifted by even a few milliseconds from the true beat incurs squared error at both the predicted and true locations, our model exhibits a higher pointwise squared error than the flattened direct forecast across all settings (\NsynExcessCommitB versus \NsynExcessDirectB; Fig.~\ref{fig:synthetic}b and Supplementary Section~S7, Tables~S23–S24).
These results confirm that when future beat times are uncertain, committing to sharp beats at predicted locations preserves detectable cardiac events at the unavoidable mathematical cost of higher pointwise squared error.

\FloatBarrier
\section{Discussion}\label{sec:discussion}

\subsection{Rethinking evaluation metrics for cardiovascular waveform forecasting}\label{sec:explain}\label{sec:simple}

A standard practice in time-series forecasting is training and benchmarking models primarily with pointwise squared error (MSE).
Our study demonstrates that for quasi-periodic cardiovascular waveforms such as ECG, ABP, and PPG, relying on pointwise MSE is fundamentally misleading because low MSE actively conflicts with preserving cardiac events under timing uncertainty.
Because heartbeat intervals fluctuate naturally from beat to beat due to autonomic and respiratory regulation~\cite{mcsharry2003}, the timing uncertainty of future beats exceeds the width of an R peak ($\pm$50~ms) within a few seconds (Fig.~\ref{fig:asymmetry}a). %
Once this occurs, minimizing MSE forces a model to output the conditional mean over phase-shifted future trajectories~\cite{gneiting2011}, averaging peaks with baselines to produce a flat, featureless wave that achieves low MSE (\NmseDirectDisc{} for Direct) while destroying all beat-level information (Table~\ref{tab:native-waveform} and Fig.~\ref{fig:synthetic}c).
Conversely, a model such as \method that faithfully generates sharp heartbeats at predicted times inevitably incurs a double-shift squared-error penalty whenever a predicted peak is offset by even a few milliseconds.
Future benchmarks for cardiovascular waveform forecasting should therefore evaluate event-level accuracy alongside samplewise fidelity.
Preserving explicit beat events also enables practical clinical applications such as beat-synchronized device control: with a median event-prediction latency of 15.43\,ms (Supplementary Section~S2, Table~S3), When2What reduces mistimed beat triggers by 19.6\% under a 50-ms hardware execution delay in our retrospective VitalDB streaming simulation (Supplementary Section~S8, Table~S25). %

\subsection{Limitations and future directions}\label{sec:crosssignal}\label{sec:limitations}

Two natural boundaries of this study point to directions for future work.
First, because beat-to-beat interval variations accumulate over time, exact millisecond-level heartbeat alignment is inherently concentrated in the near-term horizon, and \method currently commits to a single deterministic beat schedule $\hat{E}$ per window to generate one representative waveform.
A promising extension is to combine our predict, commit, and generate paradigm with probabilistic generative models to sample multiple plausible future beat schedules and waveform trajectories, providing clinicians with calibrated uncertainty intervals.
Second, our experiments trained separate single-channel models for ECG, ABP, and PPG on quality-screened recordings (Table~\ref{tab:data}; additional sensitivity analyses across detectors, clustering units, and arrhythmia strata are reported in Supplementary Sections~S1--S8).
Because electrical cardiac activation and hemodynamic pulse propagation are physiologically coupled (Fig.~\ref{fig:anchors}), extending \method to jointly forecast multimodal cardiovascular signals in continuous, unscreened clinical streams is an exciting avenue for prospective deployment.

\section{Conclusions}\label{sec:conclusion}

In this study, we introduced \method, a timing-conditioned framework for long-horizon cardiovascular waveform forecasting.
Motivated by the temporal asymmetry between the rapid seconds-scale decay of beat-time predictability and the multi-minute stability of beat morphology, \method decouples forecasting into predicting a dense timing map (\textbf{Predict}, $P$), selecting and sharpening discrete beat events (\textbf{Commit}, $C$), and generating the continuous waveform conditioned on those events and the observed history (\textbf{Generate}, $G$).
Extensive experiments demonstrate that \method consistently outperforms existing time-series and physiological forecasters, reproduces over 95\% of its predicted heartbeats in the output waveform, and generalizes robustly across external hospital and arrhythmia cohorts as well as ABP and PPG pulse signals. %
Together, our findings establish explicit timing conditioning as an effective paradigm for preserving critical physiological events in long-horizon waveform forecasting.

\section*{CRediT authorship contribution statement}
\textbf{Chaoqi Zhang:} Conceptualization, Data curation, Formal analysis, Investigation, Methodology, Project administration, Software, Validation, Visualization, Writing -- original draft, Writing -- review \& editing.
\textbf{Shujie Yan:} Data curation, Investigation, Software, Validation.
\textbf{Zhao Zhang:} Data curation, Investigation, Validation.
\textbf{Yanhua Sun:} Data curation, Investigation.
\textbf{Yu Wang:} Investigation, Software, Validation.
\textbf{Li Li:} Conceptualization, Supervision.
\textbf{Lingzhong Meng:} Conceptualization, Supervision.
\textbf{Xiaoyi Chen:} Conceptualization, Methodology, Project administration, Supervision, Writing -- review \& editing.
\textbf{Haixu Tang:} Conceptualization, Funding acquisition, Project administration, Resources, Supervision, Writing -- review \& editing.

\section*{Declaration of competing interest}
The authors declare that they have no known competing financial interests or personal relationships that could have appeared to influence the work reported in this paper.

\section*{Funding}
This research did not receive any specific grant from funding agencies in the public, commercial, or not-for-profit sectors.

\section*{Ethics statement}
This study used previously collected, de-identified physiological waveform datasets from VitalDB, MOVER, and PhysioNet in accordance with their respective data-use requirements.
No new participant data were collected, and no identifiable patient information was accessed.

\section*{Data and code availability}
The physiological waveform datasets analyzed in this study are available from their original repositories, including VitalDB, MOVER, and PhysioNet, subject to their respective access conditions and data-use agreements~\cite{lee2022,samad2023,goldberger2000,eun2026arrhythmia,tsutsui2025,nsrdb,petrutiu2007}.
The code and reproduction scripts for \method are available at \url{https://github.com/prncoprs/when2what-release.git}.
Derived research artifacts will be shared in accordance with the applicable source licenses and data-use agreements.
Original third-party datasets will not be redistributed.

\section*{Declaration of generative AI and AI-assisted technologies in the manuscript preparation process}
During the preparation of this work, the authors used Claude and ChatGPT for assisting with data analysis, internal review and Gemini for polishing the paper. The authors reviewed and edited the output as needed and take full responsibility for the content of the published article.

\bibliographystyle{elsarticle-num}
\bibliography{refs}

\begin{thebibliography}{10}
\expandafter\ifx\csname url\endcsname\relax
  \def\url#1{\texttt{#1}}\fi
\expandafter\ifx\csname urlprefix\endcsname\relax\def\urlprefix{URL }\fi
\expandafter\ifx\csname href\endcsname\relax
  \def\href#1#2{#2} \def\path#1{#1}\fi

\bibitem{ASA2025Monitoring}
{American Society of Anesthesiologists},
  \href{https://www.asahq.org/standards-and-practice-parameters/standards-for-basic-anesthetic-monitoring}{Standards
  for basic anesthetic monitoring}, revised October 15, 2025; accessed October
  26, 2026 (2025).
\newline\urlprefix\url{https://www.asahq.org/standards-and-practice-parameters/standards-for-basic-anesthetic-monitoring}

\bibitem{lee2022}
H.-C. Lee, Y.~Park, S.~B. Yoon, S.~M. Yang, D.~Park, C.-W. Jung, {VitalDB,} a
  high-fidelity multi-parameter vital signs database in surgical patients,
  Scientific Data 9~(1) (2022) 279.
\newblock \href {https://doi.org/10.1038/s41597-022-01411-5}
  {\path{doi:10.1038/s41597-022-01411-5}}.

\bibitem{hatib2018}
F.~Hatib, Z.~Jian, S.~Buddi, C.~Lee, J.~Settels, K.~Sibert, J.~Rinehart,
  M.~Cannesson, Machine-learning algorithm to predict hypotension based on
  high-fidelity arterial pressure waveform analysis, Anesthesiology 129~(4)
  (2018) 663--674.
\newblock \href {https://doi.org/10.1097/aln.0000000000002300}
  {\path{doi:10.1097/aln.0000000000002300}}.

\bibitem{Jo2022WaveformHypotension}
Y.-Y. Jo, J.-H. Jang, J.-m. Kwon, H.-C. Lee, C.-W. Jung, S.~Byun, H.-G. Jeong,
  Predicting intraoperative hypotension using deep learning with waveforms of
  arterial blood pressure, electroencephalogram, and electrocardiogram:
  Retrospective study, PLOS ONE 17~(8) (2022) e0272055.
\newblock \href {https://doi.org/10.1371/journal.pone.0272055}
  {\path{doi:10.1371/journal.pone.0272055}}.

\bibitem{nie2023}
Y.~Nie, N.~H. Nguyen, P.~Sinthong, J.~Kalagnanam, A time series is worth 64
  words: Long-term forecasting with transformers, in: International Conference
  on Learning Representations (ICLR), 2023.

\bibitem{zacarias2024}
H.~Zacarias, J.~A.~L. Marques, V.~Felizardo, M.~Pourvahab, N.~M. Garcia, {ECG}
  forecasting system based on long {Short-Term} memory, Bioengineering 11~(1)
  (2024) 89.
\newblock \href {https://doi.org/10.3390/bioengineering11010089}
  {\path{doi:10.3390/bioengineering11010089}}.

\bibitem{li2026}
Z.~Li, Y.~Ouyang, J.~Tan, J.~Zhu, W.~Lu, R.~Xie, Y.~Xu, Z.~Gou, K.~Xie,
  {Clairvoyant-Net:} a lightweight and adaptive long sequence time-series
  forecasting framework for electrocardiogram, Knowledge-Based Systems 342
  (2026) 115916.
\newblock \href {https://doi.org/10.1016/j.knosys.2026.115916}
  {\path{doi:10.1016/j.knosys.2026.115916}}.

\bibitem{gneiting2011}
T.~Gneiting, Making and evaluating point forecasts, Journal of the American
  Statistical Association 106~(494) (2011) 746--762.
\newblock \href {https://doi.org/10.1198/jasa.2011.r10138}
  {\path{doi:10.1198/jasa.2011.r10138}}.

\bibitem{ruizbarroso2025}
P.~Ruiz-Barroso, F.~M. Castro, J.~Miranda, D.-A. Constantinescu, D.~Atienza,
  N.~Guil, {FADE:} forecasting for anomaly detection on {ECG}, Computer Methods
  and Programs in Biomedicine 267 (2025) 108780.
\newblock \href {http://arxiv.org/abs/2502.07389} {\path{arXiv:2502.07389}},
  \href {https://doi.org/10.1016/j.cmpb.2025.108780}
  {\path{doi:10.1016/j.cmpb.2025.108780}}.

\bibitem{hooi2017}
B.~Hooi, S.~Liu, A.~Smailagic, C.~Faloutsos, {BeatLex:} summarizing and
  forecasting time series with patterns, in: Machine Learning and Knowledge
  Discovery in Databases, Vol. 10535 of Lecture Notes in Computer Science,
  Springer International Publishing, 2017, pp. 3--19.
\newblock \href {https://doi.org/10.1007/978-3-319-71246-8_1}
  {\path{doi:10.1007/978-3-319-71246-8_1}}.

\bibitem{lin2026}
M.~Lin, X.~Zhang, Y.~Hong, Y.~Wen, X.~Lu, Q.~Wu, H.~Chen, C.~Chen,
  {RDiffGAN-ECG:} high-quality specific r-peaks conditional {ECG} generation
  using denoising diffusion generative adversarial network with state space
  models, Biomedical Signal Processing and Control 112 (2026) 108544.
\newblock \href {https://doi.org/10.1016/j.bspc.2025.108544}
  {\path{doi:10.1016/j.bspc.2025.108544}}.

\bibitem{neifar2024diffecg}
N.~Neifar, A.~Ben-Hamadou, A.~Mdhaffar, M.~Jmaiel,
  \href{https://arxiv.org/abs/2306.01875v3}{{DiffECG:} a versatile
  probabilistic diffusion model for {ECG} signals synthesis} (2024).
\newblock \href {http://arxiv.org/abs/2306.01875} {\path{arXiv:2306.01875}}.
\newline\urlprefix\url{https://arxiv.org/abs/2306.01875v3}

\bibitem{zhou2021}
H.~Zhou, S.~Zhang, J.~Peng, S.~Zhang, J.~Li, H.~Xiong, W.~Zhang, Informer:
  Beyond efficient transformer for long sequence {Time-Series} forecasting, in:
  AAAI Conference on Artificial Intelligence, Vol.~35, 2021, pp. 11106--11115.

\bibitem{zeng2023}
A.~Zeng, M.~Chen, L.~Zhang, Q.~Xu, Are transformers effective for time series
  forecasting?, Proceedings of the AAAI Conference on Artificial Intelligence
  37~(9) (2023) 11121--11128.
\newblock \href {http://arxiv.org/abs/2205.13504} {\path{arXiv:2205.13504}},
  \href {https://doi.org/10.1609/aaai.v37i9.26317}
  {\path{doi:10.1609/aaai.v37i9.26317}}.

\bibitem{zhang2026htf}
C.~Zhang, Y.~Wang, H.~Tang, From {HL} to {H+L-1} parameters: A
  {Hankel-Toeplitz} forecaster for {Long-Term} time series forecasting, arXiv
  preprint arXiv:2609.33984 (2026).
\newblock \href {https://doi.org/10.48550/arXiv.2609.33984}
  {\path{doi:10.48550/arXiv.2609.33984}}.

\bibitem{das2024}
A.~Das, W.~Kong, R.~Sen, Y.~Zhou, A decoder-only foundation model for
  time-series forecasting, in: International Conference on Machine Learning
  (ICML), Vol. 235 of Proceedings of Machine Learning Research, 2024, pp.
  10148--10167.

\bibitem{yaldiz2026}
C.~O. Yaldiz, D.~J. Lin, A.~H. Gazi, G.~Cestero, C.~Chen, B.~K. Bracken,
  A.~Winder, S.~Lynn, R.~Sameni, O.~T. Inan, {Real-Time} autoregressive
  forecast of cardiac features for psychophysiological applications, IEEE
  Journal of Biomedical and Health Informatics 30~(8) (2026) 7019--7030.
\newblock \href {https://doi.org/10.1109/jbhi.2025.3546148}
  {\path{doi:10.1109/jbhi.2025.3546148}}.

\bibitem{salinas2020}
D.~Salinas, V.~Flunkert, J.~Gasthaus, T.~Januschowski, {DeepAR:} probabilistic
  forecasting with autoregressive recurrent networks, International Journal of
  Forecasting 36~(3) (2020) 1181--1191.
\newblock \href {https://doi.org/10.1016/j.ijforecast.2019.07.001}
  {\path{doi:10.1016/j.ijforecast.2019.07.001}}.

\bibitem{rasul2021}
K.~Rasul, C.~Seward, I.~Schuster, R.~Vollgraf, Autoregressive denoising
  diffusion models for multivariate probabilistic time series forecasting, in:
  International Conference on Machine Learning (ICML), Vol. 139 of Proceedings
  of Machine Learning Research, 2021, pp. 8857--8868.

\bibitem{ren2021}
Y.~Ren, C.~Hu, X.~Tan, T.~Qin, S.~Zhao, Z.~Zhao, T.-Y. Liu, {FastSpeech} 2:
  Fast and {High-Quality} {End-to-End} text to speech, in: International
  Conference on Learning Representations (ICLR), 2021.

\bibitem{ciamarra2024}
A.~Ciamarra, F.~Becattini, L.~Seidenari, A.~Del~Bimbo, {FLODCAST}: Flow and
  depth forecasting via multimodal recurrent architectures, Pattern Recognition
  150 (2024) 110337.
\newblock \href {https://doi.org/10.1016/j.patcog.2024.110337}
  {\path{doi:10.1016/j.patcog.2024.110337}}.

\bibitem{mcsharry2003}
P.~McSharry, G.~Clifford, L.~Tarassenko, L.~Smith, A dynamical model for
  generating synthetic electrocardiogram signals, IEEE Transactions on
  Biomedical Engineering 50~(3) (2003) 289--294.
\newblock \href {https://doi.org/10.1109/tbme.2003.808805}
  {\path{doi:10.1109/tbme.2003.808805}}.

\bibitem{kim2026}
J.~Kim, K.~Park, L.~Chen, K.~Kim, Shape and amplitude decoupling in pulsatile
  physiological signal synthesis and its evaluation, Nature Communications
  17~(1) (2026) 5876.
\newblock \href {https://doi.org/10.1038/s41467-026-72299-7}
  {\path{doi:10.1038/s41467-026-72299-7}}.

\bibitem{leguen2019}
V.~Le~Guen, N.~Thome, Shape and time distortion loss for training deep time
  series forecasting models, in: Advances in Neural Information Processing
  Systems (NeurIPS), Vol.~32, 2019, pp. 4189--4201.

\bibitem{cuturi2017}
M.~Cuturi, M.~Blondel, {Soft-DTW:} a differentiable loss function for
  {Time-Series}, in: International Conference on Machine Learning (ICML),
  Vol.~70 of Proceedings of Machine Learning Research, 2017, pp. 894--903.

\bibitem{murphy1988}
A.~H. Murphy, Skill scores based on the mean square error and their
  relationships to the correlation coefficient, Monthly Weather Review 116~(12)
  (1988) 2417--2424.
\newblock \href
  {https://doi.org/10.1175/1520-0493(1988)116<2417:ssbotm>2.0.co;2}
  {\path{doi:10.1175/1520-0493(1988)116<2417:ssbotm>2.0.co;2}}.

\bibitem{jolliffe2012}
I.~T. Jolliffe, D.~B. Stephenson (Eds.), Forecast Verification: A
  Practitioner's Guide in Atmospheric Science, 2nd Edition, Wiley, 2011.
\newblock \href {https://doi.org/10.1002/9781119960003}
  {\path{doi:10.1002/9781119960003}}.

\bibitem{pantompkins1985}
J.~Pan, W.~J. Tompkins, A {Real-Time} {QRS} detection algorithm, IEEE
  Transactions on Biomedical Engineering BME-32~(3) (1985) 230--236.
\newblock \href {https://doi.org/10.1109/tbme.1985.325532}
  {\path{doi:10.1109/tbme.1985.325532}}.

\bibitem{makowski2021}
D.~Makowski, T.~Pham, Z.~J. Lau, J.~C. Brammer, F.~Lespinasse, H.~Pham,
  C.~Sch{\"o}lzel, S.~H.~A. Chen, {NeuroKit2:} a python toolbox for
  neurophysiological signal processing, Behavior Research Methods 53~(4) (2021)
  1689--1696.
\newblock \href {https://doi.org/10.3758/s13428-020-01516-y}
  {\path{doi:10.3758/s13428-020-01516-y}}.

\bibitem{gu2023}
A.~Gu, T.~Dao, Mamba: {Linear-Time} sequence modeling with selective state
  spaces, arXiv preprint arXiv:2312.00752 (2023).
\newblock \href {https://doi.org/10.48550/arXiv.2312.00752}
  {\path{doi:10.48550/arXiv.2312.00752}}.

\bibitem{samad2023}
M.~Samad, M.~Angel, J.~Rinehart, Y.~Kanomata, P.~Baldi, M.~Cannesson, Medical
  informatics operating room vitals and events repository {(MOVER):} a
  public-access operating room database, JAMIA Open 6~(4) (2023) ooad084.
\newblock \href {https://doi.org/10.1093/jamiaopen/ooad084}
  {\path{doi:10.1093/jamiaopen/ooad084}}.

\bibitem{tsutsui2025}
K.~Tsutsui, S.~B. Brimer, N.~Ben-Moshe, J.~M. Sellal, J.~Oster, H.~Mori,
  Y.~Ikeda, T.~Arai, S.~Nakano, R.~Kato, J.~A. Behar, {SHDB-AF:} a japanese
  holter {ECG} database of atrial fibrillation, Scientific Data 12~(1) (2025)
  454.
\newblock \href {https://doi.org/10.1038/s41597-025-04777-4}
  {\path{doi:10.1038/s41597-025-04777-4}}.

\bibitem{zong2003}
W.~Zong, T.~Heldt, G.~B. Moody, R.~G. Mark, An open-source algorithm to detect
  onset of arterial blood pressure pulses, in: Computers in Cardiology, 2003,
  2003, pp. 259--262.
\newblock \href {https://doi.org/10.1109/CIC.2003.1291140}
  {\path{doi:10.1109/CIC.2003.1291140}}.

\bibitem{elgendi2013}
M.~Elgendi, I.~Norton, M.~Brearley, D.~Abbott, D.~Schuurmans, Systolic peak
  detection in acceleration photoplethysmograms measured from emergency
  responders in tropical conditions, PLoS ONE 8~(10) (2013) e76585.
\newblock \href {https://doi.org/10.1371/journal.pone.0076585}
  {\path{doi:10.1371/journal.pone.0076585}}.

\bibitem{goldberger2000}
A.~L. Goldberger, L.~A.~N. Amaral, L.~Glass, J.~M. Hausdorff, P.~C. Ivanov,
  R.~G. Mark, J.~E. Mietus, G.~B. Moody, C.-K. Peng, H.~E. Stanley,
  {PhysioBank, PhysioToolkit, and PhysioNet}: Components of a new research
  resource for complex physiologic signals, Circulation 101~(23) (2000)
  e215--e220.
\newblock \href {https://doi.org/10.1161/01.cir.101.23.e215}
  {\path{doi:10.1161/01.cir.101.23.e215}}.

\bibitem{eun2026arrhythmia}
D.~Eun, K.~Shim, H.~Lee, Y.~Lim, H.~Lim, H.~Lee, H.-C. Lee, {VitalDB Arrhythmia
  Database: An Anesthesiologist-Validated Large-Scale Intraoperative Arrhythmia
  Dataset with Beat and Rhythm Labels}, [dataset], version 1.0.0 (2026).
\newblock \href {https://doi.org/10.13026/axd6-wm13}
  {\path{doi:10.13026/axd6-wm13}}.

\bibitem{nsrdb}
{The Beth Israel Deaconess Medical Center, The Arrhythmia Laboratory}, {The
  MIT-BIH Normal Sinus Rhythm Database}, PhysioNet, [dataset], version 1.0.0
  (1990).
\newblock \href {https://doi.org/10.13026/C2NK5R} {\path{doi:10.13026/C2NK5R}}.

\bibitem{petrutiu2007}
S.~Petrutiu, A.~V. Sahakian, S.~Swiryn, Abrupt changes in fibrillatory wave
  characteristics at the termination of paroxysmal atrial fibrillation in
  humans, EP Europace 9~(7) (2007) 466--470.
\newblock \href {https://doi.org/10.1093/europace/eum096}
  {\path{doi:10.1093/europace/eum096}}.

\bibitem{ltafdbdata}
G.~Moody, S.~Swiryn, {Long Term AF Database}, PhysioNet, [dataset], version
  1.0.0 (2008).
\newblock \href {https://doi.org/10.13026/C2QG6Q} {\path{doi:10.13026/C2QG6Q}}.

\bibitem{moody2001}
G.~Moody, R.~Mark, The impact of the {MIT-BIH} arrhythmia database, IEEE
  Engineering in Medicine and Biology Magazine 20~(3) (2001) 45--50.
\newblock \href {https://doi.org/10.1109/51.932724}
  {\path{doi:10.1109/51.932724}}.

\end{thebibliography}

\clearpage
\phantomsection
\addcontentsline{toc}{section}{Supplementary Material}
\section*{Supplementary Material}
\setcounter{section}{0}
\setcounter{subsection}{0}
\setcounter{table}{0}
\setcounter{figure}{0}
\setcounter{equation}{0}
\renewcommand{\thesection}{S\arabic{section}}
\renewcommand{\thetable}{S\arabic{table}}
\renewcommand{\thefigure}{S\arabic{figure}}
\renewcommand{\theequation}{S\arabic{equation}}
\renewcommand{\theHsection}{supp.\arabic{section}}
\renewcommand{\theHtable}{supp.\arabic{table}}
\renewcommand{\theHfigure}{supp.\arabic{figure}}
\renewcommand{\theHequation}{supp.\arabic{equation}}
\section*{Overview}
This supplementary document is organized to follow the progression of the main manuscript:
\begin{itemize}
    \item \textbf{Section~S1 (Evaluation metrics and predictability time scale):} Mathematical definitions of chronological event matching, the random-phase chance baseline $F_1^{\mathrm{c}}$, chance-adjusted event skill $S$, and exponential decay fits for the predictability time scale $\tau$ across heart-rate-variability strata and awake cohorts (supporting Section~3.1 of the main text).
    \item \textbf{Section~S2 (Network architecture, baseline rules, and computational cost):} Layer-by-layer specifications of the shared autoencoder, spectral frequency gate, Mamba--Attention backbone, discrete event commitment rule $C$, non-learned baseline rules, and GPU latency and memory benchmarks across batch sizes (supporting Sections~3.2--3.4 of the main text).
    \item \textbf{Section~S3 (Dataset preprocessing, reference construction, and statistical protocol):} Causal resampling, multi-detector ECG reference fusion, quality-screening characteristics, blinded human verification, ABP and PPG pulse-anchor verification, and bootstrap protocols (supporting Section~4.1 of the main text).
    \item \textbf{Section~S4 (Complete comparison with forecasting baselines, RQ1):} Full 30-row comparison table with 95\% confidence intervals and hyperparameter sensitivity variants across all baseline families (supporting Section~4.2 of the main text).
    \item \textbf{Section~S5 (Extended ablation study and beat-level correspondence, RQ2):} Full confidence intervals for conditioning and component ablations, Lockbox replications, bin-by-bin 60-s horizon tables, matching-tolerance sensitivity, Shapley error decomposition, and beat-level correspondence tables (supporting Section~4.3 of the main text).
    \item \textbf{Section~S6 (Extended generalization analyses, RQ3):} Complete confidence intervals across losses and backbones, Split-MSE tuning grid, external hospital (MOVER) and atrial fibrillation (SHDB-AF all-rhythm and active-AF) cohorts, ABP and PPG pulse waveforms, and waveform detector sensitivity (supporting Section~4.4 of the main text).
    \item \textbf{Section~S7 (Complete synthetic experiment grid, RQ4):} Generator parameters and full 16-cell results with 95\% confidence intervals for committed and soft-map inference (supporting Section~4.5 of the main text).
    \item \textbf{Section~S8 (Retrospective beat-synchronized device triggering analysis):} First-beat timing and streaming hardware-delayed trigger simulations across cohorts and delay settings (supporting Section~5.1 of the main text).
\end{itemize}
\setcounter{section}{0}

\section{Evaluation metrics and predictability time scale}\label{sec:s-metrics}

\paragraph{Mathematical formulation of event matching and chance-adjusted skill}
We evaluate beat-timing accuracy along two evaluation paths: the event channel, which scores the discrete beat times $\hat{E} = \{\hat{e}_1, \dots, \hat{e}_{\hat{K}}\}$ output by the timing predictor $P$ and commit step $C$, and the waveform channel, which scores the beat times extracted from the generated continuous waveform $\hat{\mathbf{y}}$ by the standard NeuroKit2 detector~\cite{makowski2021}.
In both channels, predicted events are sorted chronologically ($\hat{e}_1 < \dots < \hat{e}_{\hat{K}}$) and paired greedily one-to-one with the ground-truth reference events $E = \{e_1, \dots, e_K\}$ over the full 60-s forecast horizon.
Specifically, each predicted event $\hat{e}_j$ is matched to the closest unmatched reference event $e_k$ if and only if $| \hat{e}_j - e_k | \le \delta$, where $\delta = 50$~ms is the primary matching tolerance (with $\delta \in \{25, 100\}$~ms evaluated in sensitivity analyses).

After full-horizon matching, events are partitioned into seven disjoint lead-time bins $b$: $[0,2)$, $[2,5)$, $[5,10)$, $[10,20)$, $[20,30)$, $[30,45)$, and $[45,60)$~s.
Matched pairs (true positives, $\mathrm{TP}$) and unmatched reference events (false negatives, $\mathrm{FN}$) are assigned to bin $b$ according to their reference timestamp $e_k \in b$, whereas unmatched predicted events (false positives, $\mathrm{FP}$) are assigned to bin $b$ according to their predicted timestamp $\hat{e}_j \in b$.
For a given window $w$, training seed $s$, and bin $b$, the bin-level $F_1$ score is
\begin{equation}\label{eq:s-f1}
F_1(w, s, b) = \frac{2\,\mathrm{TP}(w, s, b)}{2\,\mathrm{TP}(w, s, b) + \mathrm{FP}(w, s, b) + \mathrm{FN}(w, s, b)},
\end{equation}
with $F_1(w, s, b) = 0$ when the denominator is zero.
To account for differences in heart rate across windows, we compute the window- and bin-specific chance baseline $F_1^{\mathrm{c}}(w, b)$ as the expected $F_1$ score of a constant-rate periodic pulse train $E_{\mathrm{met}}(\phi) = \{\phi + m\,\overline{\mathrm{RR}}_w : m \in \mathbb{Z}_{\ge 0}\} \cap [0, 60)$~s whose period equals the window's true mean beat interval $\overline{\mathrm{RR}}_w$ and whose starting phase $\phi$ is drawn uniformly from $[0, \overline{\mathrm{RR}}_w)$:
\begin{equation}\label{eq:s-chance}
F_1^{\mathrm{c}}(w, b) = \frac{1}{\overline{\mathrm{RR}}_w}\int_{0}^{\overline{\mathrm{RR}}_w} F_1\bigl(E_{\mathrm{met}}(\phi), E_w, b\bigr)\,\mathrm{d}\phi.
\end{equation}
Within each window $w$ and bin $b$, $F_1(w, s, b)$ is first averaged across the $S_{\mathrm{seed}}$ random training seeds to obtain $\bar{F}_1(w, b) = \frac{1}{S_{\mathrm{seed}}}\sum_{s} F_1(w, s, b)$.
We then average $\bar{F}_1(w, b)$ and $F_1^{\mathrm{c}}(w, b)$ across all $N_{\mathrm{win}}$ windows in the cohort to obtain $\bar{F}_{1,b}$ and $\bar{F}_{1,b}^{\mathrm{c}}$, and compute the chance-adjusted event skill in bin $b$:
\begin{equation}\label{eq:s-skill}
S_b = \frac{\bar{F}_{1,b} - \bar{F}_{1,b}^{\mathrm{c}}}{1 - \bar{F}_{1,b}^{\mathrm{c}}}.
\end{equation}
The primary near-term event skill over the first 10~s is the unweighted mean across the first three bins:
\begin{equation}\label{eq:s-primary}
S_{0\text{--}10} = \frac{1}{3}\bigl(S_{[0,2)} + S_{[2,5)} + S_{[5,10)}\bigr).
\end{equation}

\paragraph{Predictability time scale $\tau$ across physiological states}
To quantify how rapidly beat-time predictability decays with lead time $t$, we compute the event skill $S$ in thirty 2-s lead-time bins over 0--60~s and fit an exponential decay curve, $a\,e^{-t/\tau}+c$, using bounded least squares.
Confidence intervals for $\tau$ are obtained by refitting the exponential curve on each of 1,000 shared patient-cluster bootstrap replicates.
Table~\ref{tab:s-tau} summarizes the fitted decay time constants $\tau$ across predictors, heart-rate-variability strata quantified by the root mean square of successive RR differences (RMSSD), and ambulatory awake sinus recordings from the MIT-BIH Normal Sinus Rhythm Database (NSRDB)~\cite{nsrdb,goldberger2000} and the Long Term AF Database (LTAFDB)~\cite{petrutiu2007,ltafdbdata}.
As shown in Table~\ref{tab:s-tau}, beat-time predictability decays within a few seconds across all settings, with shorter time scales observed when heart-rate variability is higher or when subjects are awake rather than under general anesthesia.

\begin{table}[!htbp]
\caption{Fitted beat-time predictability time scale $\tau$ across predictors, heart-rate-variability strata, and cohorts.}\label{tab:s-tau}
\centering
\begin{tabularx}{\linewidth}{@{}>{\raggedright\arraybackslash}X>{\raggedright\arraybackslash}Xcc@{}}
\toprule
Cohort and physiological stratum & Predictor & Fitted $\tau$ (s) & 95\% CI (s) \\
\midrule
VitalDB ECG test cohort (all windows) & Simple RR extrapolation & \NtauCNDisc & [6.85, 7.70] \\
VitalDB ECG test cohort (all windows) & \method{} timing head ($P$) & \NtauWWDisc & [7.02, 7.70] \\
VitalDB ECG Lockbox cohort (all windows) & Simple RR extrapolation & \NhypOne & [6.49, 7.32] \\
VitalDB ECG test, lowest RMSSD quintile (most regular) & \method{} timing head ($P$) & \NtauRmssdLow & [8.16, 9.60] \\
VitalDB ECG test, highest RMSSD quintile (most variable) & \method{} timing head ($P$) & \NtauRmssdHigh & [5.83, 7.13] \\
External NSRDB (ambulatory awake sinus rhythm) & \method{} timing head ($P$) & \NtauNsrdb & [1.56, 2.17] \\
External LTAFDB (ambulatory awake sinus segments) & \method{} timing head ($P$) & \NtauLtafSinus & [3.16, 4.04] \\
\bottomrule
\end{tabularx}
\end{table}

\section{Network architecture, baseline rules, and computational cost}\label{sec:s-training}

\paragraph{Architectural details and training configuration}
Table~\ref{tab:s-train} summarizes the architectural and optimization hyperparameters shared by \method{}, Direct, and Direct-MT on ECG.
In the shared convolutional autoencoder ($\mathrm{Enc}, \mathrm{Dec}$), three strided 1D convolutional blocks with strides $10 \times 10 \times 10$ compress the 120-s history ($H = 60{,}000$ samples at 500~Hz) into 60 tokens of 2~s each, while a symmetric transposed-convolutional decoder maps tokens back to 500-Hz waveforms.
To initialize the 30 horizon query tokens with frequency-domain context, we compute the discrete Fourier transform magnitudes of the history below 10~Hz, weight them via a learnable sigmoid frequency gate, and project them to the 512-dimensional token space.
In the Generate ($G$) stage, the committed timing map $\tilde{m}(t)$ is downsampled by a strided convolutional stem and added to the 30 horizon query tokens prior to the Mamba--Attention backbone for coarse token-level alignment, and is concatenated with the upsampled decoder features in a final 1D convolutional layer at 500~Hz for fine sample-level anchoring.
In the TCN backbone variant, the bidirectional Mamba--Attention stack in both the predictor $P$ and generator $G$ is replaced with a causal dilated temporal convolutional network while keeping all other settings fixed (three random seeds for TCN Direct and five random seeds each for TCN Direct-MT and TCN \method{}).
The ABP and PPG models use the same architecture and training recipe with signal-specific preprocessing and three random seeds per model.

\begin{table}[!htbp]
\caption{Architecture and training configuration shared by the matched ECG models.}\label{tab:s-train}
\centering
\begin{tabularx}{\linewidth}{@{}>{\raggedright\arraybackslash}X>{\raggedright\arraybackslash}X@{}}
\toprule
Component & Specification \\
\midrule
Input and output & 120-s history ($H = 60{,}000$), 60-s horizon ($L = 30{,}000$) at 500~Hz, z-scored per window \\
Autoencoder ($\mathrm{Enc}, \mathrm{Dec}$) & Convolutional autoencoder, strides $10\times10\times10$ (2-s tokens), channels 128 and 256 \\
Spectral embedding & Historical Fourier magnitudes below 10~Hz with a learned frequency gate \\
Sequence backbone & 4 layers, width 512, bidirectional Mamba blocks~\cite{gu2023} with self-attention every second layer \\
Timing target $m(t)$ & Unit-height Gaussian bumps, $\sigma = 10$~ms, truncated at $\pm 4\sigma$ and summed across events \\
Timing loss $\mathcal{L}_{\mathrm{time}}$ & Peak-weighted MSE ($w = 50$) plus soft overlap $F_1$ loss ($\lambda_{\mathrm{F}} = 0.5$) \\
Total loss weights & Timing loss $1.0$, waveform loss $1.0$, history reconstruction $\lambda_{\mathrm{h}} = 0.3$; finite-difference weights $1.0$ \\
Split-MSE variant & Inner amplitude band $|y|\le 1.6$ in history z-score units, inner-to-outer weight ratio $5 : 1$ \\
Optimizer & AdamW, peak learning rate $3\times10^{-4}$, weight decay $10^{-4}$, gradient clipping $1.0$ \\
Training schedule & 3,000 warm-up steps, cosine decay over 60,000 steps, batch size 24, bf16 precision \\
Checkpoint selection & Lowest validation loss evaluated every 200 steps; 5 random seeds for ECG, 3 for pulse models \\
\bottomrule
\end{tabularx}
\end{table}

\paragraph{Discrete event commitment rule ($C$)}
Both \method{} and Direct-MT apply the same non-neural commit operator $C$ to select discrete beat times $\hat{E} = C(\hat{m})$ from the predicted timing map $\hat{m}(t)$.
Candidate peaks are identified as local maxima of $\hat{m}(t)$ that exceed a local 90th-percentile threshold computed over a 5-s sliding window, subject to a minimum threshold floor of $0.15$.
To enforce physiological refractory spacing, historical RR intervals outside $[0.25, 2.0]$~s are discarded and the median of the last five remaining historical RR intervals is computed.
If fewer than one eligible interval remains, the historical RR estimate defaults to $0.8$~s.
The minimum allowed spacing between committed events is set to $0.5\times$ this historical RR estimate and clamped to the range $[0.125, 1.0]$~s.

\paragraph{Formulation of non-learned cadence and template baselines}
To evaluate the individual stages of \method{} against non-learned physiological heuristics, we define two rule-based beat-time extrapolators and one non-learned waveform constructor:
\begin{enumerate}
    \item \textbf{Simple RR extrapolation:} Causal R peaks (or pulse anchors) are detected in the 120-s history $\mathbf{x}$. Successive intervals outside physiologically plausible bounds ($[0.25, 2.0]$~s) are excluded, and the median of the last five valid intervals, $\widehat{\mathrm{RR}}_{\mathrm{med}}$, is projected forward from the final historical beat time $e_{\mathrm{last}} \le 0$ as $\hat{e}_k = e_{\mathrm{last}} + k\,\widehat{\mathrm{RR}}_{\mathrm{med}}$ over the 60-s horizon.
    \item \textbf{Guarded cadence rule:} Extends simple RR extrapolation by applying an outlier filter to the recent historical beats to guard against premature ectopic contractions or missed historical detections before computing the extrapolation anchor and cadence period.
    \item \textbf{Static historical beat template:} Extracts the last 40 complete heartbeat segments from the causal history $\mathbf{x}$, spanning $[-250, 450]$~ms around each historical R peak, and computes their pointwise median waveform segment. To construct a 60-s forecast at a given set of predicted beat times $\hat{E}$, this median beat segment is placed at each $\hat{e}_k \in \hat{E}$; overlapping segments between closely spaced beats are joined at their temporal midpoint, and intervals between separated beats are filled with the template's isoelectric level (the median of the first and last 30~ms of the template).
\end{enumerate}

\paragraph{Computational cost, GPU memory footprint, and latency scaling}
The matched ECG models (\method{}, Direct, and Direct-MT) have identical trainable parameter counts of $24.88$~M parameters.
Table~\ref{tab:s-latency} reports the stage-by-stage median single-window inference latency measured across 1,000 timed runs after 50 warm-up runs on an NVIDIA H200 NVL MIG 1g.18gb slice.
Table~\ref{tab:s-memory} further compares the forward-pass GPU latency and peak GPU memory footprint of the two-pass \method{} architecture against single-pass Direct regression at batch size~1 and batch size~32 across two GPU slice configurations (MIG 1g.18gb and MIG 4g.71gb).
Because $P$ and $G$ share the same encoder and Mamba--Attention weights, \method{} adds less than $6\%$ peak GPU memory ($429$~MB versus $404$~MB at batch size~1; $699$~MB versus $678$~MB at batch size~32) and achieves virtually identical GPU throughput to Direct regression ($3.3$~ms versus $3.4$~ms at batch size~1, and $18.7$~ms for both models at batch size~32 on MIG 4g.71gb).

\begin{table}[!htbp]
\caption{Stage-by-stage median single-window streaming latency for a 60-s forecast (1,000 timed runs after 50 warm-up runs on an NVIDIA H200 NVL MIG 1g.18gb slice).}\label{tab:s-latency}
\centering
\begin{tabularx}{\linewidth}{@{}>{\raggedright\arraybackslash}X>{\raggedright\arraybackslash}Xc@{}}
\toprule
Pipeline stage & Operation description & Median latency (ms) \\
\midrule
Preprocessing & Window buffering and z-score normalization & 2.61 \\
History RR estimation & Causal R-peak detection on historical window & 1.93 \\
Predict ($P$) & First network pass (encoder, Mamba--Attention backbone, timing head) & 2.23 \\
Commit ($C$) & Local-threshold peak selection, refractory filtering, and Gaussian encoding $\tilde{m}$ & 8.90 \\
\cmidrule(lr){1-3}
\multicolumn{2}{@{}l}{\emph{Subtotal: Predict ($P$) + Commit ($C$) event-prediction chain}} & \textbf{15.43} \\
\midrule
Generate ($G$) & Second network pass (conditioned backbone, decoder, and 500-Hz fusion) & \textbf{4.43} \\
\bottomrule
\end{tabularx}
\end{table}

\begin{table}[!htbp]
\caption{Comparison of network forward-pass GPU latency and peak GPU memory footprint between Direct and \method{} (24.88\,M parameters each) at batch sizes 1 and 32.}\label{tab:s-memory}
\centering
{\small\setlength{\tabcolsep}{5pt}%
\begin{tabular}{@{}llcccc@{}}
\toprule
Hardware slice & Model & \makecell{Batch 1\\ latency (ms)} & \makecell{Batch 1\\ memory (MB)} & \makecell{Batch 32\\ latency (ms)} & \makecell{Batch 32\\ memory (MB)} \\
\midrule
H200 MIG 1g.18gb & Direct & 6.0 & 404 & 76.7 & 678 \\
H200 MIG 1g.18gb & \method{} (Ours) & 6.0 & 429 & 76.8 & 699 \\
\midrule
H200 MIG 4g.71gb & Direct & 3.4 & 404 & 18.7 & 678 \\
H200 MIG 4g.71gb & \method{} (Ours) & 3.3 & 429 & 18.7 & 699 \\
\bottomrule
\end{tabular}}
\end{table}

\section{Dataset preprocessing, reference construction, and statistical protocol}\label{sec:s-reference}\label{sec:s-pulse}\label{sec:s-hypotheses}

\paragraph{Dataset preprocessing}
All VitalDB~\cite{lee2022} recordings (ECG, ABP, and PPG) are sampled at 500~Hz.
External MOVER~\cite{samad2023} intraoperative ECG recordings are causally resampled from 120~Hz to 500~Hz using past samples only so that no future information enters the resampling filter.
The training-hours total reported in Table~1 of the main text (\NtrainHours{} archived hours across \NtrainPat{} patients, comprising \NtrainSeg{} lead-II segments; \NppgTrainCases{} PPG training cases and \NabpTrainCases{} ABP training cases) reflects all archived training segments assigned to the training split prior to window extraction.

\paragraph{ECG reference construction, quality screening, and blinded human verification}
To construct a high-reliability evaluation reference for VitalDB ECG, candidate R peaks from three detectors run on leads~II and V5 are merged, conflicts are resolved by local signal quality around each beat, and any single-lead gaps are filled from the complementary lead.
When synchronous finger PPG is available, it adjudicates beat existence offline without setting R-peak timestamps or entering ECG models.
Validated against manually reviewed annotations on external benchmarks, this multi-detector fusion pipeline achieved an $F_1$ score of \NexpMitF{} on MIT-BIH~\cite{moody2001} and \NexpLtafF{} on LTAFDB~\cite{ltafdbdata}.

During quality screening, every 10-s block of the 60-s forecast horizon must pass a NeuroKit2 signal-quality index threshold of $0.4$ and artifact checks for missing samples, spikes, saturation, and flat lines in at least one lead.
Table~\ref{tab:s-gate} summarizes the exclusion rates and rhythm characteristics of included versus excluded windows, and Table~\ref{tab:s-human-check} summarizes the blinded two-annotator human verification conducted across \NrefCheckedAll{} reference peaks.

\begin{table}[!htbp]
\caption{Summary of reference-quality screening across cohorts and rhythm characteristics of included versus excluded windows.}\label{tab:s-gate}
\centering
{\small\setlength{\tabcolsep}{4.5pt}%
\begin{tabularx}{\linewidth}{@{}>{\raggedright\arraybackslash}X>{\raggedright\arraybackslash}Xccc@{}}
\toprule
Cohort or annotated subset & Metric & Before screen & Included & Excluded \\
\midrule
VitalDB ECG validation cohort & Windows / patients & \NwinValPreGate{} / \NecgValPatPreGate{} & \NwinVal{} / \NpatVal{} & \NexclVal{} \\
VitalDB ECG test cohort & Windows / patients & \NwinDiscPreGate{} / \NecgDiscPatPreGate{} & \NwinDisc{} / \NpatDisc{} & \NexclDisc{} \\
VitalDB ECG Lockbox cohort & Windows / patients & \NwinLbxPreGate{} / \NecgLbxPatPreGate{} & \NwinLbx{} / \NpatLbx{} & \NexclLbx{} \\
External MOVER ECG cohort & Windows / patients & \NmoverWinPreGate{} / \NmoverPatPreGate{} & \NmoverWin{} / \NmoverPat{} & 3.1\% \\
\midrule
VitalDB test tachycardia rate & Share with HR $> 100$\,bpm & -- & \NgateHrInclDisc{} & \NgateHrExclDisc{} \\
VitalDB Arrhythmia DB~\cite{eun2026arrhythmia} & AF or atrial flutter windows & -- & \NgateRhyIncludedAf{} / \NgateRhyIncludedWin{} & \NgateRhyExcludedAf{} / \NgateRhyExcludedWin{} \\
VitalDB Arrhythmia DB~\cite{eun2026arrhythmia} & Atrioventricular block windows & -- & \NgateRhyIncludedAvb{} / \NgateRhyIncludedWin{} & \NgateRhyExcludedAvb{} / \NgateRhyExcludedWin{} \\
\bottomrule
\end{tabularx}}
\end{table}

\begin{table}[!htbp]
\caption{Blinded two-annotator human verification and residual error bounds of the fused ECG reference.}\label{tab:s-human-check}
\centering
{\small\setlength{\tabcolsep}{5pt}%
\begin{tabularx}{\linewidth}{@{}>{\raggedright\arraybackslash}X>{\raggedright\arraybackslash}Xc@{}}
\toprule
Verification stratum & Inspected scope or prevalence & Observed error rate [95\% CI] \\
\midrule
Final-round newly sampled windows & \NrefCheckPeaks{} peaks (\NrefAgreement{} inter-annotator agreement) & 0.00\% [0.00\%, \NrefCheckErrRateHi{}] \\
All verification rounds combined & \NrefCheckedAll{} peaks inspected (\NrefErrAll{} residual errors) & 0.056\% \\
Beats without synchronous PPG adjudication & \NrefNoArb{}\% of reference beats & \NrefNoArbErrFull{} \\
Beats flagged as borderline by rule & \NrefSuspKept{}\% of reference beats & \NrefSuspErrFull{} \\
Beats at refinement window boundary & \NrefEdge{}\% of reference beats & -- \\
\bottomrule
\end{tabularx}}
\end{table}

\paragraph{ABP and PPG pulse-anchor verification}
For arterial blood pressure (ABP) and photoplethysmogram (PPG) waveforms, reference events are maximal-upslope pulse anchors extracted using the slope-sum onset detector~\cite{zong2003} and the Elgendi detector~\cite{elgendi2013}.
We verified these pulse anchors against synchronous ECG reference R peaks across the evaluable portions of the forecast horizon, covering \NppgRefCoverage{} of the horizon for PPG and \NabpRefCoverage{} for ABP.
Each pulse anchor was matched one-to-one to an ECG R peak within $\pm 100$~ms of the patient-specific mode R-to-anchor delay.
Table~\ref{tab:s-pulse-ref} summarizes the correspondence between synchronous ECG R peaks and ABP/PPG pulse anchors, confirming that over 98.8\% of ECG beats have a matching pulse anchor in both signals.

\begin{table}[!htbp]
\caption{Verification of ABP and PPG maximal-upslope pulse anchors against synchronous ECG reference R peaks over the first 10~s.}\label{tab:s-pulse-ref}
\centering
{\small\setlength{\tabcolsep}{6pt}%
\begin{tabular}{@{}lcc@{}}
\toprule
Verification metric & PPG (PLETH) & ABP (ART) \\
\midrule
Evaluable windows / patients & \NppgWin{} / \NppgPat{} & \NabpWin{} / \NabpPat{} \\
Unmatched ECG beats (missed anchor rate) [95\% CI] & \NppgRefMissedFull & \NabpRefMissedFull \\
Unmatched pulse anchors (extra anchor rate) [95\% CI] & \NppgRefExtraFull & \NabpRefExtraFull \\
Median R-to-anchor physiological delay (ms) & \NppgRefDelay & \NabpRefDelay \\
Interquartile range of R-to-anchor delay (ms) & \NppgRefDelayQlo{}--\NppgRefDelayQhi{} & \NabpRefDelayQlo{}--\NabpRefDelayQhi{} \\
Within-patient delay residual SD (ms) & \NppgRefResidSd & \NabpRefResidSd \\
Share of pairs with $|\text{residual}| > 50$\,ms & \NppgRefResidGt & \NabpRefResidGt \\
Share of patients with $> 10\%$ missed/extra over 0--60\,s & \NppgRefPatGtTen & \NabpRefPatGtTen \\
\bottomrule
\end{tabular}}
\end{table}

\paragraph{Statistical protocol}
Unless stated otherwise, 95\% confidence intervals are computed using 2,000 paired patient-cluster bootstrap replicates after averaging across random seeds within each window.
Where formal practical equivalence is evaluated against pre-specified margins ($\pm 0.010$ in event skill $S$ and $\pm 0.002$~mV in per-beat RMSE), equivalence requires the paired 90\% confidence interval to lie entirely within the margin.

\section{Complete comparison with forecasting baselines (RQ1)}\label{sec:s-published}

Table~\ref{tab:s-published} reports the complete comparison of \method{} against existing time-series and physiological forecasting baselines on the VitalDB ECG test cohort, including full 95\% confidence intervals and all hyperparameter and context-length sensitivity configurations accompanying Table~2 of the main text.
Informer~\cite{zhou2021}, PatchTST~\cite{nie2023}, DLinear~\cite{zeng2023}, HTF~\cite{zhang2026htf}, and FADE~\cite{ruizbarroso2025} were trained on the VitalDB training split under their respective sequence lengths and sampling rates.
In the 100-Hz long-horizon setting, DLinear and PatchTST outputs were upsampled back to 500~Hz for waveform evaluation; FADE was trained at its native 128~Hz and resampled to 500~Hz for scoring.
BeatLex~\cite{hooi2017} was applied at 500~Hz to each 120-s history to generate a 60-s continuation.
For RR-series forecasting, TimesFM~\cite{das2024} (500M-parameter checkpoint) was evaluated zero-shot on causally detected historical RR intervals with context lengths of 16, 32, 64, and 128 beats, while PatchTST-RR, DLinear-RR, AR, GRU, and GBM models were trained on historical RR sequences from the training split.

\begin{table}[!htbp]
\caption{Complete comparison against forecasting baselines on the VitalDB ECG test cohort (\NwinDisc{} windows from \NpatDisc{} patients; point estimate and 95\% patient-cluster bootstrap confidence interval). Reference scores for \method{}: 0--10-s waveform-channel skill \NwaveWWDiscFull{}, 0--2-s waveform $F_1$ \NdthreeWwFoneFull{}, and 0--10-s event-channel skill \NnatWWDiscFull{}.}\label{tab:s-published}
\centering
{\footnotesize\setlength{\tabcolsep}{4pt}\renewcommand{\arraystretch}{0.95}%
\begin{tabularx}{\linewidth}{@{}>{\raggedright\arraybackslash}Xccc@{}}
\toprule
Method and configuration & Metric & Baseline [95\% CI] & \method{} $-$ baseline [95\% CI] \\
\midrule
\multicolumn{4}{@{}l}{\emph{Raw waveform, 500\,Hz, 120\,s history $\to$ 60\,s forecast (waveform channel)}} \\
Informer~\cite{zhou2021}, primary (robust recipe) & Skill $S$, 0--10\,s & \NdthreeInfRLevFull & \NdthreeInfRDiffFull \\
Informer~\cite{zhou2021}, sensitivity (original recipe) & Skill $S$, 0--10\,s & \NdthreeInfOLevFull & \NdthreeInfODiffFull \\
BeatLex~\cite{hooi2017}, primary & Skill $S$, 0--10\,s & \NdthreeBlxRLevFull & \NdthreeBlxRDiffFull \\
HTF~\cite{zhang2026htf}, primary (original setting, 3 seeds) & Skill $S$, 0--10\,s & \NdthreeHtfOLevFull & \NdthreeHtfODiffFull \\
\midrule
\multicolumn{4}{@{}l}{\emph{Raw waveform, 500\,Hz, 10\,s history $\to$ 2\,s or 10\,s forecast (waveform channel)}} \\
PatchTST~\cite{nie2023}, primary (adequacy criterion) & $F_1$, 0--2\,s & \NdthreePtALevFull & \NdthreePtADiffFull \\
PatchTST~\cite{nie2023}, sensitivity (10 epochs) & $F_1$, 0--2\,s & \NdthreePtAtenLevFull & \NdthreePtAtenDiffFull \\
PatchTST~\cite{nie2023}, sensitivity ($\le$ 30 epochs) & $F_1$, 0--2\,s & \NdthreePtAthirtyLevFull & \NdthreePtAthirtyDiffFull \\
DLinear~\cite{zeng2023}, primary (2-s forecast) & $F_1$, 0--2\,s & \NdthreeDlALevFull & \NdthreeDlADiffFull \\
HTF~\cite{zhang2026htf}, primary (3 seeds) & $F_1$, 0--2\,s & \NdthreeHtfALevFull & \NdthreeHtfADiffFull \\
DLinear~\cite{zeng2023}, primary (10-s forecast) & Skill $S$, 0--10\,s & \NdthreeDlAtenLevFull & \NdthreeDlAtenDiffFull \\
FADE~\cite{ruizbarroso2025}, primary (2-s forecast, 3 seeds) & $F_1$, 0--2\,s & \NdthreeFadeAtwoLevFull & \NdthreeFadeAtwoDiffFull \\
FADE~\cite{ruizbarroso2025}, primary (10-s forecast, 3 seeds) & Skill $S$, 0--10\,s & \NdthreeFadeAtenLevFull & \NdthreeFadeAtenDiffFull \\
\midrule
\multicolumn{4}{@{}l}{\emph{Raw waveform, 100\,Hz, 120\,s history $\to$ 60\,s forecast (waveform channel)}} \\
DLinear~\cite{zeng2023}, primary & Skill $S$, 0--10\,s & \NdthreeDlBLevFull & \NdthreeDlBDiffFull \\
PatchTST~\cite{nie2023}, primary & Skill $S$, 0--10\,s & \NdthreePtBLevFull & \NdthreePtBDiffFull \\
\midrule
\multicolumn{4}{@{}l}{\emph{RR-series beat-time forecasting over 60\,s (event channel)}} \\
TimesFM~\cite{das2024}, primary (zero-shot, context 128 RR) & Skill $S$, 0--10\,s & \NdthreeTfmLevFull & \NdthreeTfmDiffFull \\
TimesFM~\cite{das2024}, sensitivity (context 16 RR) & Skill $S$, 0--10\,s & \NdthreeTfmSixteenLevFull & \NdthreeTfmSixteenDiffFull \\
TimesFM~\cite{das2024}, sensitivity (context 32 RR) & Skill $S$, 0--10\,s & \NdthreeTfmThirtytwoLevFull & \NdthreeTfmThirtytwoDiffFull \\
TimesFM~\cite{das2024}, sensitivity (context 64 RR) & Skill $S$, 0--10\,s & \NdthreeTfmSixtyfourLevFull & \NdthreeTfmSixtyfourDiffFull \\
PatchTST-RR~\cite{nie2023}, primary & Skill $S$, 0--10\,s & \NdthreePtRrLevFull & \NdthreePtRrDiffFull \\
DLinear-RR~\cite{zeng2023}, primary & Skill $S$, 0--10\,s & \NdthreeDlRrLevFull & \NdthreeDlRrDiffFull \\
AR(2), autoregressive RR baseline & Skill $S$, 0--10\,s & \NdthreeArTwoLevFull & \NdthreeArTwoDiffFull \\
AR(4), autoregressive RR baseline & Skill $S$, 0--10\,s & \NdthreeArFourLevFull & \NdthreeArFourDiffFull \\
AR(8), autoregressive RR baseline & Skill $S$, 0--10\,s & \NdthreeArEightLevFull & \NdthreeArEightDiffFull \\
AR(16), autoregressive RR baseline & Skill $S$, 0--10\,s & \NdthreeArSixteenLevFull & \NdthreeArSixteenDiffFull \\
GRU (MSE), recurrent RR baseline & Skill $S$, 0--10\,s & \NdthreeGruMseLevFull & \NdthreeGruMseDiffFull \\
GRU (L1 on event times), recurrent RR baseline & Skill $S$, 0--10\,s & \NdthreeGruLoneLevFull & \NdthreeGruLoneDiffFull \\
GRU (Gaussian NLL), recurrent RR baseline & Skill $S$, 0--10\,s & \NdthreeGruNllLevFull & \NdthreeGruNllDiffFull \\
GRU (residual, Huber), recurrent RR baseline & Skill $S$, 0--10\,s & \NdthreeGruHubLevFull & \NdthreeGruHubDiffFull \\
GBM (RR deltas), gradient-boosted baseline & Skill $S$, 0--10\,s & \NdthreeGbmLevFull & \NdthreeGbmDiffFull \\
GBM (RR deltas, Huber), gradient-boosted baseline & Skill $S$, 0--10\,s & \NdthreeGbmHubLevFull & \NdthreeGbmHubDiffFull \\
\bottomrule
\end{tabularx}}
\end{table}

\section{Extended ablation study and beat-level correspondence (RQ2)}\label{sec:s-grid}\label{sec:s-followup}\label{sec:s-lockbox}

\paragraph{Full confidence intervals and Lockbox replication of the conditioning ablations}
Tables~\ref{tab:s-ablation-test} and~\ref{tab:s-lockbox} provide the complete 95\% confidence intervals for the matched conditioning ablation models (Direct, Direct-MT, and \method{}) on the VitalDB ECG test cohort and the held-out Lockbox cohort, accompanying Table~4 of the main text.

\begin{table}[!htbp]
\caption{Matched conditioning ablations on the VitalDB ECG test cohort (\NwinDisc{} windows from \NpatDisc{} patients) with 95\% patient-cluster bootstrap confidence intervals.}\label{tab:s-ablation-test}
\centering
{\setlength{\tabcolsep}{5pt}%
\begin{tabularx}{\linewidth}{@{}>{\raggedright\arraybackslash}Xccc@{}}
\toprule
Metric & Direct & Direct-MT & \method{} (Ours) \\
\midrule
\multicolumn{4}{@{}l}{\emph{Original waveform loss}} \\
Event-channel skill $S$ (0--10\,s) & -- & \makecell{\NnatDirectMTDisc\\ \NnatDirectMTDiscCI} & \makecell{\NnatWWDisc\\ \NnatWWDiscCI} \\
Waveform-channel skill $S$ (0--10\,s) & \makecell{\NwaveDirectDisc\\ \NwaveDirectDiscCI} & \makecell{\NwaveDirectMTDisc\\ \NwaveDirectMTDiscCI} & \makecell{\textbf{\NwaveWWDisc}\\ \textbf{\NwaveWWDiscCI}} \\
Waveform MSE (0--60\,s) & \makecell{\NmseDirectDisc\\ \NmseDirectDiscCI} & \makecell{\NmseDirectMTDisc\\ \NmseDirectMTDiscCI} & \makecell{\NmseWWDisc\\ \NmseWWDiscCI} \\
\midrule
\multicolumn{4}{@{}l}{\emph{Split-MSE waveform loss}} \\
Event-channel skill $S$ (0--10\,s) & -- & \makecell{\NnatDirectMTSplitDisc\\ \NnatDirectMTSplitDiscCI} & \makecell{\NnatWWSplitDisc\\ \NnatWWSplitDiscCI} \\
Waveform-channel skill $S$ (0--10\,s) & \makecell{\NwaveDirectSplitDisc\\ \NwaveDirectSplitDiscCI} & \makecell{\NwaveDirectMTSplitDisc\\ \NwaveDirectMTSplitDiscCI} & \makecell{\textbf{\NwaveWWSplitDisc}\\ \textbf{\NwaveWWSplitDiscCI}} \\
Waveform MSE (0--60\,s) & \makecell{\NmseDirectSplitDisc\\ \NmseDirectSplitDiscCI} & \makecell{\NmseDirectMTSplitDisc\\ \NmseDirectMTSplitDiscCI} & \makecell{\NmseWWSplitDisc\\ \NmseWWSplitDiscCI} \\
\bottomrule
\end{tabularx}}
\end{table}

\begin{table}[!htbp]
\caption{Matched conditioning ablations on the held-out VitalDB ECG Lockbox cohort (\NwinLbx{} windows from \NpatLbx{} patients) with 95\% patient-cluster bootstrap confidence intervals.}\label{tab:s-lockbox}
\centering
{\setlength{\tabcolsep}{5pt}%
\begin{tabularx}{\linewidth}{@{}>{\raggedright\arraybackslash}Xccc@{}}
\toprule
Metric & Direct & Direct-MT & \method{} (Ours) \\
\midrule
\multicolumn{4}{@{}l}{\emph{Original waveform loss}} \\
Event-channel skill $S$ (0--10\,s) & -- & \makecell{\NnatDirectMTLbx\\ {[0.662, 0.691]}} & \makecell{\NnatWWLbx\\ {[0.667, 0.695]}} \\
Waveform-channel skill $S$ (0--10\,s) & \makecell{\NwaveDirectLbx\\ {[0.530, 0.559]}} & \makecell{\NwaveDirectMTLbx\\ {[0.526, 0.555]}} & \makecell{\textbf{\NwaveWWLbx}\\ \textbf{{[0.637, 0.665]}}} \\
Waveform MSE (0--60\,s) & \makecell{\NmseDirectLbx\\ {[0.953, 0.976]}} & \makecell{\NmseDirectMTLbx\\ {[0.951, 0.974]}} & \makecell{\NmseWWLbx\\ {[1.681, 1.711]}} \\
\midrule
\multicolumn{4}{@{}l}{\emph{Split-MSE waveform loss}} \\
Event-channel skill $S$ (0--10\,s) & -- & \makecell{\NnatDirectMTSplitLbx\\ {[0.662, 0.690]}} & \makecell{\NnatWWSplitLbx\\ {[0.652, 0.679]}} \\
Waveform-channel skill $S$ (0--10\,s) & \makecell{\NwaveDirectSplitLbx\\ {[0.552, 0.580]}} & \makecell{\NwaveDirectMTSplitLbx\\ {[0.553, 0.582]}} & \makecell{\textbf{\NwaveWWSplitLbx}\\ \textbf{{[0.623, 0.650]}}} \\
Waveform MSE (0--60\,s) & \makecell{\NmseDirectSplitLbx\\ {[1.026, 1.049]}} & \makecell{\NmseDirectMTSplitLbx\\ {[1.026, 1.050]}} & \makecell{\NmseWWSplitLbx\\ {[1.756, 1.785]}} \\
\bottomrule
\end{tabularx}}
\end{table}

\paragraph{Component substitutions, perturbation robustness, and Shapley decomposition}
Table~\ref{tab:s-grid} reports the complete 95\% confidence intervals for the stage-by-stage component ablations on the VitalDB ECG test cohort, including timing-source swaps, static history template replacements, unthresholded raw-map conditioning, and 30\% event deletion and insertion perturbations.
To quantify the relative contributions of beat-timing error versus beat-morphology error to reconstructed-waveform squared error, we also perform a two-factor Shapley decomposition.
Let $E_{ab}$ denote the mean squared error against the ground-truth future waveform when beat segments (spanning $[-250, 450]$~ms around each R peak) are placed at reference ($a=0$) or predicted ($a=1$) beat times using ground-truth ($b=0$) or generated ($b=1$) beat morphology.
The Shapley share of excess error $D = E_{11} - E_{00}$ attributable to beat timing is given by
\begin{equation}\label{eq:s-shapley}
\phi_T = \frac{1}{2}\bigl[(E_{10} - E_{00}) + (E_{11} - E_{01})\bigr], \qquad s_T = \frac{\phi_T}{E_{11} - E_{00}}.
\end{equation}
As reported in Table~\ref{tab:s-grid}, beat-timing error accounts for \NshapleyTiming{} of the excess reconstructed-waveform MSE over the full 0--60-s horizon and \NshapleyTimingZeroTwoMv{} in the $[0,2)$~s bin.

\begin{table}[!htbp]
\caption{Component ablations and interface diagnostics on the VitalDB ECG test cohort with 95\% patient-cluster bootstrap confidence intervals.}\label{tab:s-grid}
\centering
{\small\setlength{\tabcolsep}{5pt}%
\begin{tabular}{@{}lcc@{}}
\toprule
Ablation or diagnostic comparison & Metric & Estimate [95\% CI] \\
\midrule
\multicolumn{3}{@{}l}{\emph{Conditioning path and Commit ($C$) ablations (vs.\ full \method{})}} \\
Direct (w/o $P$, $C$, and $\tilde{m}$) $-$ \method{} & Waveform skill $S$ (0--10\,s) & \ensuremath{-}0.106~[\ensuremath{-}0.114, \ensuremath{-}0.099] \\
Direct-MT (with $P$, w/o $\tilde{m}$) $-$ \method{} & Waveform skill $S$ (0--10\,s) & \ensuremath{-}0.110~[\ensuremath{-}0.117, \ensuremath{-}0.102] \\
w/o Commit $C$ (raw map $\hat{m} \to G$) $-$ \method{} & Waveform skill $S$ (0--10\,s) & \NifSoftFull \\
w/o Commit $C$ (raw map $\hat{m} \to G$) $-$ \method{} & Per-beat RMSE (mV) & \NsoftBeatRmseDiscFull \\
Direct waveform peaks $\to G$ $-$ \method{} & Waveform skill $S$ (0--10\,s) & \NdirectReadRendererFull \\
Direct (Split) waveform peaks $\to G$ $-$ \method{} & Waveform skill $S$ (0--10\,s) & \NufSplitDirectRendererFull \\
\midrule
\multicolumn{3}{@{}l}{\emph{Timing predictor ($P$) and waveform constructor ($G$ vs.\ static template) ablations}} \\
Direct-MT timing $\to G$ $-$ \method{} (Test) & Waveform skill $S$ (0--10\,s) & \NheadRendererDiscFull \\
Direct-MT timing $\to G$ $-$ \method{} (Lockbox) & Waveform skill $S$ (0--10\,s) & \NheadRendererLbxFull \\
Direct-MT (Split) timing $\to G$ $-$ \method{} & Waveform skill $S$ (0--10\,s) & \NufSplitHeadRendererFull \\
\method{} timing $\to$ Static template $-$ \method{} & Waveform skill $S$ (0--10\,s) & \ensuremath{-}0.0045~[\ensuremath{-}0.0060, \ensuremath{-}0.0033] \\
Direct-MT timing $\to$ Static template $-$ \method{} & Waveform skill $S$ (0--10\,s) & \NufHeadTemplateFull \\
Guarded cadence $\to$ Static template $-$ \method{} & Waveform skill $S$ (0--10\,s) & \NguardedTplMinusWWFull \\
\midrule
\multicolumn{3}{@{}l}{\emph{Resilience to 30\% input event perturbations and Shapley timing share}} \\
Generator $G$ $-$ Static template retention (30\% event deletion) & Waveform skill $S$ (0--10\,s) & \NifDelRobustFull \\
Generator $G$ $-$ Static template retention (30\% event insertion) & Waveform skill $S$ (0--10\,s) & \NifInsRobustFull \\
Shapley share of timing error in excess MSE (0--60\,s) & Share of mV$^2$ error & \NshapleyTimingFull \\
Shapley share of timing error in excess MSE (0--2\,s) & Share of mV$^2$ error & \NshapleyTimingZeroTwoMvFull \\
\bottomrule
\end{tabular}}
\end{table}

\begin{figure}[!htbp]
\centering
\includegraphics[width=\linewidth]{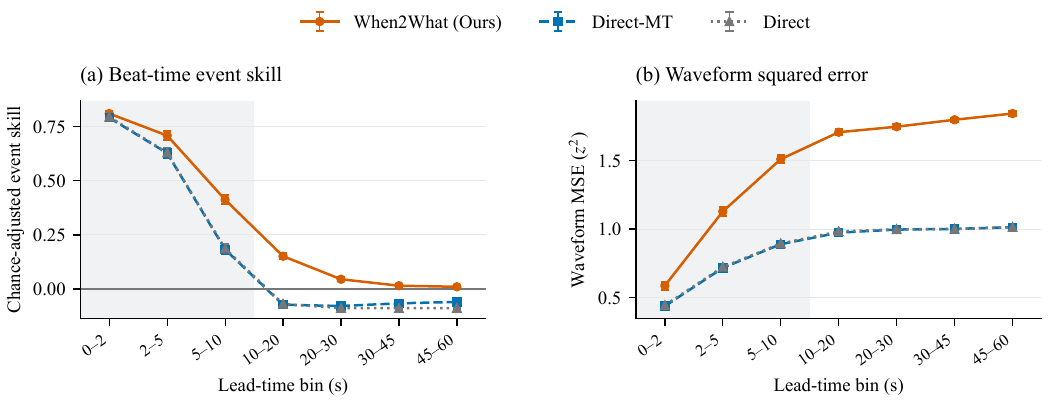}
\caption{Matched lead-time comparison across $[0,2)$, $[2,5)$, $[5,10)$, $[10,20)$, $[20,30)$, $[30,45)$, and $[45,60)$~s on the ECG test cohort (95\% patient-cluster intervals). (a)~Waveform-channel event skill $S$ at $\pm 50$~ms. (b)~Samplewise waveform MSE ($z^2$).}\label{fig:s-matched}
\end{figure}

\begin{figure}[!htbp]
\centering
\includegraphics[width=\linewidth]{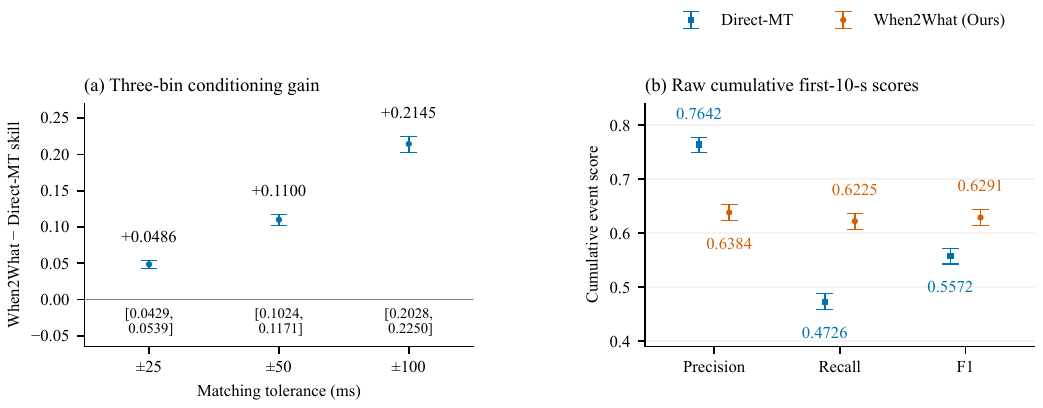}
\caption{Matching-tolerance sensitivity and cumulative 0--10-s detection scores on the ECG test cohort (95\% patient-cluster intervals). (a)~Conditioning gain (\method{} minus Direct-MT in 0--10-s event skill $S$) at $\pm 25$, $\pm 50$, and $\pm 100$~ms. (b)~Cumulative 0--10-s precision, recall, and $F_1$ at $\pm 50$~ms.}\label{fig:s-sensitivity}
\end{figure}

\paragraph{Bin-by-bin 60-s lead-time profiles on the Test and Lockbox cohorts}
Figure~\ref{fig:s-matched} and Tables~\ref{tab:s-bins-test} and~\ref{tab:s-bins-lockbox} report the complete bin-by-bin progression of waveform-channel event skill $S$ and samplewise waveform MSE ($z^2$) across all seven lead-time intervals ($[0,2)$, $[2,5)$, $[5,10)$, $[10,20)$, $[20,30)$, $[30,45)$, and $[45,60)$~s) on both the VitalDB ECG test cohort and the held-out Lockbox cohort.
In both cohorts and across every lead-time bin, \method{} achieves higher event skill than Direct and Direct-MT while incurring higher pointwise MSE due to sharp peak commitment.
Moreover, replacing the learned generator $G$ with a static historical beat template incurs substantially higher waveform MSE at longer horizons on both cohorts ($1.945$ and $2.058$ in $[30,45)$ and $[45,60)$~s on Test, and $1.957$ and $2.067$ on Lockbox).
Table~\ref{tab:s-tolerance} and Fig.~\ref{fig:s-sensitivity} report sensitivity across matching tolerances ($\pm 25$, $\pm 50$, and $\pm 100$~ms) and cumulative 0--10-s precision, recall, and $F_1$.

\begin{table}[!htbp]
\caption{Bin-by-bin waveform-channel event skill $S$ and samplewise waveform MSE ($z^2$) across the 60-s forecast horizon on the VitalDB ECG test cohort (\NwinDisc{} windows from \NpatDisc{} patients).}\label{tab:s-bins-test}
\centering
{\small\setlength{\tabcolsep}{4.5pt}%
\begin{tabular}{@{}lcccccccc@{}}
\toprule
Model variant & $[0,2)$\,s & $[2,5)$\,s & $[5,10)$\,s & $[10,20)$\,s & $[20,30)$\,s & $[30,45)$\,s & $[45,60)$\,s & Overall \\
\midrule
\multicolumn{9}{@{}l}{\emph{Waveform-channel event skill $S$ ($\pm 50$\,ms; Overall = 0--10\,s primary mean)}} \\
Direct & 0.795 & 0.629 & 0.187 & \ensuremath{-}0.068 & \ensuremath{-}0.088 & \ensuremath{-}0.088 & \ensuremath{-}0.088 & 0.537 \\
Direct-MT & 0.791 & 0.627 & 0.183 & \ensuremath{-}0.072 & \ensuremath{-}0.078 & \ensuremath{-}0.066 & \ensuremath{-}0.059 & 0.533 \\
\method{} timing $\to$ Template & 0.806 & 0.704 & 0.407 & 0.142 & 0.036 & 0.011 & 0.010 & 0.639 \\
\textbf{\method{} (Ours)} & \textbf{0.810} & \textbf{0.708} & \textbf{0.412} & \textbf{0.152} & \textbf{0.045} & \textbf{0.016} & \textbf{0.011} & \textbf{0.644} \\
\midrule
\multicolumn{9}{@{}l}{\emph{Samplewise waveform MSE in history-normalized units ($z^2$; Overall = 0--60\,s pooled)}} \\
Direct & 0.444 & 0.724 & 0.894 & 0.982 & 0.998 & 1.001 & 1.014 & 0.959 \\
Direct-MT & 0.438 & 0.717 & 0.890 & 0.974 & 0.996 & 1.001 & 1.014 & 0.957 \\
\method{} timing $\to$ Template & 0.637 & 1.133 & 1.514 & 1.714 & 1.790 & 1.945 & 2.058 & 1.789 \\
\textbf{\method{} (Ours)} & 0.587 & 1.129 & 1.510 & 1.707 & 1.747 & 1.797 & 1.843 & 1.687 \\
\bottomrule
\end{tabular}}
\end{table}

\begin{table}[!htbp]
\caption{Bin-by-bin waveform-channel event skill $S$ and samplewise waveform MSE ($z^2$) across the 60-s forecast horizon on the held-out VitalDB ECG Lockbox cohort (\NwinLbx{} windows from \NpatLbx{} patients).}\label{tab:s-bins-lockbox}
\centering
{\small\setlength{\tabcolsep}{4.5pt}%
\begin{tabular}{@{}lcccccccc@{}}
\toprule
Model variant & $[0,2)$\,s & $[2,5)$\,s & $[5,10)$\,s & $[10,20)$\,s & $[20,30)$\,s & $[30,45)$\,s & $[45,60)$\,s & Overall \\
\midrule
\multicolumn{9}{@{}l}{\emph{Waveform-channel event skill $S$ ($\pm 50$\,ms; Overall = 0--10\,s primary mean)}} \\
Direct & 0.815 & 0.647 & 0.170 & \ensuremath{-}0.071 & \ensuremath{-}0.089 & \ensuremath{-}0.091 & \ensuremath{-}0.089 & 0.544 \\
Direct-MT & 0.814 & 0.642 & 0.167 & \ensuremath{-}0.078 & \ensuremath{-}0.078 & \ensuremath{-}0.069 & \ensuremath{-}0.063 & 0.541 \\
\method{} timing $\to$ Template & 0.827 & 0.718 & 0.396 & 0.130 & 0.030 & 0.007 & 0.009 & 0.647 \\
\textbf{\method{} (Ours)} & \textbf{0.831} & \textbf{0.721} & \textbf{0.402} & \textbf{0.141} & \textbf{0.041} & \textbf{0.011} & \textbf{0.010} & \textbf{0.651} \\
\midrule
\multicolumn{9}{@{}l}{\emph{Samplewise waveform MSE in history-normalized units ($z^2$; Overall = 0--60\,s pooled)}} \\
Direct & 0.487 & 0.720 & 0.895 & 0.977 & 1.003 & 1.010 & 1.020 & 0.964 \\
Direct-MT & 0.482 & 0.716 & 0.891 & 0.970 & 1.001 & 1.010 & 1.020 & 0.962 \\
\method{} timing $\to$ Template & 0.679 & 1.154 & 1.537 & 1.710 & 1.788 & 1.957 & 2.067 & 1.798 \\
\textbf{\method{} (Ours)} & 0.625 & 1.148 & 1.533 & 1.704 & 1.748 & 1.810 & 1.850 & 1.696 \\
\bottomrule
\end{tabular}}
\end{table}

\begin{table}[!htbp]
\caption{Matching-tolerance sensitivity and cumulative 0--10-s detection metrics on the VitalDB ECG test cohort (\NwinDisc{} windows from \NpatDisc{} patients) with 95\% patient-cluster bootstrap confidence intervals.}\label{tab:s-tolerance}
\centering
{\small\setlength{\tabcolsep}{3pt}%
\begin{tabularx}{\linewidth}{@{}>{\raggedright\arraybackslash}X>{\raggedright\arraybackslash}Xccc@{}}
\toprule
Metric and tolerance & Channel & Direct-MT [95\% CI] & \method{} [95\% CI] & \method{} $-$ Direct-MT [95\% CI] \\
\midrule
Event skill $S$ ($\pm 25$\,ms) & Waveform & 0.428~[0.413, 0.443] & 0.476~[0.461, 0.491] & \ensuremath{+}0.049~[\ensuremath{+}0.043, \ensuremath{+}0.054] \\
Event skill $S$ ($\pm 50$\,ms, primary) & Waveform & 0.533~[0.517, 0.550] & 0.644~[0.628, 0.659] & \ensuremath{+}0.110~[\ensuremath{+}0.102, \ensuremath{+}0.117] \\
Event skill $S$ ($\pm 100$\,ms) & Waveform & 0.552~[0.534, 0.571] & 0.767~[0.752, 0.782] & \ensuremath{+}0.214~[\ensuremath{+}0.203, \ensuremath{+}0.225] \\
\midrule
Cumulative precision ($\pm 50$\,ms) & Waveform & 0.764~[0.750, 0.777] & 0.638~[0.623, 0.653] & -- \\
Cumulative recall ($\pm 50$\,ms) & Waveform & 0.473~[0.458, 0.488] & 0.622~[0.608, 0.637] & -- \\
Cumulative $F_1$ ($\pm 50$\,ms) & Waveform & 0.557~[0.543, 0.572] & 0.629~[0.614, 0.644] & -- \\
\bottomrule
\end{tabularx}}
\end{table}

\paragraph{Beat-level correspondence analysis}
To measure how often reference beats correctly localized by a model's timing head in $[0,10)$~s are also detected as R peaks in its generated waveform, each reference beat is classified after full-horizon $\pm 50$-ms greedy matching into $N_{11}$ (detected in both the timing head and output waveform), $N_{10}$ (timing head only), $N_{01}$ (waveform only), or $N_{00}$ (neither).
The beat reproduction (expression) rate is $N_{11}/(N_{11}+N_{10})$, computed from seed-averaged, patient-summed counts with 2,000-replicate paired patient-cluster bootstrap 95\% confidence intervals.
Strict expression additionally requires the matched timing-head and waveform peaks to lie within $50$~ms of each other.
Tables~\ref{tab:s-follow-expr}, \ref{tab:s-follow-counts}, and~\ref{tab:s-follow-boundary} report the complete rates, paired differences, event counts, and boundary-crossing statistics on the Test and Lockbox cohorts under the NeuroKit2 detector.

\begin{table}[!htbp]
\caption{Beat-level reproduction (expression) rates and paired differences [95\% patient-cluster CI] on the VitalDB ECG Test and Lockbox cohorts.}\label{tab:s-follow-expr}
\centering
{\small\setlength{\tabcolsep}{6pt}%
\begin{tabularx}{\linewidth}{@{}>{\raggedright\arraybackslash}X>{\raggedright\arraybackslash}Xcc@{}}
\toprule
Cohort & Model or paired contrast & Expression rate [95\% CI] & Strict expression rate [95\% CI] \\
\midrule
Test & Direct-MT & 72.04\%~[70.83\%, 73.38\%] & 72.00\%~[70.79\%, 73.34\%] \\
Test & \method{} (Ours) & \textbf{95.62\%~[95.38\%, 95.88\%]} & \textbf{95.62\%~[95.38\%, 95.88\%]} \\
Test & \method{} $-$ Direct-MT ($\Delta$) & \textbf{\ensuremath{+}0.2358~[\ensuremath{+}0.2232, \ensuremath{+}0.2475]} & \textbf{\ensuremath{+}0.2362~[\ensuremath{+}0.2235, \ensuremath{+}0.2480]} \\
\midrule
Lockbox & Direct-MT & 72.15\%~[70.93\%, 73.40\%] & 72.11\%~[70.89\%, 73.36\%] \\
Lockbox & \method{} (Ours) & \textbf{95.49\%~[95.22\%, 95.75\%]} & \textbf{95.49\%~[95.22\%, 95.75\%]} \\
Lockbox & \method{} $-$ Direct-MT ($\Delta$) & \textbf{\ensuremath{+}0.2334~[\ensuremath{+}0.2212, \ensuremath{+}0.2452]} & \textbf{\ensuremath{+}0.2338~[\ensuremath{+}0.2216, \ensuremath{+}0.2458]} \\
\bottomrule
\end{tabularx}}
\end{table}

\begin{table}[!htbp]
\caption{Seed-averaged reference beat categories ($N_{11}, N_{10}, N_{01}, N_{00}$) and unmatched predictions in the first 10~s on the Test and Lockbox cohorts.}\label{tab:s-follow-counts}
\centering
{\small\setlength{\tabcolsep}{6pt}%
\begin{tabular}{@{}llrrrrrr@{}}
\toprule
Cohort & Model & $N_{11}$ & $N_{10}$ & $N_{01}$ & $N_{00}$ & Unmatched head & Unmatched wave \\
\midrule
Test & Direct-MT & 12730.2 & 4941.0 & 356.0 & 9280.8 & 8902.4 & 3224.6 \\
Test & \method{} (Ours) & 16992.2 & 778.4 & 111.0 & 9426.4 & 8889.0 & 9390.0 \\
\midrule
Lockbox & Direct-MT & 12798.4 & 4939.6 & 335.4 & 9145.6 & 8942.0 & 3125.6 \\
Lockbox & \method{} (Ours) & 17064.2 & 806.2 & 110.0 & 9238.6 & 8892.8 & 9358.0 \\
\bottomrule
\end{tabular}}
\end{table}

\begin{table}[!htbp]
\caption{Boundary-crossing matches at $t = 10$~s (reference $<10$~s matched to prediction $\ge 10$~s, ``out''; or the reverse, ``in'') and windows/patients with zero timing-head matches.}\label{tab:s-follow-boundary}
\centering
{\small\setlength{\tabcolsep}{6pt}%
\begin{tabular}{@{}llrrrrrr@{}}
\toprule
Cohort & Model & Head out & Head in & Wave out & Wave in & Zero-head windows & Zero-head patients \\
\midrule
Test & Direct-MT & 11.6 & 11.4 & 2.0 & 3.2 & 54.4 & 2.0 \\
Test & \method{} (Ours) & 10.6 & 13.0 & 11.0 & 13.0 & 52.0 & 2.0 \\
\midrule
Lockbox & Direct-MT & 11.6 & 15.2 & 3.2 & 5.2 & 34.4 & 0.0 \\
Lockbox & \method{} (Ours) & 11.8 & 15.4 & 12.0 & 15.6 & 32.8 & 0.0 \\
\bottomrule
\end{tabular}}
\end{table}

\FloatBarrier
\section{Extended generalization: loss tuning, external cohorts, pulse signals, and detector sensitivity (RQ3)}\label{sec:s-loss}\label{sec:s-detector}

\paragraph{Loss function tuning, backbone architectures, and external ECG cohorts}
Tables~\ref{tab:s-dcond-full} and~\ref{tab:s-external-strata} provide the complete 95\% confidence intervals for all generalization settings in Table~5 of the main text.
To verify that the conditioning gain under FADE-style Split-MSE~\cite{ruizbarroso2025} does not depend on default loss weights, we symmetrically screened five Split-MSE configurations on validation data for both Direct-MT~(Split) and \method{}~(Split): inner-to-outer weight ratios of $5:1$, $1:1$, and $1:5$ at the default amplitude band $|y|\le 1.6$, and band widths scaled by $0.75\times$ and $1.25\times$ at ratio $5:1$.
Each configuration was screened for 20,000 steps and the selected configuration was trained for 60,000 steps across three random seeds.
As shown in Table~\ref{tab:s-dcond-full}, both the default and tuned Split-MSE configurations preserve a large, statistically significant conditioning gain, whereas applying Split-MSE without explicit timing conditioning recovers only \NrecoveryDisc{} of the original-loss conditioning gap (\NrecoveryTunedDisc{} when tuned).
We also tested adding the shape- and temporal-distortion loss DILATE~\cite{leguen2019} to Direct regression in a single default configuration; because its initial gradient magnitude was much larger than the waveform loss, it reached a waveform skill of only \NdilateSkill{}.
Across external cohorts (Table~\ref{tab:s-external-strata}), zero-shot transfer holds on both MOVER and SHDB-AF, and remains virtually identical whether SHDB-AF bootstrap resampling is clustered by Holter record or by unique participant.

\begin{table}[!htbp]
\caption{Complete 95\% confidence intervals for the conditioning gain (\method{} minus Direct-MT in 0--10-s waveform-channel skill $S$) across losses, backbones, and cohorts.}\label{tab:s-dcond-full}
\centering
{\small\setlength{\tabcolsep}{5pt}%
\begin{tabularx}{\linewidth}{@{}>{\raggedright\arraybackslash}X>{\raggedright\arraybackslash}Xccc@{}}
\toprule
Experimental setting & Cohort (clustering unit) & Direct-MT & \method{} (Ours) & Gain $\Delta$ [95\% CI] \\
\midrule
Original loss & Test (845 patients) & 0.533 & \textbf{0.644} & \textbf{\NdcondOrigDiscFull} \\
Split-MSE loss & Test (845 patients) & 0.560 & \textbf{0.627} & \textbf{\NdcondSplitDiscFull} \\
Split-MSE loss (tuned) & Test (845 patients) & 0.557 & \textbf{0.631} & \textbf{\NdcondSplitTunedDiscFull} \\
TCN backbone, original loss & Test (845 patients) & 0.534 & \textbf{0.631} & \textbf{\NdcondTCNDiscFull} \\
\midrule
Original loss & Lockbox (832 patients) & 0.541 & \textbf{0.651} & \textbf{\NdcondOrigLbxFull} \\
Split-MSE loss & Lockbox (832 patients) & 0.567 & \textbf{0.637} & \textbf{\NdcondSplitLbxFull} \\
Original loss (zero-shot) & MOVER (574 patients) & 0.534 & \textbf{0.632} & \textbf{\NdcondMOVERFull} \\
Original loss (zero-shot) & SHDB-AF (98 records) & 0.238 & \textbf{0.348} & \textbf{\NdcondSHDBFull} \\
Original loss (zero-shot) & SHDB-AF (93 participants) & 0.238 & \textbf{0.348} & \textbf{\ensuremath{+}0.110~[\ensuremath{+}0.100, \ensuremath{+}0.121]} \\
\bottomrule
\end{tabularx}}
\end{table}

\begin{table}[!htbp]
\caption{Event-channel and waveform-channel comparisons across VitalDB, MOVER, and SHDB-AF rhythm strata [95\% CI].}\label{tab:s-external-strata}
\centering
{\small\setlength{\tabcolsep}{2.5pt}%
\begin{tabularx}{\linewidth}{@{}>{\raggedright\arraybackslash}X>{\raggedright\arraybackslash}X>{\raggedright\arraybackslash}Xcc@{}}
\toprule
Cohort and stratum & Windows / clusters & Model comparison & Channel & Difference [95\% CI] \\
\midrule
VitalDB ECG Test & \NwinDisc{} / \NpatDisc{} & \method{} $-$ Direct-MT & Waveform & \NdcondOrigDiscFull \\
VitalDB ECG Test & \NwinDisc{} / \NpatDisc{} & \method{} $-$ Guarded cadence & Event & \NguardedDiffDiscFull \\
VitalDB ECG Lockbox & \NwinLbx{} / \NpatLbx{} & \method{} $-$ Direct-MT & Waveform & \NdcondOrigLbxFull \\
VitalDB ECG Lockbox & \NwinLbx{} / \NpatLbx{} & \method{} $-$ Guarded cadence & Event & \NguardedDiffLbxFull \\
\midrule
External MOVER (all) & \NmoverWin{} / \NmoverPat{} & \method{} $-$ Direct-MT & Waveform & \NdcondMOVERFull \\
External MOVER (all) & \NmoverWin{} / \NmoverPat{} & \method{} $-$ Guarded cadence & Event & \NmoverGuardedDiffFull \\
External SHDB-AF (all) & \NshdbWin{} / \NshdbRec{} & \method{} $-$ Direct-MT & Waveform & \NdcondSHDBFull \\
External SHDB-AF (AF subset) & \NshdbAfWin{} / \NshdbAfRec{} & \method{} $-$ Guarded cadence & Event & \NafGuardedDiffFull \\
\bottomrule
\end{tabularx}}
\end{table}

\paragraph{Complete confidence intervals on ABP and PPG pulse waveforms}
Table~\ref{tab:s-pulse-full} reports the complete 95\% patient-cluster bootstrap confidence intervals for the ABP and PPG evaluations accompanying Table~6 of the main text.

\begin{table}[!htbp]
\caption{Complete 95\% confidence intervals for ABP and PPG pulse waveform evaluations over the first 10~s.}\label{tab:s-pulse-full}
\centering
{\small\setlength{\tabcolsep}{5pt}%
\begin{tabularx}{\linewidth}{@{}>{\raggedright\arraybackslash}X>{\raggedright\arraybackslash}Xccc@{}}
\toprule
Signal & Evaluation metric & Baseline & \method{} (Ours) & Difference $\Delta$ [95\% CI] \\
\midrule
PPG & Waveform-channel event $F_1$ (0--10\,s, vs.\ Direct) & 0.671 & \textbf{0.686} & \textbf{\NppgWaveFoneVsDirectFull} \\
PPG & Event-channel skill $S$ (0--10\,s, vs.\ simple RR rule) & 0.621 & \textbf{0.666} & \textbf{\NppgSkillVsNaiveFull} \\
PPG & Predictability time scale $\tau$ (s, vs.\ simple RR rule) & -- & -- & \NppgTauVsNaiveFull \\
\midrule
ABP & Waveform-channel event $F_1$ (0--10\,s, vs.\ Direct) & 0.681 & \textbf{0.719} & \textbf{\NabpWaveFoneVsDirectFull} \\
ABP & Event-channel skill $S$ (0--10\,s, vs.\ simple RR rule) & 0.660 & \textbf{0.679} & \textbf{\NabpSkillVsNaiveFull} \\
ABP & Predictability time scale $\tau$ (s, vs.\ simple RR rule) & -- & -- & \NabpTauVsNaiveFull \\
\bottomrule
\end{tabularx}}
\end{table}

\paragraph{Waveform detector sensitivity}
In the main manuscript, we use the standard NeuroKit2 R-peak detector~\cite{makowski2021} as our waveform readout because it achieves the highest accuracy on ground-truth future ECG waveforms.
To verify that the conditioning gain of \method{} over Direct-MT does not depend on a single peak-detection algorithm, we also evaluated two validation-tuned local-percentile threshold detectors: a shared threshold detector tuned jointly across models, and a per-model threshold detector tuned separately for each model on validation forecasts.
Table~\ref{tab:s-readout} shows that \method{} achieves a large, statistically significant conditioning gain over Direct-MT across all three detectors.

\begin{table}[!htbp]
\caption{Waveform readout detector sensitivity on the VitalDB ECG validation and test cohorts [95\% CI].}\label{tab:s-readout}
\centering
\begin{tabularx}{\linewidth}{@{}>{\raggedright\arraybackslash}Xccc@{}}
\toprule
Metric & NeuroKit2 (standard) & Shared threshold detector & Per-model threshold detector \\
\midrule
$F_1$ on ground-truth ECG (validation) & \NroCFoneVal & \NroAFoneVal & -- \\
$F_1$ on ground-truth ECG (test) & \NroCFoneDiscFull & \NroAFoneDiscFull & -- \\
Conditioning gain, original loss (test) & \NdcondOrigDiscFull & \NdcondOrigDiscRoAFull & \NdcondOrigDiscRoBFull \\
Split-MSE / original gain ratio (test) & \NratioDiscFull & \NratioDiscRoAFull & \NratioDiscRoBFull \\
\bottomrule
\end{tabularx}
\end{table}

\FloatBarrier
\section{Complete synthetic experiment grid (RQ4)}\label{sec:s-synthetic}

\paragraph{Synthetic generator and experimental protocol}
In the controlled synthetic experiment (Section~4.5 of the main text), 100-Hz ECG recordings (20-s history, 10-s horizon) are generated using the dynamical model of McSharry et al.~\cite{mcsharry2003} across four levels of beat-time innovation standard deviation (T1--T4) and four levels of per-beat morphology variability (M1--M4; Table~\ref{tab:s-generator}).
Each cell of the $4 \times 4$ grid comprises \num{12000} training, \num{1000} validation, and 600 test records.
For each test record, 64 Monte-Carlo future realizations from the same history are averaged to form the theoretical MSE-optimal waveform (average of possible futures)~\cite{gneiting2011}.
Direct regression, multi-task, and event-conditioned models share a convolutional backbone and are trained across three random seeds per cell.
Tables~\ref{tab:s-synthetic} and~\ref{tab:s-soft} report the complete 16-cell results with 95\% record-bootstrap confidence intervals for committed-event inference ($\tilde{m}$) and unthresholded soft-map inference ($\hat{m}$), respectively.

\begin{table}[!htbp]
\caption{Synthetic ECG generator parameters across timing uncertainty levels (T1--T4) and morphology variability levels (M1--M4).}\label{tab:s-generator}
\centering
\begin{tabular}{@{}lcccc@{}}
\toprule
Timing uncertainty level & T1 & T2 & T3 & T4 \\
\midrule
Per-beat RR innovation SD $\sigma_\varepsilon$ (ms) & 2 & 6 & 18 & 54 \\
Conditional beat-time SD over first 8 beats (ms) & \NsynTimingSDlow & \NsynTimingSDmid & \NsynTimingSDthree & \NsynTimingSDhigh \\
\midrule
Morphology variability level & M1 & M2 & M3 & M4 \\
\midrule
Amplitude/width log-normal jitter $\sigma_\eta$ & 0.02 & 0.06 & 0.15 & 0.35 \\
Per-beat template-switching probability $h$ & 0.00 & 0.01 & 0.04 & 0.12 \\
\bottomrule
\end{tabular}
\end{table}

\begin{table}[!htbp]
\caption{Complete $4 \times 4$ synthetic grid (\method{} committed forecast minus Direct regression) with 95\% record-bootstrap confidence intervals.}\label{tab:s-synthetic}
\centering
\begin{tabular}{@{}lcc@{}}
\toprule
Grid cell & Event $F_1$ difference [95\% CI] & Squared error difference [95\% CI] \\
\midrule
T1M1 & \NsynGainToneMoneFull & \NsynMseToneMoneFull \\
T1M2 & \NsynGainToneMtwoFull & \NsynMseToneMtwoFull \\
T1M3 & \NsynGainToneMthreeFull & \NsynMseToneMthreeFull \\
T1M4 & \NsynGainToneMfourFull & \NsynMseToneMfourFull \\
T2M1 & \NsynGainTtwoMoneFull & \NsynMseTtwoMoneFull \\
T2M2 & \NsynGainTtwoMtwoFull & \NsynMseTtwoMtwoFull \\
T2M3 & \NsynGainTtwoMthreeFull & \NsynMseTtwoMthreeFull \\
T2M4 & \NsynGainTtwoMfourFull & \NsynMseTtwoMfourFull \\
T3M1 & \NsynGainTthreeMoneFull & \NsynMseTthreeMoneFull \\
T3M2 & \NsynGainTthreeMtwoFull & \NsynMseTthreeMtwoFull \\
T3M3 & \NsynGainTthreeMthreeFull & \NsynMseTthreeMthreeFull \\
T3M4 & \NsynGainTthreeMfourFull & \NsynMseTthreeMfourFull \\
T4M1 & \NsynGainTfourMoneFull & \NsynMseTfourMoneFull \\
T4M2 & \NsynGainTfourMtwoFull & \NsynMseTfourMtwoFull \\
T4M3 & \NsynGainTfourMthreeFull & \NsynMseTfourMthreeFull \\
T4M4 & \NsynGainTfourMfourFull & \NsynMseTfourMfourFull \\
\bottomrule
\end{tabular}
\end{table}

\begin{table}[!htbp]
\caption{Soft-map inference variant ($\hat{m}$ without discrete commit sharpening) minus Direct regression across the $4 \times 4$ synthetic grid [95\% CI].}\label{tab:s-soft}
\centering
\begin{tabular}{@{}lcc@{}}
\toprule
Grid cell & Event $F_1$ difference [95\% CI] & Squared error difference [95\% CI] \\
\midrule
T1M1 & \NsynSoftGainToneMoneFull & \NsynSoftMseToneMoneFull \\
T1M2 & \NsynSoftGainToneMtwoFull & \NsynSoftMseToneMtwoFull \\
T1M3 & \NsynSoftGainToneMthreeFull & \NsynSoftMseToneMthreeFull \\
T1M4 & \NsynSoftGainToneMfourFull & \NsynSoftMseToneMfourFull \\
T2M1 & \NsynSoftGainTtwoMoneFull & \NsynSoftMseTtwoMoneFull \\
T2M2 & \NsynSoftGainTtwoMtwoFull & \NsynSoftMseTtwoMtwoFull \\
T2M3 & \NsynSoftGainTtwoMthreeFull & \NsynSoftMseTtwoMthreeFull \\
T2M4 & \NsynSoftGainTtwoMfourFull & \NsynSoftMseTtwoMfourFull \\
T3M1 & \NsynSoftGainTthreeMoneFull & \NsynSoftMseTthreeMoneFull \\
T3M2 & \NsynSoftGainTthreeMtwoFull & \NsynSoftMseTthreeMtwoFull \\
T3M3 & \NsynSoftGainTthreeMthreeFull & \NsynSoftMseTthreeMthreeFull \\
T3M4 & \NsynSoftGainTthreeMfourFull & \NsynSoftMseTthreeMfourFull \\
T4M1 & \NsynSoftGainTfourMoneFull & \NsynSoftMseTfourMoneFull \\
T4M2 & \NsynSoftGainTfourMtwoFull & \NsynSoftMseTfourMtwoFull \\
T4M3 & \NsynSoftGainTfourMthreeFull & \NsynSoftMseTfourMthreeFull \\
T4M4 & \NsynSoftGainTfourMfourFull & \NsynSoftMseTfourMfourFull \\
\bottomrule
\end{tabular}
\end{table}

\FloatBarrier
\section{Retrospective beat-synchronized device triggering analysis}\label{sec:s-use}

To evaluate whether near-term beat predictions can support beat-synchronized clinical actuation under a hardware execution delay, we conducted retrospective first-beat and streaming trigger simulations accompanying Section~5.1 of the main text.
Table~\ref{tab:s-trigger} summarizes the relative reduction in mistimed triggers achieved by \method{} compared with the guarded cadence rule across the VitalDB test and Lockbox cohorts.

\begin{table}[!htbp]
\caption{Relative reduction in mistimed beat triggers achieved by \method{} versus the guarded cadence rule [95\% CI].}\label{tab:s-trigger}
\centering
\begin{tabularx}{\linewidth}{@{}>{\raggedright\arraybackslash}X>{\raggedright\arraybackslash}Xcc@{}}
\toprule
Simulation task and setting & Cohort & Relative reduction & 95\% CI \\
\midrule
Immediate next heartbeat ($\pm 50$\,ms tolerance, all beats) & VitalDB Test & \NfirstBeatDisc & [11.0\%, 28.5\%] \\
Immediate next heartbeat ($\pm 50$\,ms tolerance, non-ectopic) & VitalDB Test & \NfirstBeatNonEct & [14.9\%, 35.1\%] \\
Immediate next heartbeat ($\pm 50$\,ms tolerance, ectopic beats) & VitalDB Test & \NectopicDisc & [\ensuremath{-}16.8\%, \ensuremath{+}10.9\%] \\
Immediate next heartbeat ($\pm 50$\,ms tolerance, all beats) & VitalDB Lockbox & \NfirstBeatLbx & [2.2\%, 28.2\%] \\
\midrule
Streaming device trigger (50-ms delay, $\pm 50$\,ms tolerance) & VitalDB Test & \NdeviceFifty & [12.9\%, 26.2\%] \\
Streaming device trigger (50-ms delay, $\pm 50$\,ms tolerance) & VitalDB Lockbox & \NdeviceFiftyLbx & [16.3\%, 27.7\%] \\
Streaming device trigger (100-ms delay, $\pm 20$\,ms tolerance) & VitalDB Test & \NdeviceHundred & [\ensuremath{-}4.4\%, \ensuremath{+}4.7\%] \\
Streaming device trigger (50-ms delay, $\pm 50$\,ms tolerance) & External MOVER & \NdeviceFiftyMover & [\ensuremath{-}18.8\%, \ensuremath{+}2.1\%] \\
\bottomrule
\end{tabularx}
\end{table}

\FloatBarrier

\end{document}